\documentclass[10pt,twocolumn,letterpaper]{article}

\usepackage{cvpr}              

\usepackage{graphicx}
\usepackage{braket}
\usepackage{amsmath}
\usepackage{amssymb}

\usepackage{pifont}
\usepackage{xcolor}

\usepackage{booktabs}
\usepackage{subcaption}
\usepackage{ragged2e}
\usepackage{adjustbox}

\definecolor{cvprblue}{rgb}{0.21,0.49,0.74}
\usepackage[pagebackref,breaklinks,colorlinks,allcolors=cvprblue]{hyperref}

\def\paperID{} 
\def\confName{CVPR}
\def\confYear{2026}

\title{CLIP-like Model as a Foundational Density Ratio Estimator}

\author{
Fumiya Uchiyama$^{1,2}$ \quad
Rintaro Yanagi$^{2}$ \quad
Shohei Taniguchi$^{1}$ \quad
Shota Takashiro$^{1}$ \\
Masahiro Suzuki$^{1}$ \quad
Hirokatsu Kataoka$^{2}$ \quad
Yusuke Iwasawa$^{1}$ \quad
Yutaka Matsuo$^{1}$
\\[0.5em]
$^{1}$The University of Tokyo \quad $^{2}$AIST
\\[0.3em]
{\tt\small fumiya.uchiyama@weblab.t.u-tokyo.ac.jp}
}

\begin{document}
\maketitle
\begin{abstract}
Density ratio estimation is a core concept in statistical machine learning because it provides a unified mechanism for tasks such as importance weighting, divergence estimation, and likelihood-free inference, but its potential in vision and language models has not been fully explored. Modern vision-language encoders such as CLIP and SigLIP are trained with contrastive objectives that implicitly optimize log density ratios between joint and marginal image–text distributions, which implicitly learn similarity scores proportional to log density ratios. However, prior work has largely focused on their embedding utility, and the density-ratio structure induced by contrastive learning has not been systematically examined or exploited in multimodal applications.
To address this gap, we reinterpret CLIP-style models as pretrained and general-purpose density ratio estimators and show that this perspective enables new algorithmic capabilities.
We present a unified explanation of how contrastive objectives estimate density ratios and propose two practical applications: Importance Weight Learning and KL divergence estimation.
Our Importance Weight Learning method requires only a single additional prompt and improves F1 scores by up to 7 points.
We further show that CLIP-based density ratios support estimation of KL divergences that quantify how conditioning on an image or text alters the distribution of the other modality. Through qualitative examples and an N-gram analysis of captions, we find that these divergences capture semantic diversity and mode structure in multimodal data. Leveraging this property, we introduce a simple KL-guided data curation method that achieves performance competitive with LAION2B filtering.
Our code is available at \href{https://github.com/fumiyauchiyama/CLIP_Density_Ratio}{https://github.com/fumiyauchiyama/CLIP\_Density\_Ratio}.
\end{abstract}    
\section{Introduction}
\label{sec:intro}


\newcommand{\imgcard}[3]{%
  \begin{adjustbox}{valign=c,minipage={0.165\linewidth}}
    \centering
    \includegraphics[width=\linewidth,height=2.2cm,keepaspectratio]{#1} 
    \\
    {\scriptsize Score: \textcolor{#3}{#2}\par}
  \end{adjustbox}%
}

\begin{figure*}[t]
  \centering
  \begin{subfigure}{\linewidth}
    \centering
    \imgcard{}{10.127}{red!80!black}\hfill
    \imgcard{}{10.127}{red!80!black}\hfill
    \imgcard{}{10.127}{red!80!black}\hfill
    \imgcard{}{10.127}{red!80!black}\hfill
    \imgcard{}{10.127}{red!80!black}\hfill
    \imgcard{}{10.127}{red!80!black}\\[0.5em]
    \imgcard{}{10.127}{red!80!black}\hfill
    \imgcard{}{10.127}{red!80!black}\hfill
    \imgcard{}{10.127}{red!80!black}\hfill
    \imgcard{}{10.127}{red!80!black}\hfill
    \imgcard{}{10.126}{red!80!black}\hfill
    \imgcard{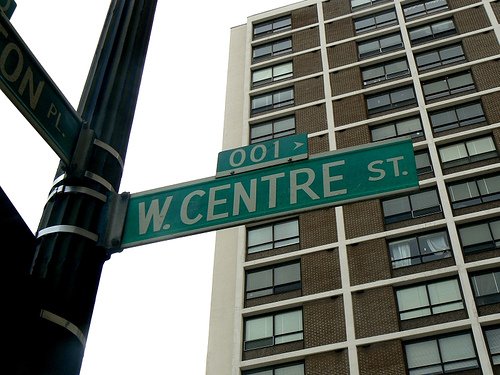}{10.126}{red!80!black}
    \caption{Top 12 images}
  \end{subfigure}
  \hfill
  \vspace{0.5em}
  \begin{subfigure}{\linewidth}
    \centering
    \imgcard{}{4.7777}{blue!80!black}\hfill
    \imgcard{}{4.8449}{blue!80!black}\hfill
    \imgcard{}{4.9000}{blue!80!black}\hfill
    \imgcard{}{4.9305}{blue!80!black}\hfill
    \imgcard{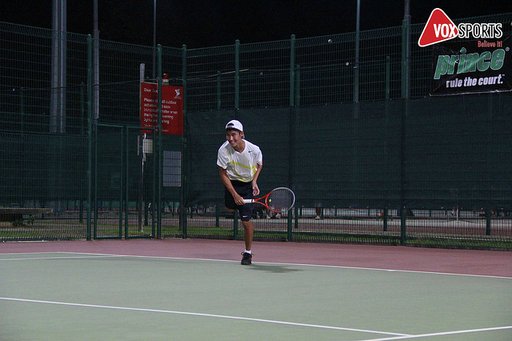}{4.9409}{blue!80!black}\hfill
    \imgcard{}{4.9569}{blue!80!black}\\
    \imgcard{}{4.9732}{blue!80!black}\hfill
    \imgcard{}{5.0248}{blue!80!black}\hfill
    \imgcard{}{5.0277}{blue!80!black}\hfill
    \imgcard{}{5.0465}{blue!80!black}\hfill
    \imgcard{}{5.0507}{blue!80!black}\hfill
    \imgcard{}{5.0534}{blue!80!black}
    \caption{Bottom 12 images}
  \end{subfigure}
  \caption{
  Top and bottom image examples in MSCOCO captions ranked by KL divergence \cref{eq:dkl} through cosine similarity of CLIP.
  While the bottom samples have the same contexts like sports, the top samples have diverse contexts, including non-English contents.
  }
  \label{fig:mscoco_image_samples}
\end{figure*}

A ratio of two probability densities, known as a density ratio, is a fundamental concept in statistical machine learning. Density ratios provide a principled framework for comparing distributions and underlie many foundational techniques, including importance weighting, divergence estimation, and likelihood-free inference. In computer vision, density ratios have been successfully applied to tasks such as image quality assessment~\cite{salimans2016improved} and training generative models~\cite{kingma2013auto,goodfellow2014generative}. Moreover, density ratios are central to domain adaptation and covariate shift correction, enabling models to generalize from one distribution to another by reweighting samples according to their relative likelihood under target and source distributions.

Estimating density ratios directly from sample data is generally more efficient and robust than estimating each probability density separately, which is prone to cumulative errors. Existing methods for direct density ratio estimation include KLIEP~\cite{NIPS2007_be83ab3e}, uLSIF~\cite{JMLR:v10:kanamori09a}, and approaches that formulate density ratio estimation as a logistic regression problem~\cite{10.1145/1273496.1273507,9c632959-7342-323f-94e5-63ab13707d43,d597df41-9436-3b3c-b74d-83c9ecf2f4a8}.
Noise Contrastive Estimation (NCE)~\cite{JMLR:v13:gutmann12a} is a widely used technique for unnormalized statistical models. It estimates density ratios by discriminating data samples from noise, and has been applied to contrastive learning, for example in Word2Vec~\cite{NIPS2013_9aa42b31}. Importantly, NCE demonstrates that contrastive objectives model the logarithm of the density ratio between two distributions. This principle is directly relevant to modern large-scale image–text models, where contrastive learning objectives are used at scale.

Modern large-scale image–text models such as CLIP~\cite{Radford2021LearningTV} and SigLIP~\cite{zhai2023sigmoid} are trained using InfoNCE or NCE, which are contrastive objectives that implicitly model density ratios. In CLIP, InfoNCE optimizes similarity scores between image–text pairs, which correspond to the logarithm of the ratio between the conditional distribution of an image given a text and the marginal distribution of the image, and analogously for text distributions. SigLIP replaces the softmax in CLIP with a sigmoid, and can likewise be interpreted as a variant of NCE that models the density ratio between prior and posterior distributions. Collectively, we refer to these models as CLIP-like models.

Despite their theoretical ability to estimate high-dimensional multimodal density ratios, these models have predominantly been applied to embedding or retrieval tasks. Their full potential as general-purpose density ratio estimators remains largely unexplored, offering a unique opportunity to enable novel algorithmic applications and insights in vision–language learning. If CLIP-like models are interpreted through the lens of density ratios, several practical applications emerge naturally. For instance, importance weighting can guide domain adaptation in image–text pretraining; KL divergences can quantify semantic informativeness; and data curation can be performed by identifying samples that most strongly influence the model’s distribution. These applications suggest that contrastive pretraining encodes a form of probabilistic reasoning, beyond its conventional use for representation learning.

In this work, we systematically explore CLIP-like models as density ratio estimators. First, we provide a unified interpretation of how similarity scores on InfoNCE and NCE encode density ratios in CLIP and SigLIP for both image and text distributions (\cref{sec:method}). Based on this foundation, we demonstrate their effectiveness in two representative applications.

The first is a simple Importance-Weighted Learning (IWL) method~\cite{Sugiyama_Suzuki_Kanamori_2012} for domain-specification of image foundation models. Our method requires only a single prompt and leverages density ratios estimated from a pretrained CLIP-like model. Experiments show that pretraining a new CLIP model using this method improves downstream task performance by up to 7\% in accuracy and F1-score. 

The second application is KL divergence estimation between image and text distributions. These KL divergences quantify how conditioning on one modality shifts the distribution of the other. We propose two estimation strategies: (i) directly using density ratios with sample data, and (ii) approximating KL divergence by interpreting CLIP-like models as exponential family distributions~\cite{oyama-etal-2023-norm}. Empirical analysis on MSCOCO captions (\cref{sec:kl_feature}) demonstrates that captions and images with diverse contextual content have high KL values. Finally, we leverage KL divergences to guide data curation for pretraining, achieving downstream performance competitive with LAION2B filtering.

Our contributions are summarized as follows:
\begin{itemize}
\item \textbf{CLIP-like models as density ratio estimators.}\\ We revisit the interpretation of CLIP-like models as density ratio estimators. We clarify how CLIP and SigLIP, based on InfoNCE and NCE, learn the density ratio of conditional and marginal distributions through similarity scores.\\
\item \textbf{Importance-weighted learning using density ratios.}\\ We leverage density ratios estimated from pretrained CLIP-like models to reweight training samples in a prompt-driven manner, effectively performing domain adaptation. This directly exploits the probabilistic interpretation of similarity scores, and improves downstream accuracy and F1-score with minimal additional supervision.\\ 
\item \textbf{KL-based measurement of semantic diversity.}\\ We formulate KL divergences between image and text distributions using density ratios, revealing that high KL values correspond to samples with diverse contextual content. This connects theoretical density estimation to interpretable properties of multimodal data.\\
\item \textbf{Density ratio-guided data curation.}\\ Using KL-based metrics, we filter pretraining datasets to select the most informative samples. Despite using only pretrained embeddings and simple density ratios, this approach achieves performance competitive with standard large-scale filtering methods, highlighting the practical impact of our framework.
\end{itemize}
\section{Related Work}
\label{sec:relatedwork}

\subsection{Density Ratio Estimation and Its Applications}
Classical approaches to density ratio estimation include KLIEP~\cite{NIPS2007_be83ab3e}, uLSIF~\cite{JMLR:v10:kanamori09a}, and LogReg~\cite{10.1145/1273496.1273507}. These methods directly estimate the ratio between two distributions from samples, without explicitly modeling each density. 
The idea of LogReg is similar to that of Noise Contrastive Estimation~\cite{JMLR:v13:gutmann12a}, which discriminates data from noise to model the density ratio.
These methods are widely used for importance weighting~\cite{iwl_apps_1,iwl_apps_2,iwl_apps_3}, where the ratio between training and test distributions is required to construct an unbiased loss under distribution shift. Also, estimation of the KL divergence and mutual information is another application of the density ratio estimator since the KL divergence has the log-ratio of two distributions within its definition~\cite{mi_from_densratio,mi_apps_1,mi_apps_2,mi_apps_3}.

While these methods provide principled formulations, they typically require a tailored estimator for each pair of distributions.
In contrast, CLIP-like models, so-called ``vision-language foundation models'', are trained on vast and diverse image–text pairs, and their learning algorithms encode rich relationships between marginal and conditional distributions of images and texts to a scalar. This raises a largely unexplored question: to what extent can CLIP-like models be reused as off-the-shelf, general-purpose density ratio estimators for downstream tasks, without additional training?
Despite their growing use as universal feature extractors~\cite{Radford2021LearningTV,zhai2023sigmoid,li2023blip,llava}, this potential has not been systematically investigated for applications such as domain adaptation or distributional analysis.

\subsection{Density Ratio Relationship in Contrastive Learning}
Noise Contrastive Estimation (NCE) and its variant InfoNCE are central to contrastive learning in both word embeddings~\cite{NIPS2013_9aa42b31} and CLIP-like models. In the context of word embeddings, skip-gram with negative sampling is known to approximate point-wise mutual information $\log\tfrac{p(x|y)}{p(x)}$~\cite{church-hanks-1990-word}, which corresponds to the density ratio that CLIP-like models encode~\cite{NIPS2014_b7866697}. More recent work interprets word contrastive objectives as exponential-family models and shows that the 2-form of the embeddings approximates the KL divergence between the marginal distribution and the conditional distribution given context words~\cite{oyama-etal-2023-norm}. 

SigLIP~\cite{zhai2023sigmoid} replaces the softmax loss of CLIP with the sigmoid function, aligning the learning dynamics with logistic regression between marginal and conditional distributions. Prior analyses, however, are largely restricted to the natural language processing domain or theoretical characterizations. To the best of our knowledge, no existing work interprets vision–language models such as CLIP and SigLIP as high-dimensional density-ratio estimators, nor leverages this interpretation for downstream applications such as domain adaptation or KL divergence estimation.



\subsubsection{Data Curation for Large Multi-modal Pre-training}
High-quality data curation is essential for efficient pretraining of large-scale image–text models. Recent benchmarks such as DataComp~\cite{datacomp} evaluate filtering strategies based on alignment scores, heuristic rules, or quality classifiers. Metrics like CLIPScore~\cite{hessel-etal-2021-clipscore} assess image–text consistency and effectively identify misaligned or noisy samples.

Existing filtering methods focus primarily on alignment, without quantifying how individual samples influence the underlying distributions. In contrast, our KL divergence–based approach measures the distributional impact of each sample via the density ratio interpretation of CLIP-like models. This provides a complementary axis to alignment-based filtering, enabling more efficient utilization of large multi-modal datasets and highlighting samples that exert substantial influence on distributional structure.

\section{CLIP-like models as Density Ratio Estimators}
\label{sec:method}

In this section, we formalize how CLIP-like models trained with NCE or InfoNCE implicitly estimate the density ratio between the conditional and marginal distributions of text given an image, and symmetrically for the image given a text. This interpretation allows us to view the similarity scores learned by these models as principled density ratio estimators for high-dimensional multimodal data.

Let $p_T(t)$ denote the marginal probability of a text $t$, and $p_T(\cdot)$ the overall text distribution. The conditional probability of a text $t$ given an image $i$ is denoted as $p_{T}(t|i)$ and the conditional text distribution given an image $i$ is denoted as $p_{T}(\cdot|i)$. CLIP and SigLIP learn embeddings $v_i$ and $v_t$ satisfying
\begin{equation}
  \label{eq:densratio}
  \frac{p_{T}(t|i)}{p_{T}(t)} = \frac{\exp(a \langle v_t, v_i \rangle)}{Z(i)},
\end{equation}
where $a$ is a scaling factor called logit scale, and $Z(i):=\mathbb{E}_{t\sim p_T(\cdot)}[\exp(a \langle v_t, v_i \rangle)]$ is a normalizing term depending on the image $i$.
\Cref{eq:densratio} shows that the logarithm of the density ratio between the conditional text distribution given an image and the text distribution is proportional to the inner product of the embeddings. This provides a direct connection between the learned similarity scores and the underlying density ratio, allowing CLIP-like models to be interpreted as implicit density ratio estimators. A similar analysis has been explored in the context of word embeddings~\cite[Eq.~(8)]{NIPS2014_b7866697}.

Analogously, for the image distribution $p_I(\cdot)$ and conditional image distribution given a text $p_I(\cdot|t)$, we have
\begin{equation}
  \frac{p_{I}(i|t)}{p_{I}(i)} = \frac{\exp(a \langle v_i, v_t \rangle)}{Z(t)}.
  \label{eq:densratio_txt}
\end{equation}
demonstrating the symmetric nature of density ratio estimation across modalities.
In the supplemental materials, we provide a detailed derivation for why this formulation holds for both CLIP and SigLIP, organizing arguments from \cite{JMLR:v13:gutmann12a,oord2018representation} in the context of these modern image–text models.

\section{Application: A Simple Importance-Weighting Learning}
\label{sec:apps}
Importance-Weighted Learning (IWL) \cite{Sugiyama_Suzuki_Kanamori_2012} addresses domain adaptation under covariate shift~\cite{covshift}, where the test input distribution differs from training, but the conditional distribution of outputs given inputs remains the same.

Formally, let $y$ denote the output of any modality and $x$ an image input. A distributional change between the training data and the test data is called a covariate shift when
\begin{align}
    p^{\mathrm{train}}(\cdot|x) &= p^{\mathrm{test}}(\cdot|x) \\
    p_I^{\mathrm{train}}(x) &\neq p_I^{\mathrm{test}}(x)
\end{align}
where $p_I^{\mathrm{train}}(x)$ and $p_I^{\mathrm{test}}(x)$ are the distribution of an input image, and $p^{\mathrm{train}}(\cdot|x)$ and $p^{\mathrm{test}}(\cdot|x)$ are the conditional output distribution given an image $x$.

Denoting a loss function calculated from an output predicted from $x$ and its ground-truth by $l(x)$, we want to reduce the test loss $L_\mathrm{test}:=\mathbb{E}_{x\sim p_I^\mathrm{test}(\cdot)}[l(x)]$, rather than train loss $L_\mathrm{train}:=\mathbb{E}_{x\sim p_I^\mathrm{train}(\cdot)}[l(x)]$.
With some modifications, $L_\mathrm{test}$ can be estimated by the training distribution and density ratio of inputs between $p_I^\mathrm{train}(\cdot)$ and $p_I^\mathrm{test}(\cdot)$ that
\begin{align}
    L_\mathrm{test}
    &= \int l(x)p_I^\mathrm{test}(x)dx \notag\\
    &= \int l(x)\frac{p_I^\mathrm{test}(x)}{p_I^\mathrm{train}(x)}p_I^\mathrm{train}(x)dx \notag\\
    &= \mathbb{E}_{x\sim p_I^\mathrm{train}(\cdot)}\left[\frac{p_I^\mathrm{test}(x)}{p_I^\mathrm{train}(x)}l(x)\right].
\end{align}
This formulation allows unbiased evaluation of the test loss using only training data and the density ratio, without requiring test labels.
Traditionally, $r(x)$ is estimated by training a separate density ratio estimator. In our framework, we leverage CLIP-like models to estimate this ratio using \cref{eq:densratio_txt}:
\begin{equation}
    \frac{p_I^\mathrm{test}(x)}{p_I^\mathrm{train}(x)} \propto \exp(a\langle v_x, v_t \rangle)
\end{equation}
assuming the test distribution corresponds approximately to a conditional distribution given a prompt $t$, \ie $p_I^\mathrm{test}(\cdot) \approx p_I(\cdot|t)$. The normalizing term $Z(t)$ in \cref{eq:densratio_txt} is constant across samples and can be ignored for weighting purposes.

We evaluated this approach by re-weighting pre-training for CLIP on domain-specific datasets. Three prompts were used: “A photo of food”, “A photo of pets”, and “A photo of flowers”. The re-weighted CLIP pretraining loss is defined as:

\begin{align}
  L'_{\mathrm{CLIP}}
  &= 
    -\sum_{j=1}^N
    \underbrace{
    e^{a\langle u_{i_j}, u_{t^\dagger}\rangle}
    \log\!
    \left[
      \frac{
        \exp(s(t_j, i_j))
      }{
        \sum_{k=1}^{N}
        \exp(s(t_k, i_j))
      }
    \right]
    }_{\text{image-to-text prediction}}
    \notag
    \\
  &\quad-\sum_{j=1}^N
    \underbrace{
    e^{a\langle u_{i_j}, u_{t^\dagger}\rangle}
    \log\!
    \left[
      \frac{
        \exp(s(t_j, i_j))
      }{
        \sum_{k=1}^{N}
        \exp(s(t_j, i_k))
      }
    \right]
    }_{\text{text-to-image prediction}}
  \label{eq:iwl_clip}
\end{align}
where $u$ is the embedding of the other pre-trained CLIP and $t^\dagger$ is the prompt that describes the domain of the test data. Optimizing \cref{eq:iwl_clip} is proportionally equal to minimizing the loss on the desired domain defined by the prompt.
For simplicity, the same weighting is applied to both image-to-text and text-to-image terms. We used the embedding of CLIP (ViT-L-14) trained on the LAION dataset \cite{ilharco_gabriel_2021_5143773} as $u$, and scaled down the logit scale $a$ from almost 100 to 10, because large $a$ will increase loss values exponentially, and lead to overflow on mixed-precision training. Pre-training was done by the codebase of OpenCLIP\cite{ilharco_gabriel_2021_5143773} with a few modifications, and CC12M\cite{changpinyo2021cc12m} is used as the pre-training data. For details of pre-training, please refer to the supplemental materials.
As the baseline, we also pre-trained a model whose loss is not weighted by importance.

\begin{figure*}[t]
  \centering
  \begin{subfigure}{0.33\linewidth}
    \includegraphics[width=\linewidth]{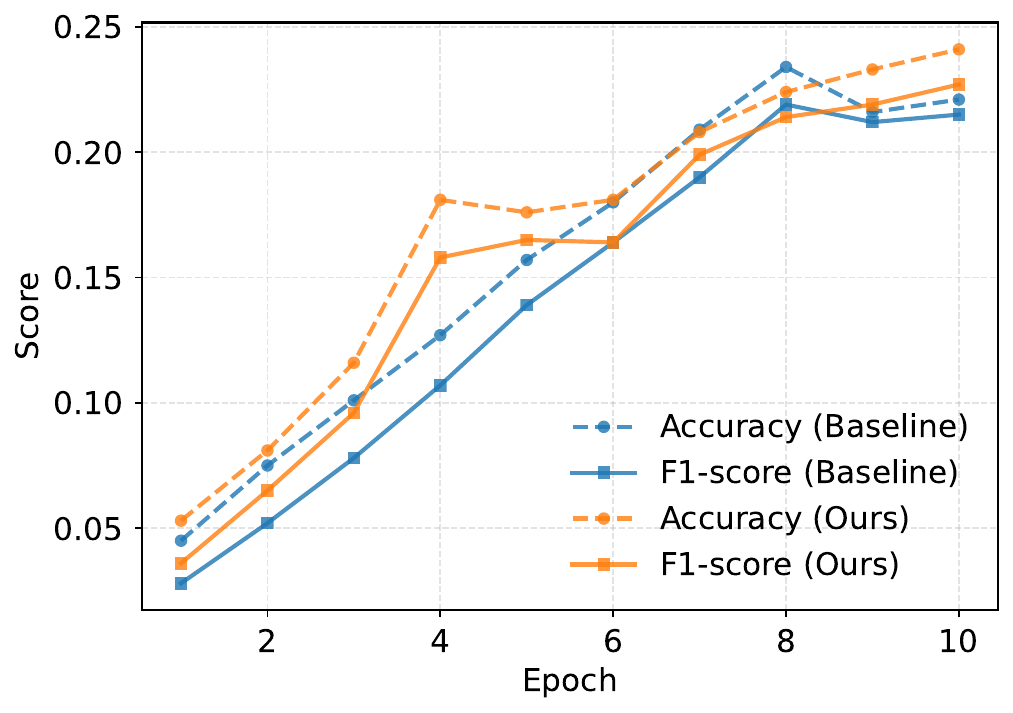}
    \caption{Food101}
  \end{subfigure}
  \hfill
  \begin{subfigure}{0.33\linewidth}
    \includegraphics[width=\linewidth]{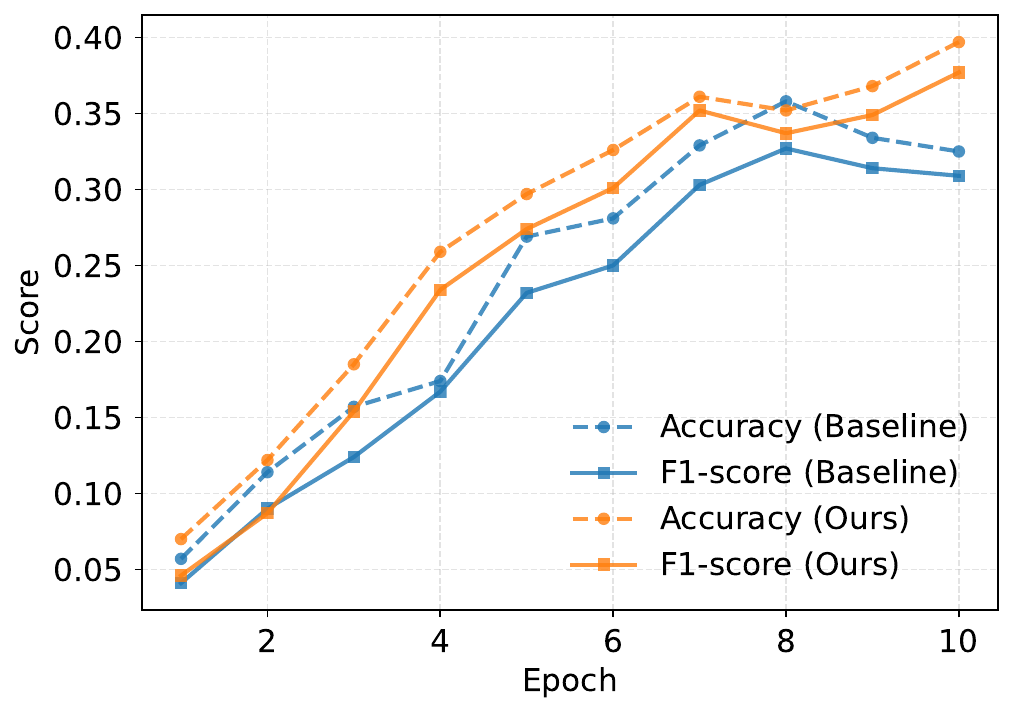}
    \caption{Oxford-IIIT Pet Dataset}
  \end{subfigure}
  \hfill
  \begin{subfigure}{0.33\linewidth}
    \includegraphics[width=\linewidth]{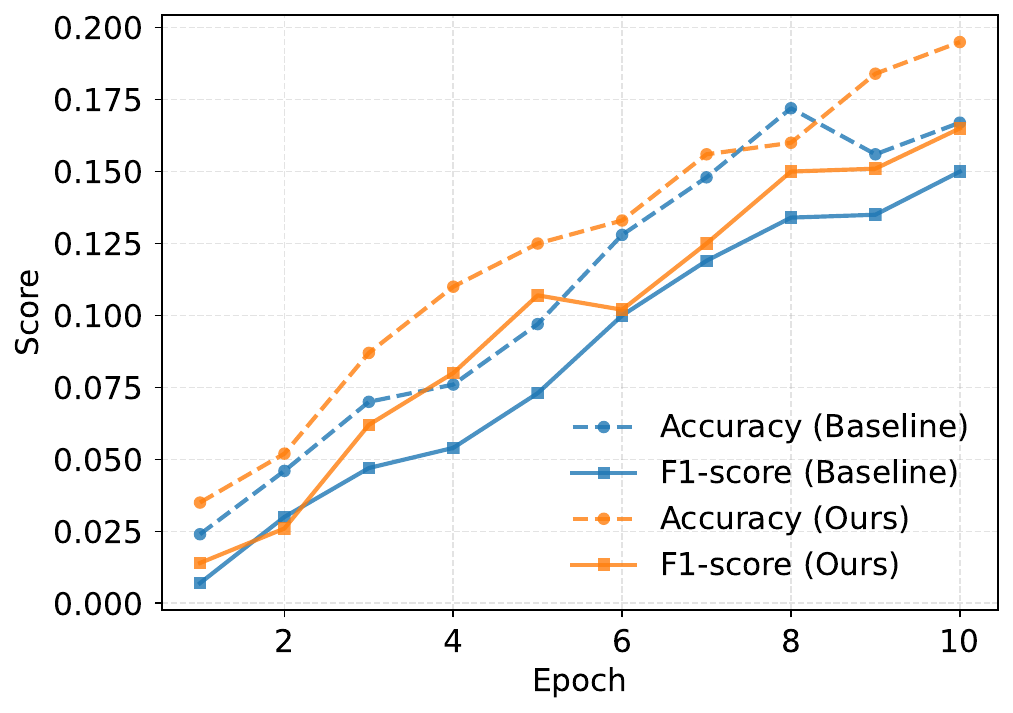}
    \caption{Flowers102}
  \end{subfigure}
  \caption{Zero-shot classification performance comparison between baseline and our IWL method across three downstream datasets.}
  \label{fig:iwl_performance}
\end{figure*}

\Cref{fig:iwl_performance} shows the performance on zero-shot classification task of Food101~\cite{bossard14}, The Oxford-IIIT Pet Dataset~\cite{parkhi12a} and Flowers102~\cite{Nilsback08} on CLIP benchmark~\cite{cherti_2025_15403103}. Across accuracy and F1-score, our IWL method consistently outperforms the baseline that uses the standard CLIP loss.
In particular, our method achieves faster performance gains, surpassing the baseline by 2–8 percentage points in accuracy and 3–7 percentage points in F1-score on the Oxford-IIIT Pet Dataset at the 4–6 epoch checkpoints.

Intuitively, IWL re-weights pretraining samples according to their relevance to the target domain. Samples whose images are more compatible with the domain prompt receive larger weights, while samples that are not related to the target domain receive smaller weights. 
This soft selection differs from hard filtering strategies that drop all low-scoring samples, and can be advantageous when the proxy metrics are imperfect or when the domain is only loosely specified by a single prompt. 
Though the logit scale remains relatively small due to the precision of the numerical format, a larger logit scale, such as 100, which is closer to the logit scale that CLIP-like models have, would raise further improvements.

Our experiments suggest importance weighting with CLIP-based density ratios is a simple yet practical alternative to designing dedicated domain-specific datasets from scratch with high costs of correcting and annotations.
\section{Application: KL Divergence}
\label{sec:apps_kl}
In this section, we explore how conditioning an image or a text affects the distributions of the other modality, measured via KL divergences. This concept is known as Information Gain, and these variants are used in keyword extraction and image-quality assessment \cite{oyama-etal-2023-norm,salimans2016improved}. First, we derive how to estimate these KL divergences with \cref{eq:densratio,eq:densratio_txt} and sample data in \cref{sec:apps_kl_intro}, then analyze their properties (\cref{sec:kl_feature}) and apply them to data curation (\cref{sec:apps_kl_datacomp}).

\subsection{Estimating KL divergences using CLIP-like models}
\label{sec:apps_kl_intro}
Using the density ratio estimated by CLIP-like models explained in \cref{eq:densratio,eq:densratio_txt}, we can estimate the following two KL divergences.
\begin{align}
  D_\mathrm{KL}(i) 
  &:= \mathrm{KL}(p_{T}(\cdot|i) \| p_{T}(\cdot)) 
  \label{eq:dkl}
  \\
  D_\mathrm{KLR}(i) 
  &:= \mathrm{KL} (p_{T}(\cdot) \| p_{T}(\cdot|i))
  \label{eq:dklr}
\end{align}
Especially, $D_\mathrm{KL}(i)$ is called Information Gain~\cite{oyama-etal-2023-norm}.
\Cref{eq:dkl,eq:dklr} are approximated by \cref{eq:densratio} and text samples $\mathcal{D}_T$
\begin{align}
  D_\mathrm{KL}(i) 
  &\approx
  \sum_{t\in\mathcal{D}_T}
  a\langle v_{t},v_i\rangle 
  \mathrm{softmax}_i(t)
  \notag
  \\
  &\quad
  -\log
  \sum_{t\in\mathcal{D}_T}
  a\langle v_{t},v_i\rangle 
  e^{a\langle v_{t},v_i\rangle}
  +\log |\mathcal{D}_T|
  \label{eq:dkl_approx}
  \\
  D_\mathrm{KLR}(i) 
  &\approx
  \log
  \sum_{t\in\mathcal{D}_T}
  e^{a\langle v_{t},v_i\rangle}
  -\log |\mathcal{D}_T|
  \notag
  \\
  &\quad- \frac{1}{|\mathcal{D}_T|}
  \sum_{t\in\mathcal{D}_T}
  a\langle v_t,v_i\rangle.
  \label{eq:dklr_approx}
\end{align}
For detailed derivations, please refer to the supplemental materials.
By symmetry of the learning objective of CLIP-like models, the derived approximation holds also for the other modality, \ie, $D_\mathrm{KL}(t):= \mathrm{KL}(p_{I}(\cdot|t) \| p_{I}(\cdot)) $ and $D_\mathrm{KLR}(t):= \mathrm{KL} (p_{I}(\cdot) \| p_{I}(\cdot|t)) $ can be estimated with \cref{eq:densratio_txt} and image samples $\mathcal{D}_I$ by reversing modalities:
\begin{align}
  D_\mathrm{KL}(t) 
  &\approx
  \sum_{i\in\mathcal{D}_I}
  a\langle v_{t},v_i\rangle 
  \mathrm{softmax}_i(i)
  \notag
  \\
  &\quad
  -\log
  \sum_{i\in\mathcal{D}_I}
  a\langle v_{t},v_i\rangle 
  e^{a\langle v_{t},v_i\rangle}
  +\log |\mathcal{D}_I|
  \label{eq:txt_dkl_approx}
  \\
  D_\mathrm{KLR}(t) 
  &\approx
  \log
  \sum_{i\in\mathcal{D}_I}
  e^{a\langle v_{t},v_i\rangle}
  -\log |\mathcal{D}_I|
  \notag
  \\
  &\quad- \frac{1}{|\mathcal{D}_I|}
  \sum_{i\in\mathcal{D}_I}
  a\langle v_t,v_i\rangle.
  \label{eq:txt_dklr_approx}
\end{align}

\citet{oyama-etal-2023-norm} introduced approximation of \cref{eq:dkl} in terms of word statistics via the embedding of word2vec (skip-gram with negative sampling~\cite{NIPS2013_9aa42b31}) trained on NCE.

\Cref{eq:densratio} is transformed as
\begin{equation}
  p_{T}(t|i) = \frac{p_{T}(t)\exp(\langle a v_t, v_i \rangle)}{\mathbb{E}_{t\sim p_T(\cdot)}[e^{a \langle v_t, v_i \rangle}]}.
\end{equation}
For fixed $i$ and $v_i$, this can be regarded as an instance of exponential families~\cite{10.1214/aos/1176345779,Efron_2022}.
\begin{equation}
  p_{T}(t|v_i) = p_{T}(t)\exp(a \langle v_t, v_i \rangle-\psi(v_i))
\end{equation}
where $a v_t$ is sufficient statistics, $v_i$ is a natural parameter, and $\psi(v_i):=\log\mathbb{E}_{t\sim p_T(\cdot)}[\exp(a \langle v_t, v_i \rangle)]$ is the normalization function that makes $\int p_{T}(t|v_i)dv_t$ equal to one. As a well-known property of exponential families, the KL divergence between two exponential families with different natural parameters is approximated as follows:
\begin{align}
  &\mathrm{KL}(p_{T}(\cdot|v_1)\|p_{T}(\cdot|v_2)) \notag\\
  &= a^2(v_1-v_2)^\top G(v_i) (v_1-v_2) + o(\|v_1-v_2\|^2).
\end{align}
Assuming that there exists $v_0$ representing the marginal distribution $p_{T}(\cdot)=p_{T}(\cdot|v_0)$, \citet{oyama-etal-2023-norm} showed that $\bar{v}_I:=\mathbb{E}_{i\sim p_I(\cdot)}[v_i]$ approximates $v_0$ and 
\begin{align}
  D_\mathrm{KL}(i) &\approx a^2(v_i-\bar{v}_I)^\top G_T (v_i-\bar{v}_I)\label{eq:oyama}\\
  G_T&:=\int(v_t-\bar{v}_T)(v_t-\bar{v}_T)^\top p_T(t)dv_t \notag\\
  \bar{v}_T&:=\mathbb{E}_{t\sim p_T(\cdot)}[v_t] \notag
\end{align}
ignoring higher orders. Note that we rewrite the theoretical result of \cite{oyama-etal-2023-norm} for continuous expression. In this work, we treat the approximation \cref{eq:oyama} and its variant:
\begin{align}
    D_\mathrm{W}(i)&:=a^2(v_i-\hat{v}_I)^\top \hat{G}_T (v_i-\hat{v}_I) \label{eq:dw_approx_img}\\
    D_\mathrm{C}(i)&:=a^2\|v_i-\hat{v}_I\|^2 \label{eq:dc_approx_img}
\end{align}
where $\hat{v}_T:=\frac{1}{|\mathcal{D}_T|}\sum_{t\in\mathcal{D}_T}v_t$ and $\hat{v}_I:=\frac{1}{|\mathcal{D}_I|}\sum_{i\in\mathcal{D}_I}v_i$ are the average vector of text and image embeddings
, and $\hat{G}_T:=\frac{1}{|\mathcal{D}_T|}\sum_{t\in\mathcal{D}_T}(v_t-\hat{v}_T)(v_t-\hat{v}_T)^\top$ and $\hat{G}_I:=\frac{1}{|\mathcal{D}_I|}\sum_{i\in\mathcal{D}_I}(v_i-\hat{v}_I)(v_i-\hat{v}_I)^\top$ are the covariance matrix of the embeddings. $D_\mathrm{W}(t)$ and $D_\mathrm{C}(t)$ are also defined by reversing modalities. These are the squared (metric-induced) norm of centered embeddings.

\subsection{Comparison to metrics in related work.}
\textbf{Conformity~\cite{reb1}.}
For a fixed mean vector $\hat{v}$, $D_\mathrm{C}$ can be decomposed by $a^2(||v_i||^2 + ||\hat{v}||^2 - 2\langle v_i, \hat{v}\rangle)=-2a^2 \cos(v_i, \hat{v}) + Const.$ ($\because v_i$ are normalized). On the other hand, Conformity $\mathbb{E}_{j, j\neq i}[\langle \cos(v_i, v_j) \rangle]$ is also introduced in a previous study to formalize ``common'' themes in CLIP embeddings, measuring discrepancy of an embedding $v_i$ from the mean vector~\cite{reb1}.
Although we introduce $D_\mathrm{C}$ from the interpretation of CLIP-like models as instances of exponential family and the background is different from~\cite{reb1}, conformity is equivalent to the minus of $D_\mathrm{C}$ by ignoring constants. In fact, \cref{fig:mscoco_image_samples_dc} contains the same image as Fig.8 in~\cite{reb1}.

\textbf{Log-likelihood of whiten CLIP~\cite{reb2} and Raw norm~\cite{reb3}.}
Both encode frequency and have a negative correlation to $D_\mathrm{C}$. 
$D_\mathrm{C}$ shows a stronger correlation with the log-likelihood (Tab. \ref{tab:comparison}).
$D_\mathrm{C}$ measures the centered norm of the embedding. This means that a frequent image has a small $D_\mathrm{C}$, while a rare image has a large $D_\mathrm{C}$. This negative correlation naturally demonstrates the inverse relationship between $D_\mathrm{C}$ and frequency (log-likelihood). Raw norm, which grows more when an image is frequent, also shows a negative correlation with $D_\mathrm{C}$.
On the other hand, $D_\mathrm{KL}$ and $D_\mathrm{KLR}$ show little correlation with these metrics, suggesting they capture different information from the others.

\begin{table}[t]
    \centering
    \begin{tabular}{c|cccc}
        Metric & $D_{\mathrm{KL}}$ & $D_{\mathrm{KLR}}$ & $D_\mathrm{C}$ & $D_\mathrm{W}$ \\
        \hline
        Conformity~\cite{reb1} & -0.255 & 0.015 & \textbf{-1.000} & 0.093 \\
        $\log p(x)$~\cite{reb2} & -0.346 & 0.096 & \textbf{-0.626} & 0.389 \\
        Raw Norm~\cite{reb3} & -0.089 & -0.046 & \textbf{-0.731} & -0.120 \\
    \end{tabular}
    \caption{Pearson correlation coefficient of related metrics to ours. Each metric is for images.}
    \label{tab:comparison}
\end{table}

\subsection{KL Divergence represents semantic diversity}
\label{sec:kl_feature}
\newcommand{\textcard}[3]{%
  \begin{minipage}[c]{0.16\linewidth}
    \raggedright
    {\scriptsize #1\par}
    {\scriptsize Score: \textcolor{#3}{#2}\par}
  \end{minipage}
}
\begin{figure*}[t]
  \centering
  \begin{subfigure}{\linewidth}
    \centering
    \textcard{A Beanie Baby beside a vintage photo of a man and a woman.}{8.5172}{red!80!black}\hfill
    \textcard{Sign of Arch Street Fire Department saying "He who laughs last thinks slowest"}{8.5172}{red!80!black}\hfill
    \textcard{A girl in a blue bowtie standing next to a pink Princess Parking Only sign.}{8.5172}{red!80!black}\hfill
    \textcard{a man in a white lab coat holding a hacksaw}{8.5172}{red!80!black}\hfill
    \textcard{A smiling girl standing beside a sign that says "Princess Parking Only".}{8.5172}{red!80!black}\hfill
    \textcard{An iguana and two birds are sitting on several steps.}{8.5172}{red!80!black}\\[0.5em]
    \textcard{A street sign reads "Bacchanalia" above a stop sign.}{8.5172}{red!80!black}\hfill
    \textcard{A sign that says "Abbey Road NW8 City of Westminster".}{8.5172}{red!80!black}\hfill
    \textcard{Othello sign in front of a restaurant on a street.}{8.5172}{red!80!black}\hfill
    \textcard{This green street name sign says bodacious drive}{8.5172}{red!80!black}\hfill
    \textcard{A green street sign that says Bodacious Drive.}{8.5172}{red!80!black}\hfill
    \textcard{a clean street sign that reads bodacious dr}{8.5172}{red!80!black}
    \caption{Top 12 texts}
  \end{subfigure}
  
  \vspace{1em}
  
  \begin{subfigure}{\linewidth}
    \centering
    \textcard{There is no image to describe for this question.}{1.7408}{blue!80!black}\hfill
    \textcard{There is no image to be reviewed on this hit.}{1.7670}{blue!80!black}\hfill
    \textcard{There is no image here to provide a caption for.}{1.7773}{blue!80!black}\hfill
    \textcard{There is no image here to provide a caption for.}{1.7773}{blue!80!black}\hfill
    \textcard{There is no image here to provide a caption for.}{1.7773}{blue!80!black}\hfill
    \textcard{There is no image here to provide a caption for.}{1.7773132}{blue!80!black}\\[0.5em]
    \textcard{"There are some appealing support ready to expend. "}{1.8377}{blue!80!black}\hfill
    \textcard{unable to see this image in this particular hit}{1.8673}{blue!80!black}\hfill
    \textcard{I am not sure what this image is.}{1.8947}{blue!80!black}\hfill
    \textcard{no image again there are a lot of these lately}{1.9048}{blue!80!black}\hfill
    \textcard{I do not know what this is supposed to be..}{2.3317}{blue!80!black}\hfill
    \textcard{I am unable to see the image above.}{2.4624}{blue!80!black}
    \caption{Bottom 12 texts}
  \end{subfigure}
  
  \caption{Top and bottom captions ranked by $D_\mathrm{KL}$.}
  \label{fig:mscoco_text_samples}
\end{figure*}

\Cref{fig:mscoco_image_samples,fig:mscoco_text_samples} qualitatively demonstrate the image and text samples that have top or bottom $D_\mathrm{KL}(i)$ and $D_\mathrm{KL}(t)$ values in MSCOCO captions\cite{lin2014microsoft}. The metric is calculated by the score of CLIP ViT-L-14 trained on LAION2B, same as in \cref{sec:apps}. While the samples with the lowest $D_\mathrm{KL}$ depict common or meaningless concepts, the samples with the highest $D_\mathrm{KL}$ have diverse contexts. 

The samples in~\cref{fig:mscoco_image_samples,fig:mscoco_text_samples} and the definition of our KL divergences (\cref{eq:dkl,eq:dklr}) induce the hypothesis that $D_\mathrm{KL}$ are related to the semantic diversity. To confirm the semantic diversity of higher $D_\mathrm{KL}$ samples quantitatively, we measured N-gram probability coverage by top-K N-grams in the captions. 
First, we grouped captions by deciles of $D_\mathrm{KL}$. To investigate the relationship between $D_\mathrm{KL}$ of images and the semantic diversity of them, captions are classified into 10 groups by $D_\mathrm{KL}$ of the corresponding image. Note that there are five captions per image in MSCOCO captions. Similarly, to investigate the relationship between  $D_\mathrm{KL}$ of captions and their semantic diversity, each caption is also classified into another ten groups by $D_\mathrm{KL}$ of itself.

\begin{figure*}[t]
  \centering
  \begin{subfigure}{1\linewidth}
    \includegraphics[width=\linewidth]{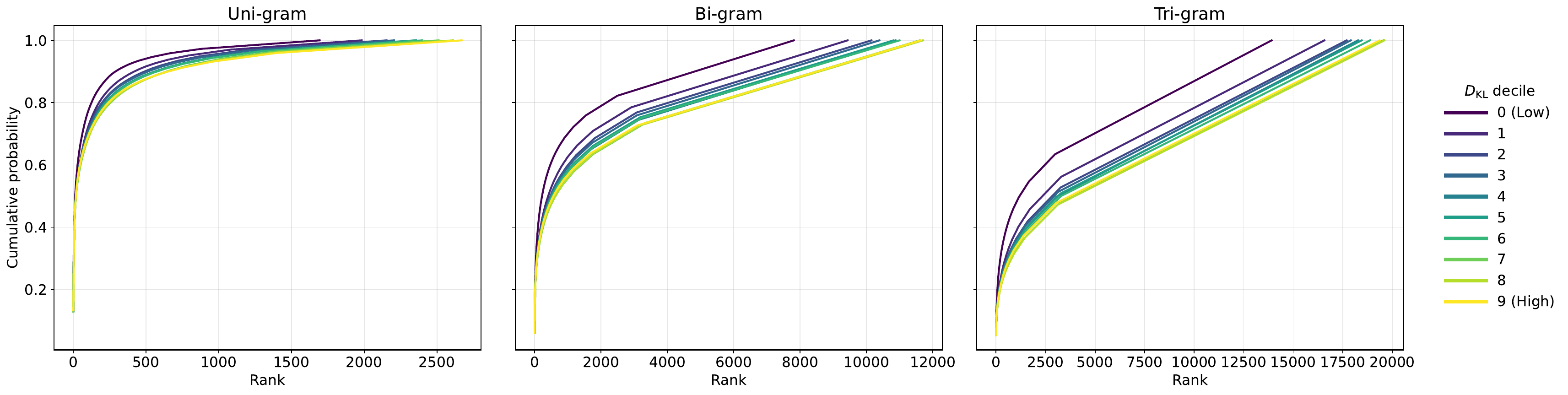}
    \caption{Cumulative N-gram probability across $D_\mathrm{KL}(i)$ deciles}
    \label{fig:mscoco_img}
  \end{subfigure}
  
  \vspace{1em}
  
  \begin{subfigure}{1\linewidth}
    \includegraphics[width=\linewidth]{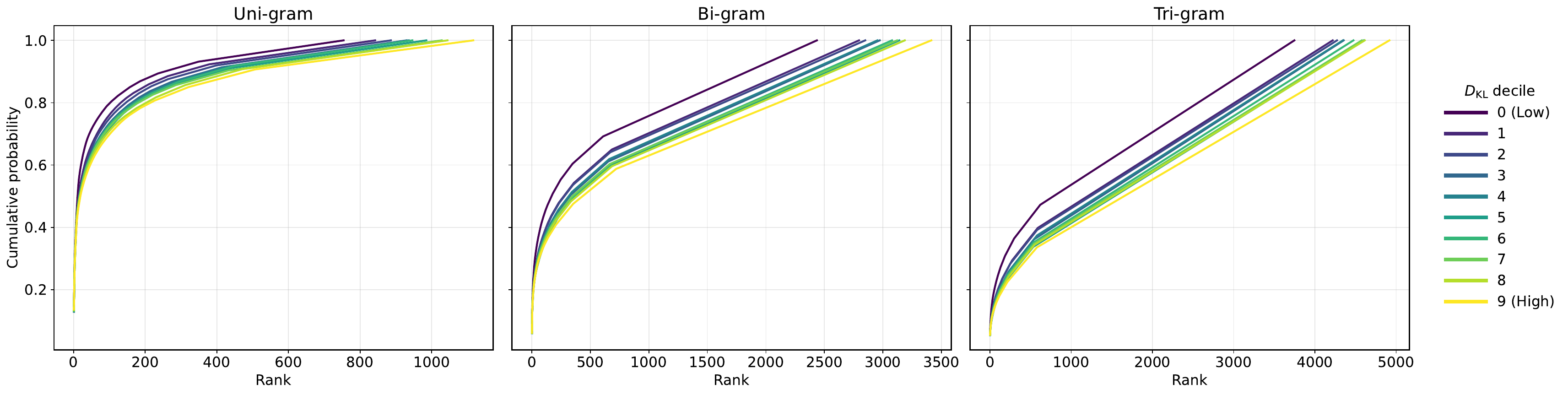}
    \caption{Cumulative N-gram probability across $D_\mathrm{KL}(t)$ deciles}
    \label{fig:mscoco_txt}
  \end{subfigure}
  \caption{
    N-gram probability coverage across $D_\mathrm{KL}$ deciles. 
    Each decile in \cref{fig:mscoco_img} is a group of captions corresponding to images $i$ which has the same level of $D_\mathrm{KL}(i)$.
    Each decile in \cref{fig:mscoco_txt} is a group of captions $t$ which has the same level of $D_\mathrm{KL}(t)$.
    }
  \label{fig:mscoco_ngram}
\end{figure*}

\Cref{fig:mscoco_ngram} shows cumulative N-gram probability by top-K N-grams (\ie what percentage of the whole text in a group of captions does top-K N-grams occupy). Both captions grouped by the corresponding image's $D_\mathrm{KL}$ and captions grouped by $D_\mathrm{KL}$ of themselves, the higher the KL divergence, the lower the cumulative probability curve is. For instance, while the top 2500 tri-grams cover 60\% of all tri-gram occurrences in the lowest image KL group, the top 2500 tri-grams of other caption groups cover almost half or fewer occurrences in each group of captions (\cref{fig:mscoco_img}). This means that the higher $D_\mathrm{KL}$ sample is a caption (or has captions) that contains diverse N-grams compared to the lower samples. These results are evidence that $D_\mathrm{KL}$ represents the semantic diversity of an image and a text. This is intuitive because conditioning a sample that has rare semantics strongly distorts the distribution of another modal. The results of KL divergences other than $D_\mathrm{KL}$ are available in the supplemental materials.

\subsection{Data Curation}
\label{sec:apps_kl_datacomp}

We applied KL divergences introduced in \cref{sec:apps_kl_intro} to curate pre-training data of CLIP. DataComp~\cite{datacomp} is a benchmark that evaluates filtering methods by filtering samples from a prepared data pool, pre-training CLIP on the filtered subset, and evaluating on diverse downstream tasks.

The data pool to filter is a large dataset of image-text pairs. To classify whether a pair $(t,i)$ should be removed or maintained, we calculated either $D_\mathrm{KL}(t)$ or $D_\mathrm{KL}(i)$, and maintained pairs possessing the top 25\% $D_\mathrm{KL}$ values. This approach is based on our hypothesis that the sample that has a specific context should be learned for aligning the image and text. We conducted data-curation and pretraining at the ``small'' scale on DataComp, then evaluated on 38 image classification and retrieval tasks. We also conducted curation by $D_\mathrm{KLR}$, $D_\mathrm{W}$ and $D_\mathrm{C}$. We sampled 10k images and texts in the data pool of DataComp for $\mathcal{D}_T$ and $\mathcal{D}_I$, and utilized CLIP embedding already prepared in the data pool to calculate KL divergences.

\begin{table}[t]
\centering
\begin{tabular}{lccc}
\hline
Metrics & Modal & IN1k Acc. & Mean metrics \\
\hline
No filtering & N/A & 0.025 & 0.132 \\
LAION2B & N/A & 0.031 & 0.133 \\
Basic & N/A & 0.03 & 0.142 \\
CLIPScore & N/A & 0.051 & 0.173 \\
\hline
$D_\mathrm{KL}$ & Text & 0.0300 & 0.1337 \\
   & Image & 0.0216 & 0.122 \\
$D_\mathrm{KLR}$ & Text & 0.0325 & 0.1344 \\
      & Image & 0.0176 & 0.1160 \\
$D_\mathrm{W}$ & Text & 0.0215 & 0.1246 \\
           & Image & 0.0144 & 0.1104 \\
$D_\mathrm{C}$ & Text & 0.0312 & 0.1319 \\
           & Image & 0.0108 & 0.0994 \\
\hline
\end{tabular}
\caption{
Evaluation results for different preprocessing methods. ``Modal'' describes which modal to calculate the KL divergence for our methods, ``IN1k'' stands for ImageNet 1k, and ``Mean metrics'' for the average performance over 38 datasets in DataComp. 
The upper rows other than $D_\mathrm{KL},D_\mathrm{KLR},D_\mathrm{W},D_\mathrm{C}$ are cited from the result of DataComp~\cite{datacomp}.
}
\label{tab:datacomp}
\end{table}


\Cref{tab:datacomp} demonstrates zero-shot classification accuracy on ImageNet1k, and average performance on 38 downstream tasks including classification and retrieval tasks. Compared with no-filtering data, the pretraining data filtered by $D_\mathrm{KL}, D_\mathrm{KLR}$, and $D_\mathrm{C}$ of text achieve 5–8 percentage points higher accuracy on the ImageNet1k zero-shot classification task and almost the same average performance despite the size of the data being one quarter. KL divergences of text are more effective than those of images, indicating that measuring text informativeness directly matches to make alignment of image and text data. CLIPScore remains a very strong baseline for data filtering, and our KL-based selection alone does not surpass CLIPScore. However, our KL-based metrics and CLIPScore target different aspects of the data: CLIPScore primarily measures alignment between an image and its caption, while KL focuses on how informative a sample is relative to the global distribution. Our results show that KL-based filtering is competitive with LAION2B filtering and basic heuristics, despite relying only on an off-the-shelf CLIP-like model and simple density ratio estimates. This suggests that KL is a complementary signal that could be combined with alignment scores such as CLIPScore, for example, by first enforcing a minimum alignment threshold and then selecting the most informative samples among those. We leave a systematic exploration of such combined criteria to future work.

\section{Limitations}
Our claims stand on a strong assumption that the density ratio of the conditional and marginal distribution is equal to that of the training data. While CLIP-like models are trained on hundreds of millions to billions of image-text pairs and show strong empirical performance, biases and other characteristics inherent in the pretraining data and modeling errors may affect the accuracy of these density ratio estimates.
Also, we have conducted experiments on Importance-Weighting Learning and data curation relatively at a small scale. Further experiments of scaling up the model training are future work.

Beyond distributional assumptions, the KL divergence measures we use are finite-sample approximations based on a restricted set of candidate captions or images. The bias and variance of these estimations depend on factors such as the embeddings, the temperature parameter, and the sampling strategy used for selecting candidate samples.

Finally, our evaluations focus on moderately sized benchmarks, specifically CC12M and the smallest datapool of DataComp, and cover only a limited set of domains. Extending the analysis to larger-scale datasets, more diverse modalities, and more challenging distribution shifts represents an important direction for future research.

\section{Conclusion}
In this work, we revisited the interpretation of the similarity scores learned by CLIP-like models from the perspective of density ratio estimation. We demonstrated that CLIP-like models effectively capture density ratios, which can be leveraged for a variety of downstream tasks, including importance-weighted learning, quantifying contextual diversity, and data curation. We hope that our findings will inspire further research exploring the use of CLIP-like models as general-purpose density ratio estimators across both vision and language domains.
\clearpage
\section{Acknowledgment}
This work was supported by the AIST policy-based budget project ``R\&D on Generative AI Foundation Models for the Physical Domain''. This work was supported by Japan Science and Technology Agency (JST) as part of Adopting Sustainable Partnerships for Innovative Research Ecosystem (ASPIRE), Grant Number JPMJAP2518. We used ABCI 3.0 provided by AIST and AIST Solutions with support from ``ABCI 3.0 Development Acceleration Use''. This work was supported by JSPS KAKENHI Grant Number 23H04974.
{
    \small
    \bibliographystyle{ieeenat_fullname}
    \bibliography{main}
}

\clearpage
\setcounter{page}{1}
\maketitlesupplementary

\section{Derivations of \cref{eq:densratio,eq:densratio_txt}}
\label{sec:derivation_density_ratio}

In SigLIP, the learning objective is to classify whether a given pair of image and text is a positive pair or a negative pair. Given an image $i$, the text paired with $i$ could be considered as a sample from the conditional text distribution $p_{T}(\cdot|i)$, while the negative texts are sampled from the marginal text distribution $p_{T}(\cdot)$.

Fixing $i$, let's consider a set that contains one positive sample text paired with $i$, and $\nu$ other texts that are not paired with $i$. We assign binary labels $C$ where $C=1$ for the positive sample and $C=0$ for the negative samples. Therefore,

\begin{gather}
  P(t|C=1;i) = p_T(t|i) \\
  P(t|C=0;i) = p_T(t).
\end{gather}

The prior probabilities of class labels are $P(C=1)=\frac{1}{1+\nu}$ and $P(C=0)=\frac{\nu}{1+\nu}$. The posterior probability of positive sample is
\begin{align}
  &P(C=1|t;i) \notag\\
  &= \frac{P(t|C=1;i)P(C=1)}{P(t|C=1;i)P(C=1)+P(t|C=0;i)P(C=0)} \notag\\
  &= \frac{p_T(t|i)}{p_T(t|i)+\nu p_T(t)}.
\end{align}

Also, the posterior probability of negative sample $P(C=0|t;i)$ is $\tfrac{\nu p_T(t)}{p_T(t|i)+\nu p_T(t)}$.
Denoting the positive sample by $t^{\mathrm{pos}(i)}$, and the set of the negative samples by $\mathrm{neg}(i)$, the log-likelihood $L_C(i)$ on a set about $i$ is

\begin{align}
  &\!L_C(i) \notag\\
  &= 
  \log
  \left[
    P(C=1|t^{\mathrm{pos}(i)};i) \cdot
    \prod_{t'\in\mathrm{neg}(i)}P(C=0|t';i)
  \right]
  \notag\\
  &=
  \log\!
  \left[
    \frac{
        p_{T}(t^{\mathrm{pos}(i)}|i)
      }{
        p_{T}(t^{\mathrm{pos}(i)}|i)
        +\nu p_{T}(t^{\mathrm{pos}(i)})
      }
  \right]
  \notag\\
  &\quad + \sum_{t'\in\mathrm{neg}(i)}
  \log\!
  \left[
    \frac{
      \nu p_{T}(t')
    }{
      p_{T}(t'|i) + \nu p_{T}(t')
    }
  \right]
  \notag\\
  &= 
  \log\!
  \left[
    \frac{
      p_{T}(t^{\mathrm{p}(i)}|i)
    }{
      p_{T}(t^{\mathrm{p}(i)}|i)
      +\nu p_{T}(t^{\mathrm{p}(i)})
    }
  \right]
  \notag\\
  &\quad + \sum_{t'\in\mathrm{neg}(i)} 
  \log\!
  \left[
    1-\frac{p_{T}(t'|i)}{p_{T}(t'|i) + \nu p_{T}(t')}
  \right].
  \label{eq:likelihood_nce}
\end{align}

On the other hand, a batch of SigLIP has $\nu +1$ elements, where the $j$th element is a text set composed of one text paired with an image $i_j$ and $\nu$ other texts, which is the same configuration as the previous one. The SigLIP loss is 
\begin{align}
  L_{\mathrm{SigLIP}}
  &= 
  -\frac{1}{N}\sum_{j=1}^N
  \Bigg[
    \log\!
    \left[
      \sigma\!(s(t^{\mathrm{pos}(i_j)}, i_j))
    \right]
  \\
  &\quad+\sum_{t'\in\mathrm{neg}(i_j)}
    \log\!
    \left[
      \sigma\!(-s(t',i_j))
    \right]
  \Bigg]
  \notag\\
  &= 
  -\frac{1}{N}\sum_{j=1}^N
  \Bigg[
    \log\!
    \left[
      \sigma\!(s(t^{\mathrm{pos}(i_j)}, i_j))
    \right]
  \\
  &\quad+\sum_{t'\in\mathrm{neg}(i_j)}
    \log\!
    \left[
      1-\sigma\!(s(t', i_j))
    \right]
  \Bigg]
  \label{eq:siglip}
\end{align}

where $N:=\nu +1$ and $s(t,i):=a\langle v_t, v_i \rangle + b$ that $b$ is a scalar called logit bias. $\sigma(x) := \frac{1}{1 + \exp(-x)}$ is sigmoid function. The last equation holds because $\sigma(-x) = 1 - \sigma(x)$.

Compared with \cref{eq:likelihood_nce} and \cref{eq:siglip}, Minimizing $L_\mathrm{NCE}$ is equal to maximizing summation of $L(i)$ \wrt images and the optimal embeddings $\hat{v}_t, \hat{v}_i$ satisfy following equation:
\begin{equation}
  \frac{p_{T}(t|i)}{p_{T}(t|i) + \nu p_{T}(t)} = \frac{1}{1 
  + \exp(-a \langle \hat{v}_t, \hat{v}_i \rangle - b)}
\end{equation}
when assuming a one-to-one mapping from text to embedding.
With some equivalent transformation, we obtain
\begin{equation}
  \frac{p_{T}(t|i)}{p_{T}(t)} = \nu e^b \exp(a \langle v_t, v_i \rangle).
\end{equation}

$\nu$ and $b$ are constant through samples, and $\int p_{T}(t|i) dt = \int p_{T}(t) \exp(a \langle v_t, v_i \rangle)/Z(i) dt = 1$, thus \cref{eq:densratio} holds and $Z(i)^{-1} = \nu e^b$ in an ideal modeling.

A batch of SigLIP is also regarded as a batch whose $j$-th element is an image set composed of one image paired with a text $t_j$ and $\nu$ other images, since the batch is made from $N$ image-text pairs and negative samples are prepared from other pairs. So \cref{eq:densratio_txt} holds symmetrically.

While SigLIP is learning binary classification, CLIP is based on multi-label classification in a batch. Let's assign an index for each sample instead of a binary label in a text set, and assume that the positive text paired with an image $i_j$ exists at the $j$-th position in a text set. We denote the $j$-th sample text by $t_j$ for $1\leq~j\leq N$. This configuration makes a multi-label classification problem, which predicts the index of the text that corresponds to the input image. The log-likelihood for image-to-text prediction is therefore:

\begin{align}
  L_D(i_j)
  &=\log\left[P(D=j|t_1,\dots ,t_N;i_j)\right] \notag\\
  &= 
    \log 
    \left[
      \frac{
          p_T(t_j|i_j)\prod_{l\neq j}p_T(t_l)
        }{
          \sum_{k=1}^N
          p_T(t_k|i_j)\prod_{l\neq k}p_T(t_l)
        }
    \right] 
    \notag\\
  &=
    \log
    \left[
      \frac{
          \frac{p_T(t_j|i_j)}{p_T(t_j)}
        }{
          \sum_{k=1}^N
          \frac{p_T(t_k|i_j)}{p_T(t_j)}
        }
    \right].
\end{align}
In the same way, the log-likelihood of text-to-image prediction is
\begin{align}
  L_D(t_j)
  &=\log\left[P(D=j|i_1,\dots ,i_N;t_j)\right] \notag\\
  &=
    \log
    \left[
      \frac{
          \frac{p_I(i_j|t_j)}{p_I(i_j)}
        }{
          \sum_{k=1}^N
          \frac{p_I(i_k|t_j)}{p_I(i_j)}
        }
    \right].
\end{align}
CLIP loss functions are designed for maximizing the log-likelihood of image-to-text prediction and text-to-image prediction as follows:
\begin{align}
  L_{\mathrm{CLIP}}
  &= 
    -\sum_{j=1}^N
    \underbrace{
    \log\!
    \left[
      \frac{
        \exp(s(t_j, i_j))
      }{
        \sum_{k=1}^{N}
        \exp(s(t_k, i_j))
      }
    \right]
    }_{\text{image-to-text prediction}}
    \notag
    \\
  &\quad-\sum_{j=1}^N
    \underbrace{
    \log\!
    \left[
      \frac{
        \exp(s(t_j, i_j))
      }{
        \sum_{k=1}^{N}
        \exp(s(t_j, i_k))
      }
    \right]
    }_{\text{text-to-image prediction}}
  \label{eq:clip}
\end{align}
where $s(t,i):=a\langle v_t, v_i \rangle$ is the score of CLIP that does not include logit bias $b$.
The optimal score $\hat{s}$ is proportional to the density ratio of conditional and marginal probabilities for both image and text modality:
\begin{align}
  \hat{s}(t,i) 
  \propto \frac{p_T(t|i)}{p_T(t)} 
  = \frac{p_I(i|t)}{p_I(i)}
  .
\end{align} 

\section{Derivations of \cref{eq:dkl_approx,eq:dklr_approx}}
\label{sec:kl_div_deriv_detail}
To treat images and text in the same manner, we abuse $p_T$ like $p_T(t)=p_T(v_t)$ and $p_T(t|i)=p_T(v_t|v_i)$.

By substituting \Cref{eq:densratio} into \Cref{eq:dkl},
\begin{align}
  D_\mathrm{KL}(i)
  &= \int
  p_T(v_{t'}|v_{i})
  \log
  \frac{p_T(v_{t'}|v_{i})}{p_T(v_{t'})}
  dv_{t'}
  \notag
  \\
  &= \int
  p_T(v_{t'})
  \frac{\exp (a\langle v_{t'},v_i\rangle)}{Z(i)}
  \log
  \frac{\exp (a\langle v_{t'},v_i\rangle)}{Z(i)}
  dv_{t'}
  \notag
  \\
  &= \frac{1}{Z(i)}\int
  p_T(v_{t'})\exp (a\langle v_{t'},v_i\rangle)
  a\langle v_{t'},v_i\rangle
  dv_{t'}
  \notag
  \\
  &\quad
  -\frac{1}{Z(i)}\log(Z(i))\int
  p_T(v_{t'})\exp (a\langle v_{t'},v_i\rangle)
  dv_{t'}
  \notag
  \\
  &=\frac{1}{Z(i)}
  \mathbb{E}_{t'\sim p_T(\cdot)}
  \left[
    a\langle v_{t'},v_i\rangle 
    e^{a\langle v_{t'},v_i\rangle}
  \right]
  -\log Z(i).
  \label{eq:dkl2}
\end{align}
Note that $Z(i)=\mathbb{E}_{t\sim p_T(\cdot)}[e^{a \langle v_t, v_i \rangle}]$. \cref{eq:dkl2} can be estimated by text samples $\mathcal{D}_T$ as
\begin{align}
  D_\mathrm{KL}(i)
  &= \frac{
    \mathbb{E}_{t'\sim p_T(\cdot)}
    \left[
      a\langle v_{t'},v_i\rangle 
      e^{a\langle v_{t'},v_i\rangle}
    \right]
  }{
    \mathbb{E}_{t\sim p_T(\cdot)}[e^{a \langle v_t, v_i \rangle}]
  }
  \notag
  \\
  &\quad
  -\log
  \mathbb{E}_{t'\sim p_T(\cdot)}
  \left[
    a\langle v_{t'},v_i\rangle 
    e^{a\langle v_{t'},v_i\rangle}
  \right]
  \notag
  \\
  &\approx\frac{
    \sum_{t'\in\mathcal{D}_T}
    a\langle v_{t'},v_i\rangle 
    e^{a\langle v_{t'},v_i\rangle}
  }{
    \sum_{t\in\mathcal{D}_T}
    e^{a \langle v_t, v_i \rangle}
  }
  \notag
  \\
  &\quad
  -\log
  \frac{1}{|\mathcal{D}_T|}
  \sum_{t'\in\mathcal{D}_T}
  a\langle v_{t'},v_i\rangle 
  e^{a\langle v_{t'},v_i\rangle}
  \notag
  \\
  &=
  \sum_{t'\in\mathcal{D}_T}
  a\langle v_{t'},v_i\rangle 
  \mathrm{softmax}_i(t')
  \notag
  \\
  &\quad
  -\log
  \frac{1}{|\mathcal{D}_T|}
  \sum_{t'\in\mathcal{D}_T}
  a\langle v_{t'},v_i\rangle 
  e^{a\langle v_{t'},v_i\rangle}
  \notag
  \\
  &=
  \sum_{t\in\mathcal{D}_T}
  a\langle v_{t},v_i\rangle 
  \mathrm{softmax}_i(t)
  \notag
  \\
  &\quad
  -\log
  \sum_{t\in\mathcal{D}_T}
  a\langle v_{t},v_i\rangle 
  e^{a\langle v_{t},v_i\rangle}
  +\log |\mathcal{D}_T|
\end{align}
where $\mathrm{softmax}_i(t):=\frac{e^{a\langle v_{t},v_i\rangle}}{\sum_{t'\in\mathcal{D}_T}e^{a \langle v_{t'}, v_i \rangle}}$.

\Cref{eq:dklr} is also transformed with \cref{eq:densratio} as follows:
\begin{align}
  D_\mathrm{KLR}(i)
  &= \int
  p_T(v_{t'})
  \log
  \frac{p_T(v_{t'})}{p_T(v_{t'}|v_{i})}
  dv_{t'}
  \notag
  \\
  &= \int
  p_T(v_{t'})
  \log
  \frac{Z(i)}{\exp (a\langle v_{t'},v_i\rangle)}
  dv_{t'}
  \notag
  \\
  &= \log Z(i)
  \int
  p_T(v_{t'})
  dv_{t'}
  \notag
  \\
  &\quad
  -\int
  p_T(v_{t'})
  a\langle v_{t'},v_i\rangle
  dv_{t'}
  \notag
  \\
  &=\log Z(i)
  -\mathbb{E}_{t'\sim p_T(\cdot)}
  \left[
    a\langle v_{t'},v_i\rangle 
  \right]
  \label{eq:dklr2}
\end{align}
With test samples $\mathcal{D}_T$,
\begin{align}
  D_\mathrm{KLR}(i)
  &\approx
  \log
  \frac{1}{|\mathcal{D}_T|}
  \sum_{t\in\mathcal{D}_T}
  e^{a\langle v_{t},v_i\rangle}
  - \frac{1}{|\mathcal{D}_T|}
  \sum_{t\in\mathcal{D}_T}
  a\langle v_t,v_i\rangle
  \notag
  \\
  &=
  \log
  \sum_{t\in\mathcal{D}_T}
  e^{a\langle v_{t},v_i\rangle}
  -\log |\mathcal{D}_T|
  \notag
  \\
  &\quad- \frac{1}{|\mathcal{D}_T|}
  \sum_{t\in\mathcal{D}_T}
  a\langle v_t,v_i\rangle
\end{align}

\section{Hyper-parameters for \cref{sec:apps}}
\label{sec:hyperparam_iwl}
We trained ViT-B/32 models on the CC12M~\cite{changpinyo2021cc12m} dataset using the official OpenCLIP training pipeline. Unless otherwise noted, we follow the default OpenCLIP configuration. We use AdamW with a learning rate of $5\times 10^{-4}$, weight decay of 0.2, and mixed-precision training (AMP). Each training run uses a single node equipped with eight GPUs with a per-GPU batch size of 1000 and eight dataloader workers, resulting in a global batch size of 8000. Models are trained for ten epochs. We enable Importance-Weighted Learning (IWL) using a publicly available ViT-L/14 model pretrained on LAION2B~\cite{ilharco_gabriel_2021_5143773} as the reference model. The end-to-end training cost is approximately 70 H200 GPU hours per model.
\section{Other results of \cref{sec:kl_feature}}
\label{sec:kl_features_others}
\Cref{fig:mscoco_image_samples_dklr,fig:mscoco_image_samples_dc,fig:mscoco_image_samples_dw} show top and bottom 18 images ranked by $D_\mathrm{KLR}$, $D_\mathrm{C}$, $D_\mathrm{W}$. Some bottom samples of $D_\mathrm{C}$ have shared concepts such as patterned animals, and the top samples of $D_\mathrm{W}$ resemble the bottom samples of $D_\mathrm{KL}$. However, unlike $D_\mathrm{KL}$, qualitative differences between top and bottom samples are less distinct. 
\Cref{fig:mscoco_text_samples_dklr,fig:mscoco_text_samples_dc,fig:mscoco_text_samples_dw} are the text samples. $D_\mathrm{KLR}$ and $D_\mathrm{W}$ capture meaningless captions at their bottom score, while $D_\mathrm{C}$ seems to represent specificity for image content.
\Cref{fig:mscoco_ngram_other} is the cumulative N-gram probability of captions grouped by each KL divergence. Interestingly, captions associated with low $D_\mathrm{W}$ show more diverse N-gram distribution. Further analysis of those metrics is left for future work.
\begin{figure*}[t]
  \centering
  \begin{subfigure}{\linewidth}
    \centering
    \imgcard{}{36.697}{red!80!black}\hfill
    \imgcard{}{36.021}{red!80!black}\hfill
    \imgcard{}{35.794}{red!80!black}\hfill
    \imgcard{}{35.591}{red!80!black}\hfill
    \imgcard{}{35.466}{red!80!black}\hfill
    \imgcard{}{35.238}{red!80!black}\\[0.5em]
    \imgcard{}{34.853}{red!80!black}\hfill
    \imgcard{}{34.726}{red!80!black}\hfill
    \imgcard{}{34.668}{red!80!black}\hfill
    \imgcard{}{34.657}{red!80!black}\hfill
    \imgcard{}{34.655}{red!80!black}\hfill
    \imgcard{}{34.371}{red!80!black}\\[0.5em]
    \imgcard{}{34.344}{red!80!black}\hfill
    \imgcard{}{34.125}{red!80!black}\hfill
    \imgcard{}{34.124}{red!80!black}\hfill
    \imgcard{}{34.091}{red!80!black}\hfill
    \imgcard{}{34.090}{red!80!black}\hfill
    \imgcard{}{33.667}{red!80!black}
    \caption{Top 18 images}
  \end{subfigure}
  \hfill
  \vspace{1em}
  \begin{subfigure}{\linewidth}
    \centering
    \imgcard{}{9.1247835}{blue!80!black}\hfill
    \imgcard{}{9.6127}{blue!80!black}\hfill
    \imgcard{}{10.961}{blue!80!black}\hfill
    \imgcard{}{12.163}{blue!80!black}\hfill
    \imgcard{}{13.026}{blue!80!black}\hfill
    \imgcard{}{13.076}{blue!80!black}\\[0.5em]
    \imgcard{}{13.123}{blue!80!black}\hfill
    \imgcard{}{13.179}{blue!80!black}\hfill
    \imgcard{}{13.219}{blue!80!black}\hfill
    \imgcard{}{13.402}{blue!80!black}\hfill
    \imgcard{}{13.728}{blue!80!black}\hfill
    \imgcard{}{13.736}{blue!80!black}\\[0.5em]
    \imgcard{}{13.761}{blue!80!black}\hfill
    \imgcard{}{13.841}{blue!80!black}\hfill
    \imgcard{}{13.858}{blue!80!black}\hfill
    \imgcard{}{13.862}{blue!80!black}\hfill
    \imgcard{}{13.93}{blue!80!black}\hfill
    \imgcard{}{14.027}{blue!80!black}
    \caption{Bottom 18 images}
  \end{subfigure}
  \caption{
  Top and bottom image examples in MSCOCO captions ranked by $D_\mathrm{KLR}$ through cosine similarity of CLIP.
  }
  \label{fig:mscoco_image_samples_dklr}
\end{figure*}

\begin{figure*}[t]
  \centering
  \begin{subfigure}{\linewidth}
    \centering
    \imgcard{}{10882}{red!80!black}\hfill
    \imgcard{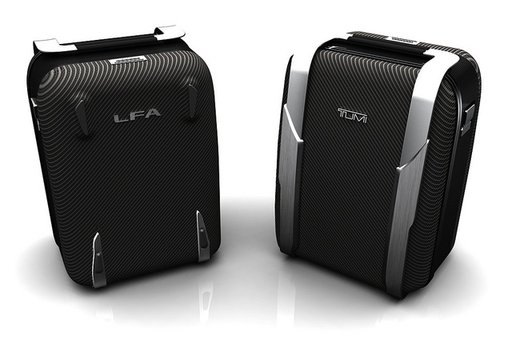}{10381}{red!80!black}\hfill
    \imgcard{}{10325}{red!80!black}\hfill
    \imgcard{}{10193}{red!80!black}\hfill
    \imgcard{}{10165}{red!80!black}\hfill
    \imgcard{}{10068}{red!80!black}\\[0.5em]
    \imgcard{}{9970.6}{red!80!black}\hfill
    \imgcard{}{9951.3}{red!80!black}\hfill
    \imgcard{}{9908.1}{red!80!black}\hfill
    \imgcard{}{9892.9}{red!80!black}\hfill
    \imgcard{}{9772.6}{red!80!black}\hfill
    \imgcard{}{9722.9}{red!80!black}\\[0.5em]
    \imgcard{}{9682.2}{red!80!black}\hfill
    \imgcard{}{9580.6}{red!80!black}\hfill
    \imgcard{}{9522.8}{red!80!black}\hfill
    \imgcard{}{9503.4}{red!80!black}\hfill
    \imgcard{}{9421.7}{red!80!black}\hfill
    \imgcard{}{9367.8}{red!80!black}
    \caption{Top 18 images}
  \end{subfigure}
  \hfill
  \vspace{1em}
  \begin{subfigure}{\linewidth}
    \centering
    \imgcard{}{4571.2}{blue!80!black}\hfill
    \imgcard{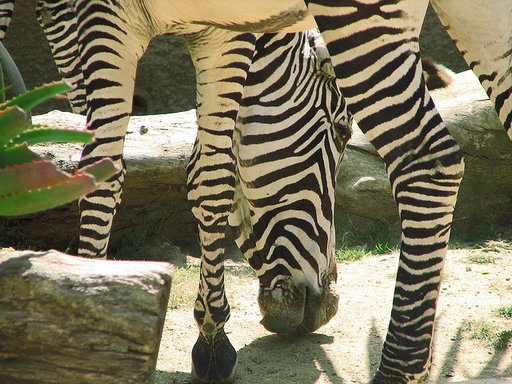}{4584.3}{blue!80!black}\hfill
    \imgcard{}{4638.3}{blue!80!black}\hfill
    \imgcard{}{4644.4}{blue!80!black}\hfill
    \imgcard{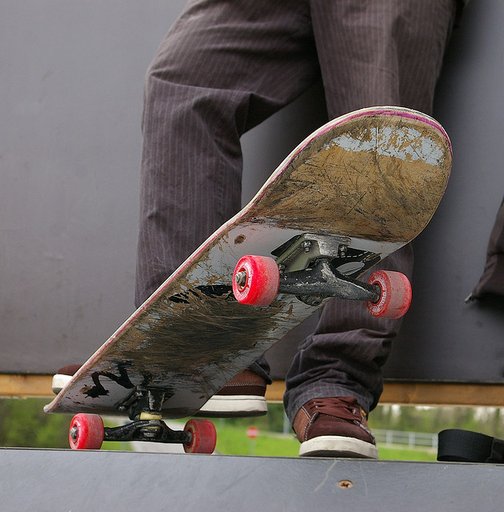}{4695.8}{blue!80!black}\hfill
    \imgcard{}{4699.4}{blue!80!black}\\[0.5em]
    \imgcard{}{4708.2}{blue!80!black}\hfill
    \imgcard{}{4713}{blue!80!black}\hfill
    \imgcard{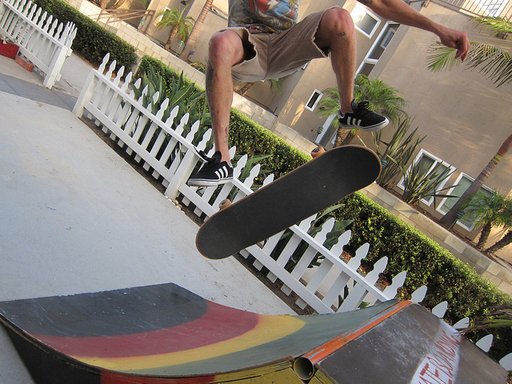}{4722.3}{blue!80!black}\hfill
    \imgcard{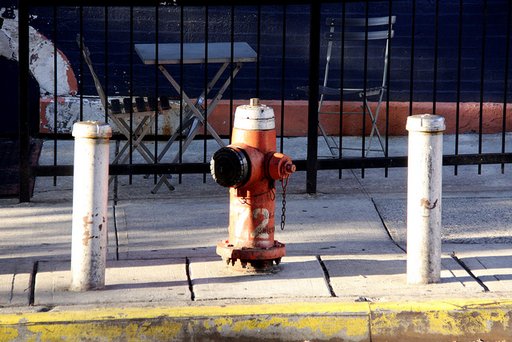}{4744.8}{blue!80!black}\hfill
    \imgcard{}{4752.4}{blue!80!black}\hfill
    \imgcard{}{4756.7}{blue!80!black}\\[0.5em]
    \imgcard{}{4760.6}{blue!80!black}\hfill
    \imgcard{}{4791.1}{blue!80!black}\hfill
    \imgcard{}{4792.8}{blue!80!black}\hfill
    \imgcard{}{4794.8}{blue!80!black}\hfill
    \imgcard{}{4799.7}{blue!80!black}\hfill
    \imgcard{}{4799.8}{blue!80!black}
    \caption{Bottom 18 images}
  \end{subfigure}
  \caption{
  Top and bottom image examples in MSCOCO captions ranked by $D_\mathrm{C}$ through cosine similarity of CLIP.
  }
  \label{fig:mscoco_image_samples_dc}
\end{figure*}

\begin{figure*}[t]
  \centering
  \begin{subfigure}{\linewidth}
    \centering
    \imgcard{}{1.6126e+06}{red!80!black}\hfill
    \imgcard{}{1.4868e+06}{red!80!black}\hfill
    \imgcard{}{1.4259e+06}{red!80!black}\hfill
    \imgcard{}{1.3936e+06}{red!80!black}\hfill
    \imgcard{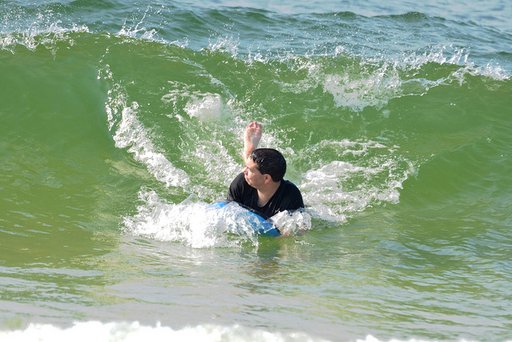}{1.3591e+06}{red!80!black}\hfill
    \imgcard{}{1.3264e+06}{red!80!black}\\[0.5em]
    \imgcard{}{1.3179e+06}{red!80!black}\hfill
    \imgcard{}{1.3175e+06}{red!80!black}\hfill
    \imgcard{}{1.3089e+06}{red!80!black}\hfill
    \imgcard{}{1.3064e+06}{red!80!black}\hfill
    \imgcard{}{1.3022e+06}{red!80!black}\hfill
    \imgcard{}{1.3011e+06}{red!80!black}\\[0.5em]
    \imgcard{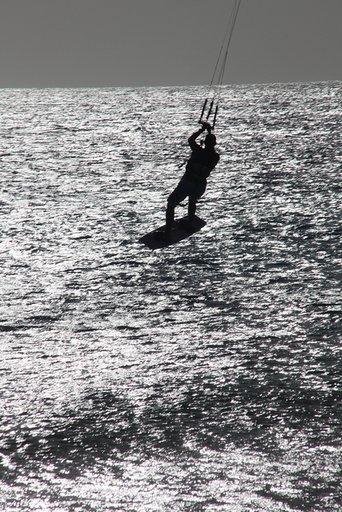}{1.2953e+06}{red!80!black}\hfill
    \imgcard{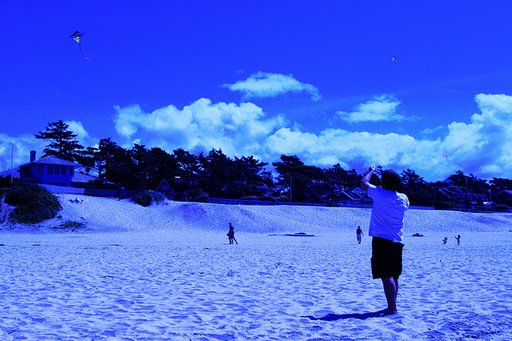}{1.2903e+06}{red!80!black}\hfill
    \imgcard{}{1.2885e+06}{red!80!black}\hfill
    \imgcard{}{1.287e+06}{red!80!black}\hfill
    \imgcard{}{1.2831e+06}{red!80!black}\hfill
    \imgcard{}{1.2795e+06}{red!80!black}
    \caption{Top 18 images}
  \end{subfigure}
  \hfill
  \vspace{1em}
  \begin{subfigure}{\linewidth}
    \centering
    \imgcard{}{2.4622e+05}{blue!80!black}\hfill
    \imgcard{}{2.9241e+05}{blue!80!black}\hfill
    \imgcard{}{2.9877e+05}{blue!80!black}\hfill
    \imgcard{}{3.0105e+05}{blue!80!black}\hfill
    \imgcard{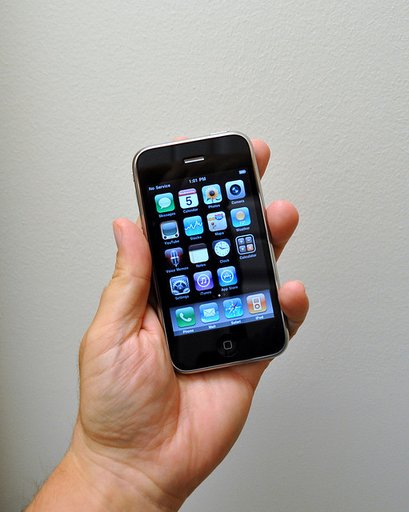}{3.0177e+05}{blue!80!black}\hfill
    \imgcard{}{3.0261e+05}{blue!80!black}\\[0.5em]
    \imgcard{}{3.1397e+05}{blue!80!black}\hfill
    \imgcard{}{3.1781e+05}{blue!80!black}\hfill
    \imgcard{}{3.18e+05}{blue!80!black}\hfill
    \imgcard{}{3.1869e+05}{blue!80!black}\hfill
    \imgcard{}{3.3028e+05}{blue!80!black}\hfill
    \imgcard{}{3.3789e+05}{blue!80!black}\\[0.5em]
    \imgcard{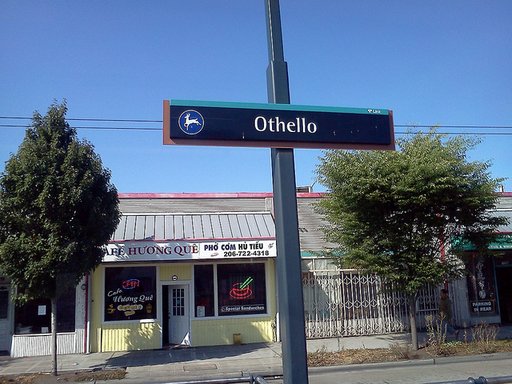}{3.4255e+05}{blue!80!black}\hfill
    \imgcard{}{3.4399e+05}{blue!80!black}\hfill
    \imgcard{}{3.4868e+05}{blue!80!black}\hfill
    \imgcard{}{3.4994e+05}{blue!80!black}\hfill
    \imgcard{}{3.5154e+05}{blue!80!black}\hfill
    \imgcard{}{3.5258e+05}{blue!80!black}
    \caption{Bottom 18 images}
  \end{subfigure}
  \caption{
  Top and bottom image examples in MSCOCO captions ranked by $D_\mathrm{W}$ through cosine similarity of CLIP.
  }
  \label{fig:mscoco_image_samples_dw}
\end{figure*}

\begin{figure*}[t]
  \centering
  \begin{subfigure}{\linewidth}
    \centering
    \textcard{A waitress and restaurant owner showing off a pizza in front of a mural of Venice.}{38.591}{red!80!black}\hfill
    \textcard{A group photo of men and boys from the Goodmayes Boys School dated April 1929.}{37.113}{red!80!black}\hfill
    \textcard{A green street sign that says Bodacious Drive.}{36.416}{red!80!black}\hfill
    \textcard{Sign of Arch Street Fire Department saying "He who laughs last thinks slowest"}{36.341}{red!80!black}\hfill
    \textcard{A bottle of beer sits next to a computer at a desk}{36.214}{red!80!black}\hfill
    \textcard{A street sign reads "Bacchanalia" above a stop sign.}{36.101}{red!80!black}\\[0.5em]
    \textcard{a street sign for Library Way and Madison Avenue above a one way arrow}{35.955}{red!80!black}\hfill
    \textcard{A ripe tangerine atop a worn wooden picnic table.}{35.796}{red!80!black}\hfill
    \textcard{A small cat sitting in the top of a banana tree.}{35.741}{red!80!black}\hfill
    \textcard{A bottle of beer sits next to the keyboard and mouse at the computer table.}{35.619}{red!80!black}\hfill
    \textcard{A black and white photo of four elderly people sitting in a bench overlooking the ocean.}{35.522}{red!80!black}\hfill
    \textcard{A class picture of the Goodmayes Boys' School in April of 1929.}{35.281}{red!80!black}\\[0.5em]
    \textcard{Two almost identical photos with some minor cropping show two young girls standing on the coast with people in the water in the background as a plane flies overhead}{35.094}{red!80!black}\hfill
    \textcard{Assortment of laptop computers displayed on table with backpacks full of electronic cords.}{35.062}{red!80!black}\hfill
    \textcard{A picture of a road sign labeled bodacious drive.}{35.05}{red!80!black}\hfill
    \textcard{A vintage baby doll with the book "Goody two Shoes" on it's lap}{34.898}{red!80!black}\hfill
    \textcard{A red fire hydrant is sitting in the woods near fallen leaves.}{34.83}{red!80!black}\hfill
    \textcard{A computer on a desk with a bottle of beer next to it}{34.745}{red!80!black}
    \caption{Top 18 texts}
  \end{subfigure}
  \vspace{1em}
  \begin{subfigure}{\linewidth}
    \centering
    \textcard{There is no image to be reviewed on this hit.}{1.5582}{blue!80!black}\hfill
    \textcard{unable to see this image in this particular hit}{1.7586}{blue!80!black}\hfill
    \textcard{There is no image here to provide a caption for.}{1.8616}{blue!80!black}\hfill
    \textcard{There is no image here to provide a caption for.}{1.8616}{blue!80!black}\hfill
    \textcard{There is no image here to provide a caption for.}{1.8616}{blue!80!black}\hfill
    \textcard{There is no image here to provide a caption for.}{1.8616}{blue!80!black}\\[0.5em]
    \textcard{no image again there are a lot of these lately}{1.9015}{blue!80!black}\hfill
    \textcard{There are some appealing support ready to expend. 
    }{1.954}{blue!80!black}\hfill
    \textcard{I am unable to see the image above.}{1.9708}{blue!80!black}\hfill
    \textcard{I am not sure what this image is.}{1.9813}{blue!80!black}\hfill
    \textcard{There is no image to describe for this question.}{2.0279}{blue!80!black}\hfill
    \textcard{I do not know what this is supposed to be..}{2.2007}{blue!80!black}\\[0.5em]
    \textcard{I am unable to see an image above.}{2.5721}{blue!80!black}\hfill
    \textcard{I am unable to see an image above.}{2.5721}{blue!80!black}\hfill
    \textcard{I am unable to see an image above.}{2.5721}{blue!80!black}\hfill
    \textcard{I am unable to see an image above.}{2.5721}{blue!80!black}\hfill
    \textcard{I am unable to see an image above.}{2.5721}{blue!80!black}\hfill
    \textcard{A being is doing something as of right now that is splendid. 
    }{3.137}{blue!80!black}
    \caption{Bottom 18 texts}
  \end{subfigure}
  \caption{Top and bottom captions ranked by $D_\mathrm{KLR}$.}
  \label{fig:mscoco_text_samples_dklr}
\end{figure*}

\begin{figure*}[t]
  \centering
  \begin{subfigure}{\linewidth}
    \centering
    \textcard{Side by side view of two oval plates, one with fork, with chicken salad sandwiches and rosy new potatoes, by an open and an unopened bottle of lager, a pepper mill, paper towel roll, basket behind.}{10591}{red!80!black}\hfill
    \textcard{A blue and black ray holding sandwich, coffee and liquor bottles.}{10313}{red!80!black}\hfill
    \textcard{Three rectangular bowls with food; Big bowl has nine meat and sesame seed patties with brown sauce, next to it, a bowl of shredded cabbage and carrots with yogurt dollop atop, and behind that is a bowl of cut broccoli and tomatoes with seasoning.}{10278}{red!80!black}\hfill
    \textcard{Buildings with large signs "Park Here" with rainy cement}{10138}{red!80!black}\hfill
    \textcard{Here is Massachusettes candidate Scott Brown's campaign trailer.}{10115}{red!80!black}\hfill
    \textcard{A kale and sweet pea home garden getting the last rays of sunlight. }{10037}{red!80!black}\\[0.5em]
    \textcard{Sign of Arch Street Fire Department saying "He who laughs last thinks slowest"}{10034}{red!80!black}\hfill
    \textcard{Four bowls of snacks: crackers, broccoli and carrots, nuts and dip}{9961.7}{red!80!black}\hfill
    \textcard{Bananas, marshmellows, chocolate chips and sprinkles in a bowl}{9957.1}{red!80!black}\hfill
    \textcard{Tube of orange and pears, surrounded by boxes of grass and one pineapple}{9946.3}{red!80!black}\hfill
    \textcard{Two apples, a bowl of food with berries and a pitcher and spoon on a purple place mat.}{9937.1}{red!80!black}\hfill
    \textcard{A woman wearing a blue t-shirt while looking at her cell phone and sitting on a bench next to a bright pink wall.}{9937}{red!80!black}\\[0.5em]
    \textcard{A view of the street signs "W 122 St.", "Seminary Row", and "Broadway" in front of an old red brick building. }{9923.6}{red!80!black}\hfill
    \textcard{United States President Barack Obama gives a speech in front of American and Russian flags.}{9913.7}{red!80!black}\hfill
    \textcard{A young redheaded woman in sunglasses and black tank top holds a black leather purse and a white umbrella.}{9906.4}{red!80!black}\hfill
    \textcard{A clock tower that is blue reading "Lenox Hill Hospital"}{9895.7}{red!80!black}\hfill
    \textcard{A runner poses beneath a "Run for Rights" sign in a green city park. }{9895}{red!80!black}\hfill
    \textcard{A bunch of small red flowers in a barnacle encrusted clay vase }{9881}{red!80!black}
    \caption{Top 18 texts}
  \end{subfigure}
  \vspace{1em}
  \begin{subfigure}{\linewidth}
    \centering
    \textcard{A lower shot of someone riding their skateboard. }{5388.3}{blue!80!black}\hfill
    \textcard{A man is performing tricks on a skateboard.}{5446.4}{blue!80!black}\hfill
    \textcard{A player in action up to bat in a baseball game.}{5463.3}{blue!80!black}\hfill
    \textcard{A player is up to bat in a baseball game.}{5464.1}{blue!80!black}\hfill
    \textcard{A train is traveling along a stretch of track.}{5464.3}{blue!80!black}\hfill
    \textcard{A person that is doing a trick on a skateboard.}{5486.2}{blue!80!black}\\[0.5em]
    \textcard{A baseball player at bat during a baseball game.}{5501.7}{blue!80!black}\hfill
    \textcard{A man is doing tricks on a skateboard.}{5505.4}{blue!80!black}\hfill
    \textcard{A man is doing tricks on a skateboard.}{5505.4}{blue!80!black}\hfill
    \textcard{A picture of a person on a skateboard.}{5506.3}{blue!80!black}\hfill
    \textcard{A rider is riding a horse at a race track.}{5511.6}{blue!80!black}\hfill
    \textcard{A person is riding on a skateboard on the street.}{5513.1}{blue!80!black}\\[0.5em]
    \textcard{A person rides a bike on the road.}{5514.4}{blue!80!black}\hfill
    \textcard{A man who is riding on a skateboard.}{5515.6}{blue!80!black}\hfill
    \textcard{A man is doing a trick on a skateboard.}{5523}{blue!80!black}\hfill
    \textcard{A man is doing a trick on a skateboard.}{5523}{blue!80!black}\hfill
    \textcard{A train that is riding on tracks through a station.}{5528.2}{blue!80!black}\hfill
    \textcard{A picture of a person playing in a tennis game.}{5549.5}{blue!80!black}
    \caption{Bottom 18 texts}
  \end{subfigure}
  \caption{Top and bottom captions ranked by $D_\mathrm{C}$.}
  \label{fig:mscoco_text_samples_dc}
\end{figure*}

\begin{figure*}[t]
  \centering
  \begin{subfigure}{\linewidth}
    \centering
    \textcard{Buses parked on a road outside a large bus station. }{3.0478e+05}{red!80!black}\hfill
    \textcard{A city road with buses and people on the sidewalk}{2.8882e+05}{red!80!black}\hfill
    \textcard{A city road with buses and bus park}{2.8675e+05}{red!80!black}\hfill
    \textcard{A housewife holds a platter of food in the kitchen.}{2.8587e+05}{red!80!black}\hfill
    \textcard{A city street with cars and street signs giving directions}{2.8556e+05}{red!80!black}\hfill
    \textcard{A bus driving in a city area with traffic signs.}{2.8245e+05}{red!80!black}\\[0.5em]
    \textcard{Several animals standing in the grass near a lake.}{2.8216e+05}{red!80!black}\hfill
    \textcard{Covered food is sitting on a kitchen counter. }{2.8135e+05}{red!80!black}\hfill
    \textcard{African animals placidly grazing in a park enclosure.}{2.803e+05}{red!80!black}\hfill
    \textcard{People crossing the street near a parked sightseeing bus.}{2.7734e+05}{red!80!black}\hfill
    \textcard{many animals in a field with trees and bushes in the background}{2.7708e+05}{red!80!black}\hfill
    \textcard{A herd of wild animals walking across a dry grass park.}{2.7704e+05}{red!80!black}\\[0.5em]
    \textcard{two public transit buses on a city street }{2.7531e+05}{red!80!black}\hfill
    \textcard{A herd of animals grazing on a dry grass field.}{2.7424e+05}{red!80!black}\hfill
    \textcard{a street filled with older cars, buses and motorcycles }{2.7331e+05}{red!80!black}\hfill
    \textcard{A group of animals grazing on a lush green field.}{2.7215e+05}{red!80!black}\hfill
    \textcard{A city street with a road sign next to some buildings}{2.7193e+05}{red!80!black}\hfill
    \textcard{A bus traveling down the street in front of a large building with a clock tower.}{2.7066e+05}{red!80!black}
    \caption{Top 18 texts}
  \end{subfigure}
  \vspace{1em}
  \begin{subfigure}{\linewidth}
    \centering
    \textcard{There is no image to be reviewed on this hit.}{18497}{blue!80!black}\hfill
    \textcard{A person is taken in this very picture.
    }{25216}{blue!80!black}\hfill
    \textcard{There is no image here to provide a caption for.}{25437}{blue!80!black}\hfill
    \textcard{There is no image here to provide a caption for.}{25437}{blue!80!black}\hfill
    \textcard{There is no image here to provide a caption for.}{25437}{blue!80!black}\hfill
    \textcard{There is no image here to provide a caption for.}{25437}{blue!80!black}\\[0.5em]
    \textcard{I am unable to see the image above.}{26402}{blue!80!black}\hfill
    \textcard{unable to see this image in this particular hit}{26929}{blue!80!black}\hfill
    \textcard{I am unable to see an image above.}{28398}{blue!80!black}\hfill
    \textcard{I am unable to see an image above.}{28398}{blue!80!black}\hfill
    \textcard{I am unable to see an image above.}{28398}{blue!80!black}\hfill
    \textcard{I am unable to see an image above.}{28398}{blue!80!black}\\[0.5em]
    \textcard{I am unable to see an image above.}{28398}{blue!80!black}\hfill
    \textcard{no image again there are a lot of these lately}{29587}{blue!80!black}\hfill
    \textcard{I am not sure what this image is.}{30736}{blue!80!black}\hfill
    \textcard{I do not know what this is supposed to be..}{31176}{blue!80!black}\hfill
    \textcard{An individual is taken in this very picture. 
    }{32908}{blue!80!black}\hfill
    \textcard{An individual is taken in this very picture. 
    }{32908}{blue!80!black}
    \caption{Bottom 18 texts}
  \end{subfigure}
  \caption{Top and bottom captions ranked by $D_\mathrm{W}$.}
  \label{fig:mscoco_text_samples_dw}
\end{figure*}

\begin{figure*}[t]
  \centering
  \begin{subfigure}{0.49\linewidth}
    \includegraphics[width=\linewidth]{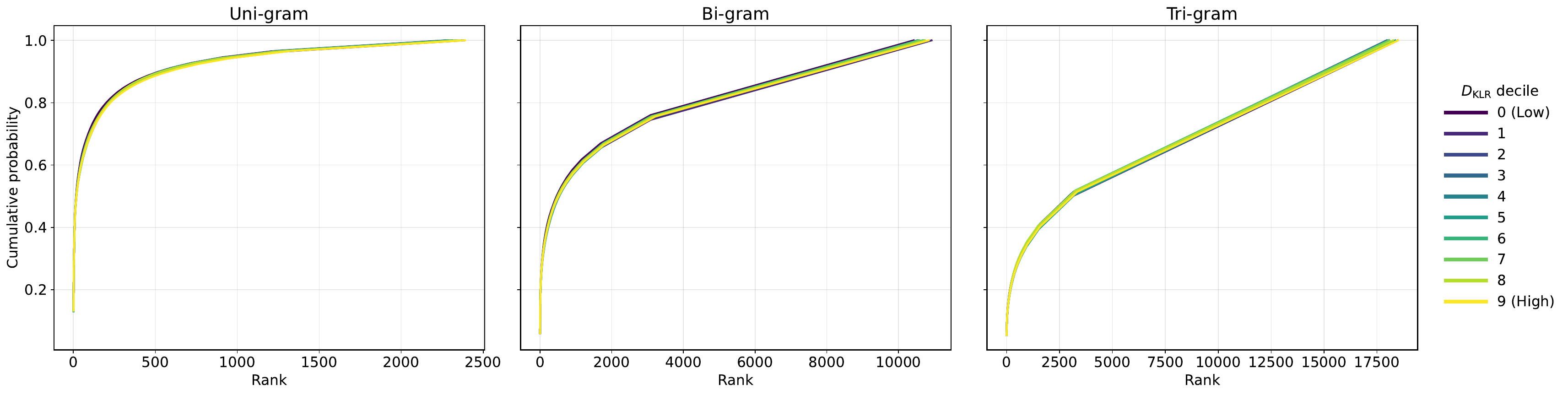}
    \caption{Cumulative N-gram probability across $D_\mathrm{KLR}(i)$ deciles}
    \label{fig:mscoco_img_dklr}
  \end{subfigure}
  \vspace{1em}
  \begin{subfigure}{0.49\linewidth}
    \includegraphics[width=\linewidth]{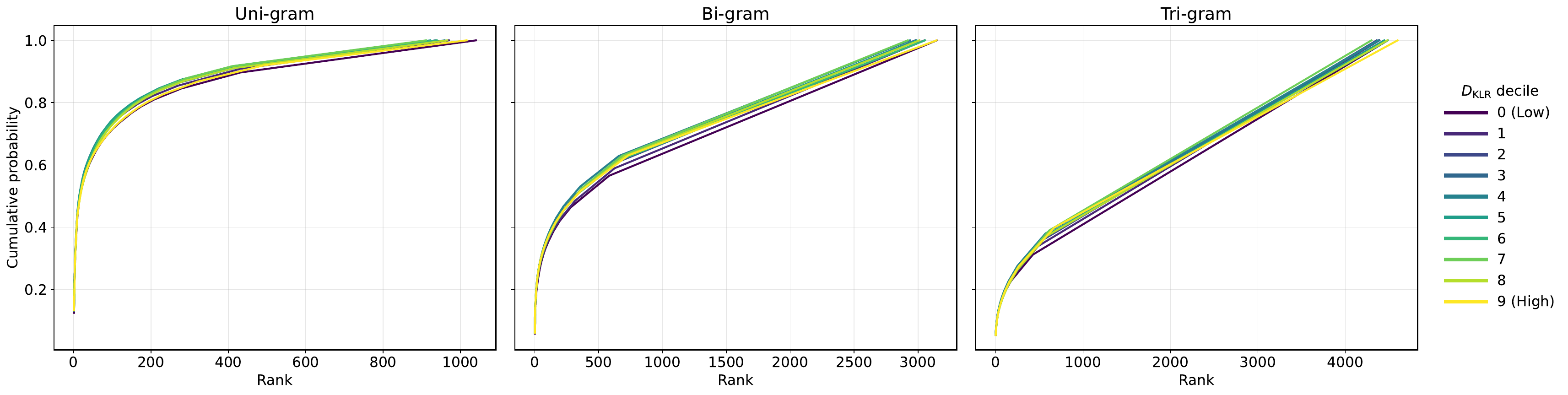}
    \caption{Cumulative N-gram probability across $D_\mathrm{KLR}(t)$ deciles}
    \label{fig:mscoco_txt_dklr}
  \end{subfigure}
  \vspace{1em}
  \begin{subfigure}{0.49\linewidth}
    \includegraphics[width=\linewidth]{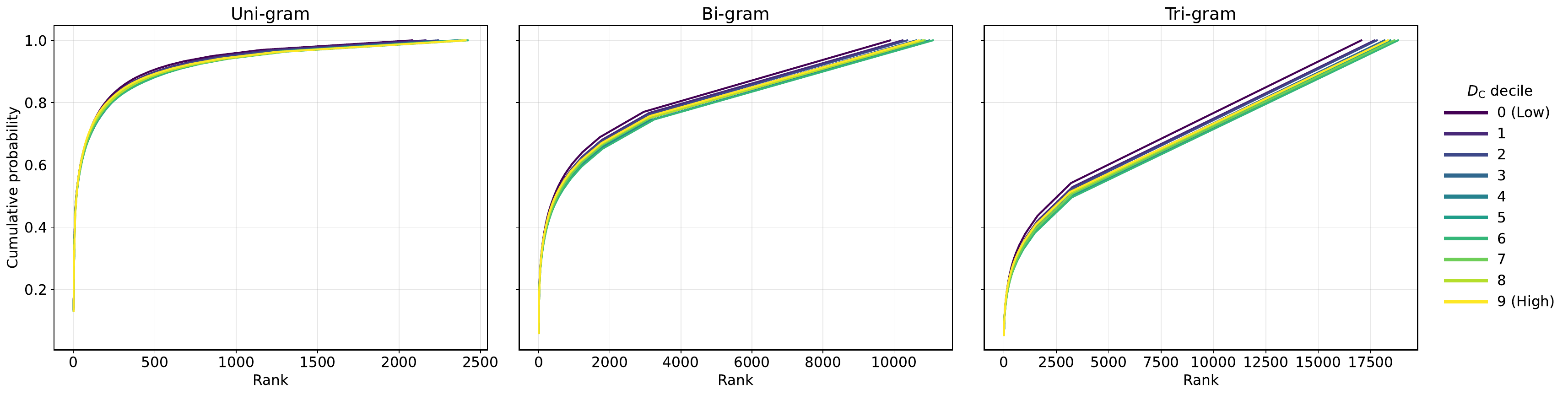}
    \caption{Cumulative N-gram probability across $D_\mathrm{C}(i)$ deciles}
    \label{fig:mscoco_img_dc}
  \end{subfigure}
  \vspace{1em}
  \begin{subfigure}{0.49\linewidth}
    \includegraphics[width=\linewidth]{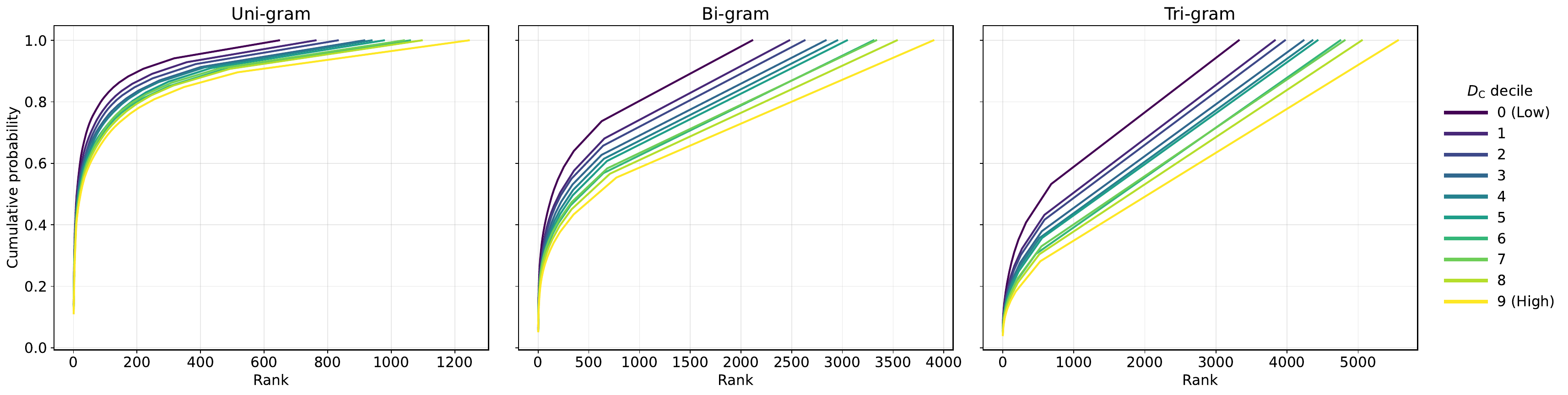}
    \caption{Cumulative N-gram probability across $D_\mathrm{C}(t)$ deciles}
    \label{fig:mscoco_txt_dc}
  \end{subfigure}
  \vspace{1em}
  \begin{subfigure}{0.49\linewidth}
    \includegraphics[width=\linewidth]{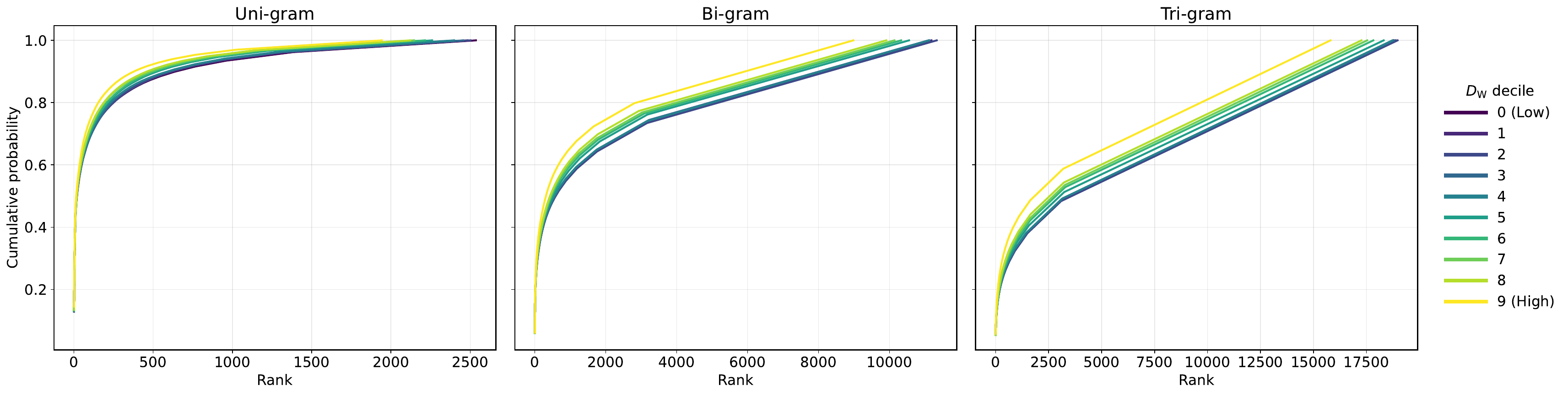}
    \caption{Cumulative N-gram probability across $D_\mathrm{W}(i)$ deciles}
    \label{fig:mscoco_img_dw}
  \end{subfigure}
  \vspace{1em}
  \begin{subfigure}{0.49\linewidth}
    \includegraphics[width=\linewidth]{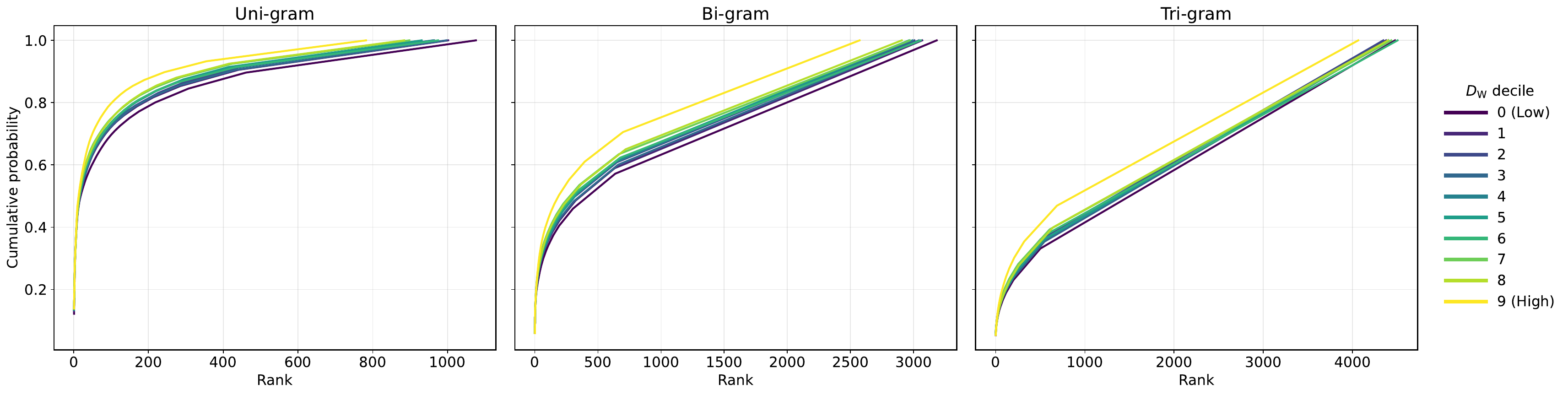}
    \caption{Cumulative N-gram probability across $D_\mathrm{W}(t)$ deciles}
    \label{fig:mscoco_txt_dw}
  \end{subfigure}
  \caption{
    N-gram probability coverage across KL deciles. 
    Each decile in \cref{fig:mscoco_img_dklr,fig:mscoco_img_dc,fig:mscoco_img_dw} is a group of captions corresponding to images $i$ which has the same level of $D_\mathrm{KLR}(i)$, $D_\mathrm{C}(i)$ and $D_\mathrm{W}(i)$.
    Each decile in \cref{fig:mscoco_txt_dklr,fig:mscoco_txt_dc,fig:mscoco_txt_dw} is a group of captions $t$ which has the same level of each KL divergence.
    }
  \label{fig:mscoco_ngram_other}
\end{figure*}

\section{SigLIP Results}
\label{sec:siglip}
In \cref{sec:apps,sec:apps_kl}, we empirically examined the applications of CLIP-like models as density ratio estimators using CLIP, which is commonly used in various tasks. In this section, we report the result of Importance-Weighted Learning and KL divergence estimation using ViT-B-16 SigLIP pretrained on the webli dataset~\cite{zhai2023sigmoid}.
 
\begin{figure*}[t]
  \centering
  \begin{subfigure}{0.33\linewidth}
    \includegraphics[width=\linewidth]{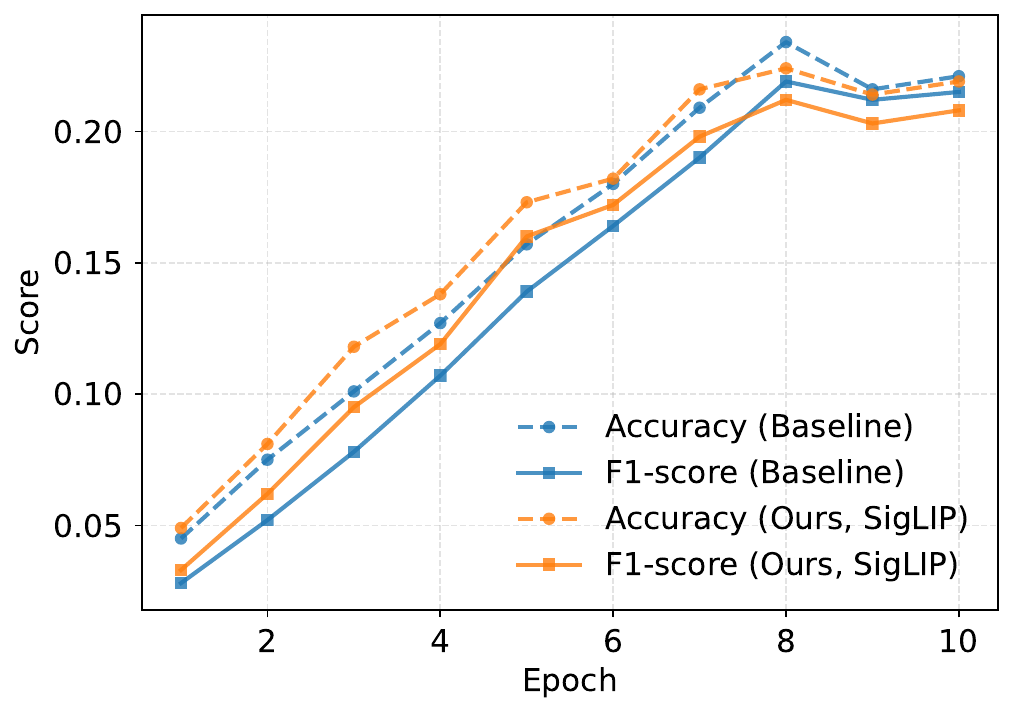}
    \caption{Food101}
  \end{subfigure}
  \hfill
  \begin{subfigure}{0.33\linewidth}
    \includegraphics[width=\linewidth]{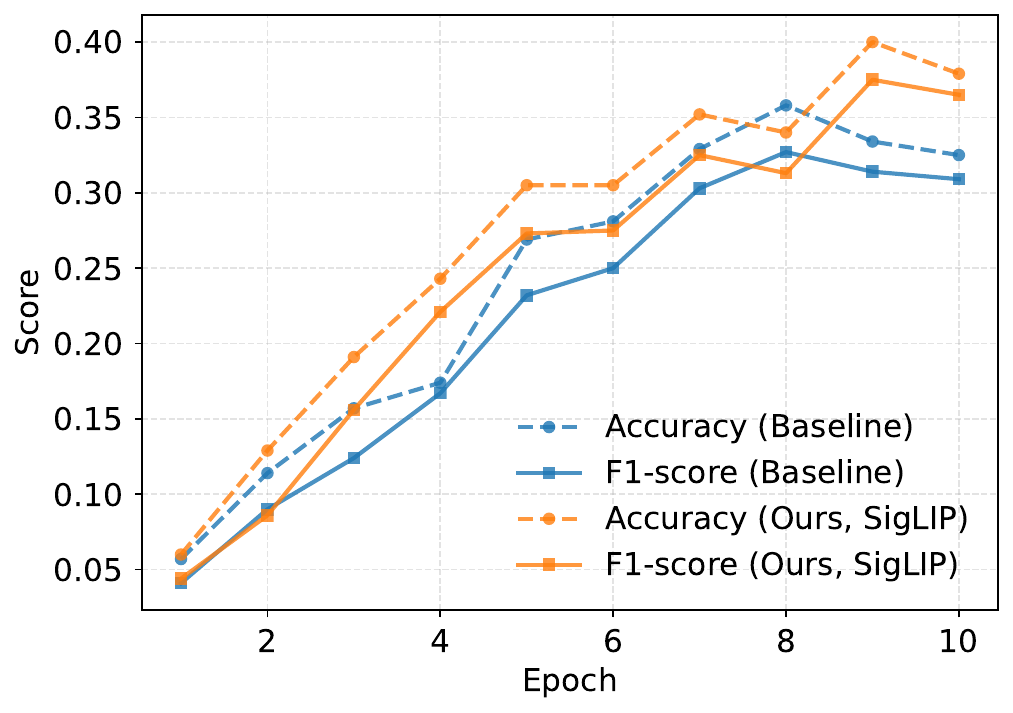}
    \caption{Oxford-IIIT Pet Dataset}
  \end{subfigure}
  \hfill
  \begin{subfigure}{0.33\linewidth}
    \includegraphics[width=\linewidth]{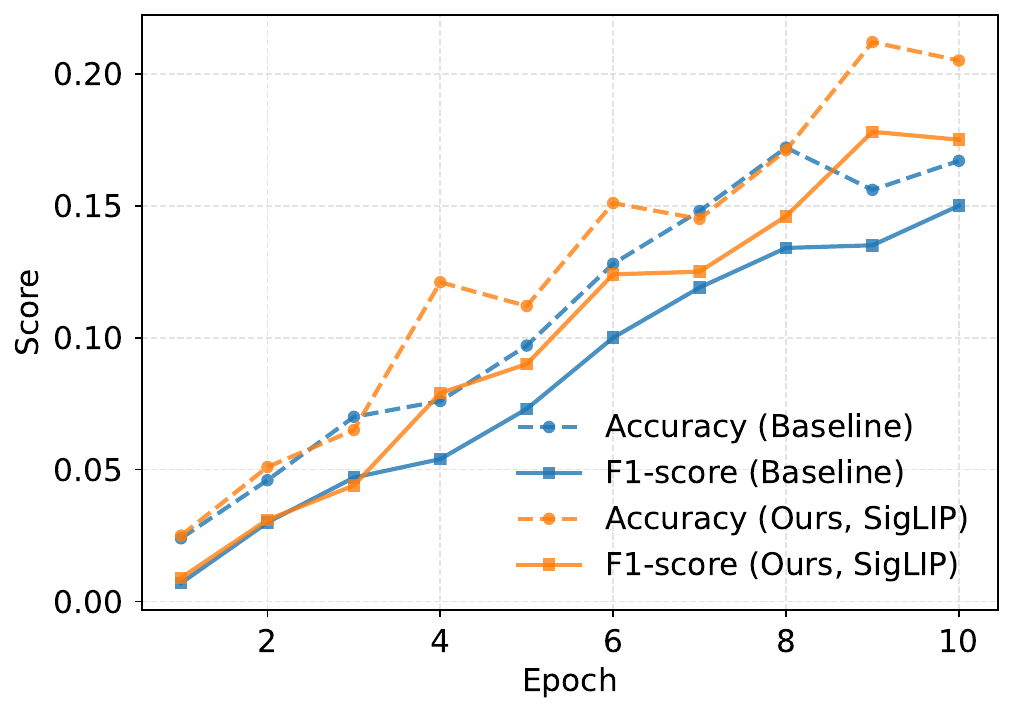}
    \caption{Flowers102}
  \end{subfigure}
  \caption{Zero-shot classification performance comparison between baseline and our IWL method using SigLIP similarity scores across three downstream datasets.}
  \label{fig:iwl_performance_siglip}
\end{figure*}

\paragraph{Importance-Weighted Learning}
\Cref{fig:iwl_performance_siglip} shows the downstream task performance for each domain compared with no importance weighting. Although the performance of IWL does not explicitly surpass training without importance weighting on Food101, our method mostly demonstrates higher downstream task performance than the baseline consistent with the results observed with CLIP.

\begin{figure*}[t]
  \centering
  \begin{subfigure}{\linewidth}
    \centering
    \imgcard{}{10.125}{red!80!black}\hfill
    \imgcard{}{10.125}{red!80!black}\hfill
    \imgcard{}{10.125}{red!80!black}\hfill
    \imgcard{}{10.124}{red!80!black}\hfill
    \imgcard{}{10.124}{red!80!black}\hfill
    \imgcard{}{10.124}{red!80!black}\\[0.5em]
    \imgcard{}{10.124}{red!80!black}\hfill
    \imgcard{}{10.124}{red!80!black}\hfill
    \imgcard{}{10.123}{red!80!black}\hfill
    \imgcard{}{10.122}{red!80!black}\hfill
    \imgcard{}{10.121}{red!80!black}\hfill
    \imgcard{}{10.121}{red!80!black}\\[0.5em]
    \imgcard{}{10.121}{red!80!black}\hfill
    \imgcard{}{10.119}{red!80!black}\hfill
    \imgcard{}{10.118}{red!80!black}\hfill
    \imgcard{}{10.118}{red!80!black}\hfill
    \imgcard{}{10.118}{red!80!black}\hfill
    \imgcard{}{10.115}{red!80!black}
    \caption{Top 18 images}
  \end{subfigure}
  \hfill
  \vspace{1em}
  \begin{subfigure}{\linewidth}
    \centering
    \imgcard{}{4.5523}{blue!80!black}\hfill
    \imgcard{}{4.5826}{blue!80!black}\hfill
    \imgcard{}{4.6101}{blue!80!black}\hfill
    \imgcard{}{4.6176}{blue!80!black}\hfill
    \imgcard{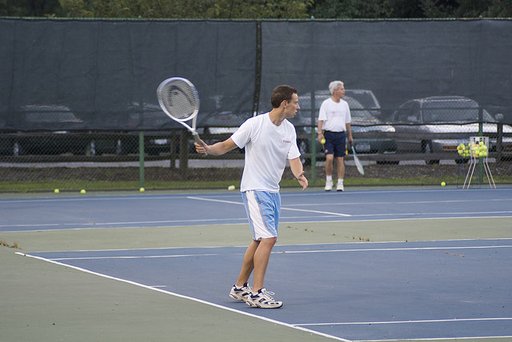}{4.6235}{blue!80!black}\hfill
    \imgcard{}{4.6365}{blue!80!black}\\[0.5em]
    \imgcard{}{4.6366}{blue!80!black}\hfill
    \imgcard{}{4.6437}{blue!80!black}\hfill
    \imgcard{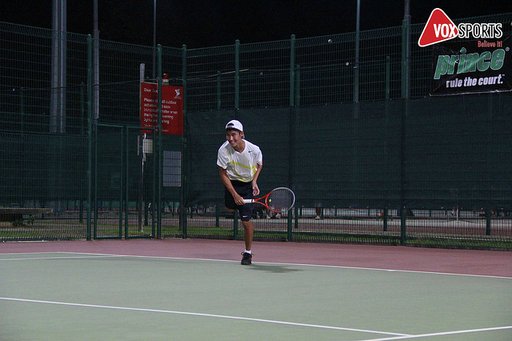}{4.6612}{blue!80!black}\hfill
    \imgcard{}{4.6696}{blue!80!black}\hfill
    \imgcard{}{4.6764}{blue!80!black}\hfill
    \imgcard{}{4.6847}{blue!80!black}\\[0.5em]
    \imgcard{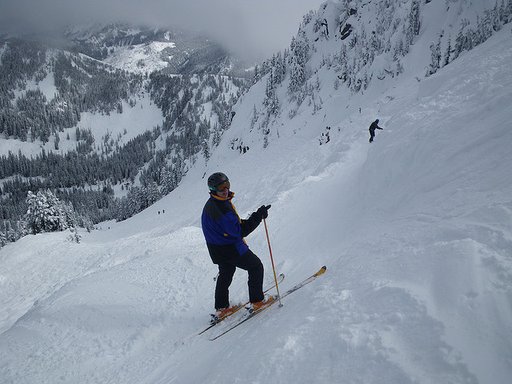}{4.6915}{blue!80!black}\hfill
    \imgcard{}{4.6975}{blue!80!black}\hfill
    \imgcard{}{4.703}{blue!80!black}\hfill
    \imgcard{}{4.7091}{blue!80!black}\hfill
    \imgcard{}{4.7124}{blue!80!black}\hfill
    \imgcard{}{4.7142}{blue!80!black}
    \caption{Bottom 18 images}
  \end{subfigure}
  \caption{
  Top and bottom image examples in MSCOCO captions ranked by $D_\mathrm{KL}$ through similarity score of SigLIP.
  }
  \label{fig:mscoco_image_samples_siglip_dkl}
\end{figure*}

\begin{figure*}[t]
  \centering
  \begin{subfigure}{\linewidth}
    \centering
    \imgcard{}{28.137}{red!80!black}\hfill
    \imgcard{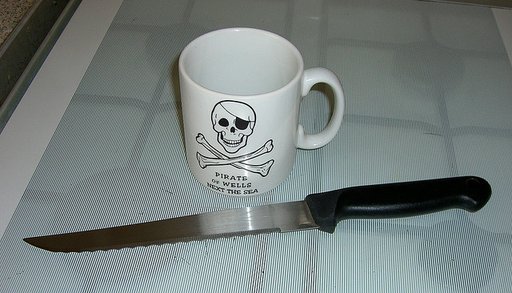}{27.946}{red!80!black}\hfill
    \imgcard{}{27.828}{red!80!black}\hfill
    \imgcard{}{27.751}{red!80!black}\hfill
    \imgcard{}{27.466}{red!80!black}\hfill
    \imgcard{}{27.138}{red!80!black}\\[0.5em]
    \imgcard{}{27.122}{red!80!black}\hfill
    \imgcard{}{27.104}{red!80!black}\hfill
    \imgcard{}{26.907}{red!80!black}\hfill
    \imgcard{}{26.881}{red!80!black}\hfill
    \imgcard{}{26.861}{red!80!black}\hfill
    \imgcard{}{26.829}{red!80!black}\\[0.5em]
    \imgcard{}{26.731}{red!80!black}\hfill
    \imgcard{}{26.649}{red!80!black}\hfill
    \imgcard{}{26.376}{red!80!black}\hfill
    \imgcard{}{26.18}{red!80!black}\hfill
    \imgcard{}{26.069}{red!80!black}\hfill
    \imgcard{}{25.848}{red!80!black}
    \caption{Top 18 images}
  \end{subfigure}
  \hfill
  \vspace{1em}
  \begin{subfigure}{\linewidth}
    \centering
    \imgcard{}{7.1854}{blue!80!black}\hfill
    \imgcard{}{8.1158}{blue!80!black}\hfill
    \imgcard{}{8.1368}{blue!80!black}\hfill
    \imgcard{}{8.9992}{blue!80!black}\hfill
    \imgcard{}{9.385}{blue!80!black}\hfill
    \imgcard{}{9.3991}{blue!80!black}\\[0.5em]
    \imgcard{}{9.5895}{blue!80!black}\hfill
    \imgcard{}{9.6297}{blue!80!black}\hfill
    \imgcard{}{9.6459}{blue!80!black}\hfill
    \imgcard{}{9.6512}{blue!80!black}\hfill
    \imgcard{}{9.6843}{blue!80!black}\hfill
    \imgcard{}{9.7217}{blue!80!black}\\[0.5em]
    \imgcard{}{9.9351}{blue!80!black}\hfill
    \imgcard{}{10.008}{blue!80!black}\hfill
    \imgcard{}{10.086}{blue!80!black}\hfill
    \imgcard{}{10.086}{blue!80!black}\hfill
    \imgcard{}{10.096}{blue!80!black}\hfill
    \imgcard{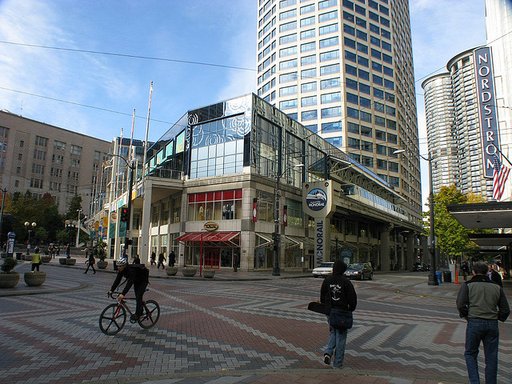}{10.129}{blue!80!black}
    \caption{Bottom 18 images}
  \end{subfigure}
  \caption{
  Top and bottom image examples in MSCOCO captions ranked by $D_\mathrm{KLR}$ through similarity score of SigLIP.
  }
  \label{fig:mscoco_image_samples_siglip_dklr}
\end{figure*}

\begin{figure*}[t]
  \centering
  \begin{subfigure}{\linewidth}
    \centering
    \imgcard{}{11467}{red!80!black}\hfill
    \imgcard{}{11286}{red!80!black}\hfill
    \imgcard{}{11149}{red!80!black}\hfill
    \imgcard{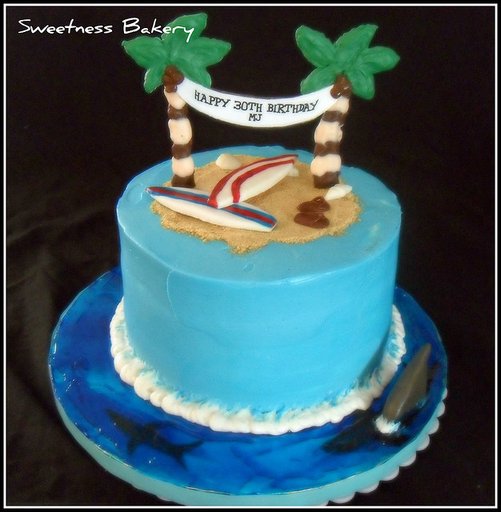}{11043}{red!80!black}\hfill
    \imgcard{}{10885}{red!80!black}\hfill
    \imgcard{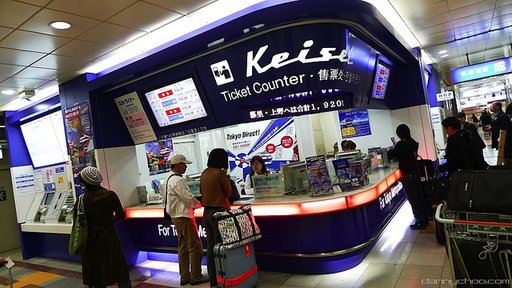}{10768}{red!80!black}\\[0.5em]
    \imgcard{}{10535}{red!80!black}\hfill
    \imgcard{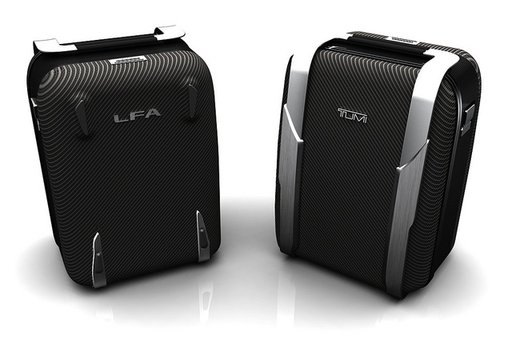}{10413}{red!80!black}\hfill
    \imgcard{}{10345}{red!80!black}\hfill
    \imgcard{}{10254}{red!80!black}\hfill
    \imgcard{}{10157}{red!80!black}\hfill
    \imgcard{}{10128}{red!80!black}\\[0.5em]
    \imgcard{}{10107}{red!80!black}\hfill
    \imgcard{}{10104}{red!80!black}\hfill
    \imgcard{}{10069}{red!80!black}\hfill
    \imgcard{}{10063}{red!80!black}\hfill
    \imgcard{}{10054}{red!80!black}\hfill
    \imgcard{}{10038}{red!80!black}
    \caption{Top 18 images}
  \end{subfigure}
  \hfill
  \vspace{1em}
  \begin{subfigure}{\linewidth}
    \centering
    \imgcard{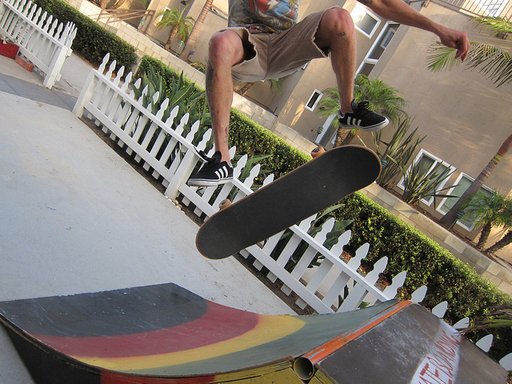}{4842.1}{blue!80!black}\hfill
    \imgcard{}{4939.7}{blue!80!black}\hfill
    \imgcard{}{4991.8}{blue!80!black}\hfill
    \imgcard{}{4997.6}{blue!80!black}\hfill
    \imgcard{}{5035.3}{blue!80!black}\hfill
    \imgcard{}{5059.2}{blue!80!black}\\[0.5em]
    \imgcard{}{5073.6}{blue!80!black}\hfill
    \imgcard{}{5080.9}{blue!80!black}\hfill
    \imgcard{}{5087}{blue!80!black}\hfill
    \imgcard{}{5087}{blue!80!black}\hfill
    \imgcard{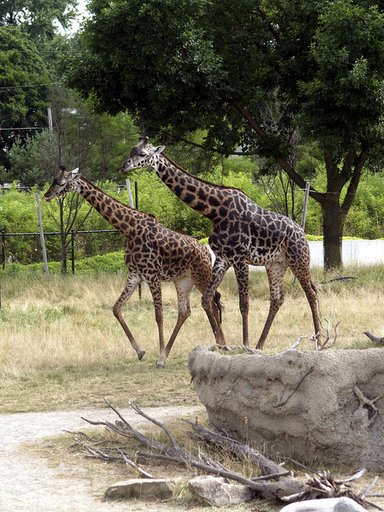}{5101.3}{blue!80!black}\hfill
    \imgcard{}{5118.9}{blue!80!black}\\[0.5em]
    \imgcard{}{5120.2}{blue!80!black}\hfill
    \imgcard{}{5142.4}{blue!80!black}\hfill
    \imgcard{}{5157.3}{blue!80!black}\hfill
    \imgcard{}{5163.7}{blue!80!black}\hfill
    \imgcard{}{5178.5}{blue!80!black}\hfill
    \imgcard{}{5187.7}{blue!80!black}
    \caption{Bottom 18 images}
  \end{subfigure}
  \caption{
  Top and bottom image examples in MSCOCO captions ranked by $D_\mathrm{C}$ through similarity score of SigLIP.
  }
  \label{fig:mscoco_image_samples_siglip_dc}
\end{figure*}

\begin{figure*}[t]
  \centering
  \begin{subfigure}{\linewidth}
    \centering
    \imgcard{}{9.9201e+05}{red!80!black}\hfill
    \imgcard{}{9.4566e+05}{red!80!black}\hfill
    \imgcard{}{9.4245e+05}{red!80!black}\hfill
    \imgcard{}{9.2688e+05}{red!80!black}\hfill
    \imgcard{}{9.2559e+05}{red!80!black}\hfill
    \imgcard{}{9.2119e+05}{red!80!black}\\[0.5em]
    \imgcard{}{9.1814e+05}{red!80!black}\hfill
    \imgcard{}{9.0534e+05}{red!80!black}\hfill
    \imgcard{}{8.9769e+05}{red!80!black}\hfill
    \imgcard{}{8.9319e+05}{red!80!black}\hfill
    \imgcard{}{8.9e+05}{red!80!black}\hfill
    \imgcard{}{8.8573e+05}{red!80!black}\\[0.5em]
    \imgcard{}{8.8511e+05}{red!80!black}\hfill
    \imgcard{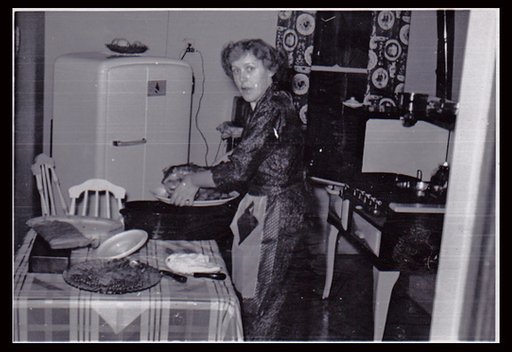}{8.8367e+05}{red!80!black}\hfill
    \imgcard{}{8.8146e+05}{red!80!black}\hfill
    \imgcard{}{8.8138e+05}{red!80!black}\hfill
    \imgcard{}{8.803e+05}{red!80!black}\hfill
    \imgcard{}{8.7785e+05}{red!80!black}
    \caption{Top 18 images}
  \end{subfigure}
  \hfill
  \vspace{1em}
  \begin{subfigure}{\linewidth}
    \centering
    \imgcard{}{1.8034e+05}{blue!80!black}\hfill
    \imgcard{}{2.1343e+05}{blue!80!black}\hfill
    \imgcard{}{2.1995e+05}{blue!80!black}\hfill
    \imgcard{}{2.2101e+05}{blue!80!black}\hfill
    \imgcard{}{2.2447e+05}{blue!80!black}\hfill
    \imgcard{}{2.2733e+05}{blue!80!black}\\[0.5em]
    \imgcard{}{2.2815e+05}{blue!80!black}\hfill
    \imgcard{}{2.3096e+05}{blue!80!black}\hfill
    \imgcard{}{2.325e+05}{blue!80!black}\hfill
    \imgcard{}{2.3379e+05}{blue!80!black}\hfill
    \imgcard{}{2.3395e+05}{blue!80!black}\hfill
    \imgcard{}{2.3622e+05}{blue!80!black}\\[0.5em]
    \imgcard{}{2.3663e+05}{blue!80!black}\hfill
    \imgcard{}{2.4005e+05}{blue!80!black}\hfill
    \imgcard{}{2.4187e+05}{blue!80!black}\hfill
    \imgcard{}{2.4212e+05}{blue!80!black}\hfill
    \imgcard{}{2.455e+05}{blue!80!black}\hfill
    \imgcard{}{2.4559e+05}{blue!80!black}
    \caption{Bottom 18 images}
  \end{subfigure}
  \caption{
  Top and bottom image examples in MSCOCO captions ranked by $D_\mathrm{W}$ through similarity score of SigLIP.
  }
  \label{fig:mscoco_image_samples_siglip_dw}
\end{figure*}

\paragraph{KL Divergence Estimation}
\Cref{fig:mscoco_image_samples_siglip_dkl,fig:mscoco_image_samples_siglip_dklr,fig:mscoco_image_samples_siglip_dc,fig:mscoco_image_samples_siglip_dw} show top and bottom 18 images ranked by the four KL-based metrics $D_\mathrm{KL}$, $D_\mathrm{KLR}$, $D_\mathrm{C}$, $D_\mathrm{W}$ and \cref{fig:mscoco_text_samples_siglip_dkl,fig:mscoco_text_samples_siglip_dklr,fig:mscoco_text_samples_siglip_dc,fig:mscoco_text_samples_siglip_dw} are the text samples.
\Cref{fig:mscoco_ngram_siglip} is the cumulative N-gram probability of captions grouped by the four metrics estimated by SigLIP. Consistent with CLIP's results, $D_\mathrm{KL}$ captures semantic diversity in terms of N-gram frequency.

\begin{figure*}[t]
  \centering
  \begin{subfigure}{\linewidth}
    \centering
    \textcard{Round glass table full of potted plants and description cards}{8.5172}{red!80!black}\hfill
    \textcard{A sheet cake with a lake scene displays the words, "What is the Pickup Line Today?"}{8.5172}{red!80!black}\hfill
    \textcard{A heart shaped, glass container filled with origami birds.}{8.5172}{red!80!black}\hfill
    \textcard{A clear heart shaped jar filled with many colors of paper cranes.}{8.5172}{red!80!black}\hfill
    \textcard{A sign that says "Abbey Road NW8 City of Westminster".}{8.5172}{red!80!black}\hfill
    \textcard{A girl in a blue bowtie standing next to a pink Princess Parking Only sign.}{8.5172}{red!80!black}\\[0.5em]
    \textcard{A woman standing next to a princess parking only sign.}{8.5172}{red!80!black}\hfill
    \textcard{Elvis impersonator sitting atop a metal sculpture of a bull.}{8.5172}{red!80!black}\hfill
    \textcard{A car covered in dolls and doll parts with people in the background.}{8.5172}{red!80!black}\hfill
    \textcard{a white mug  showing pirate skull and bones and a large knife on a counter top.}{8.5172}{red!80!black}\hfill
    \textcard{Two almost identical photos with some minor cropping show two young girls standing on the coast with people in the water in the background as a plane flies overhead}{8.5172}{red!80!black}\hfill
    \textcard{A smiling girl standing beside a sign that says "Princess Parking Only".}{8.5172}{red!80!black}\\[0.5em]
    \textcard{a home bar with different drink ingredients under a large decorative sign saying, "PUB".}{8.5172}{red!80!black}\hfill
    \textcard{A person sits on the floor beside flip flops and a teddy bear.}{8.5172}{red!80!black}\hfill
    \textcard{A white frosted donut with sprinkles and a jump rope with leather roping next to it.}{8.5172}{red!80!black}\hfill
    \textcard{A mirror image of a red haired doll clock.}{8.5172}{red!80!black}\hfill
    \textcard{a woman selling jewelry laid on a blanket on the sidewalk}{8.5172}{red!80!black}\hfill
    \textcard{An old truck, painted over blue in the desert}{8.5172}{red!80!black}
    \caption{Top 18 texts}
  \end{subfigure}
  \vspace{1em}
  \begin{subfigure}{\linewidth}
    \centering
    \textcard{An individual is taken in this very picture. 
    }{1.7682}{blue!80!black}\hfill
    \textcard{An individual is taken in this very picture. 
    }{1.7682}{blue!80!black}\hfill
    \textcard{A being is doing something as of right now that is splendid. 
    }{1.8934}{blue!80!black}\hfill
    \textcard{There is no image to describe for this question.}{1.9816}{blue!80!black}\hfill
    \textcard{A thing is in the outline and it shows up like something 
    }{2.0293}{blue!80!black}\hfill
    \textcard{An individual is in the open view in the picture. 
    }{2.1309}{blue!80!black}\\[0.5em]
    \textcard{There is no image here to provide a caption for.}{2.1893}{blue!80!black}\hfill
    \textcard{There is no image here to provide a caption for.}{2.1893}{blue!80!black}\hfill
    \textcard{There is no image here to provide a caption for.}{2.1893}{blue!80!black}\hfill
    \textcard{There is no image here to provide a caption for.}{2.1893}{blue!80!black}\hfill
    \textcard{A picture of a person on a skateboard.}{2.1898}{blue!80!black}\hfill
    \textcard{no image again there are a lot of these lately}{2.225}{blue!80!black}\\[0.5em]
    \textcard{I am unable to see the image above.}{2.256}{blue!80!black}\hfill
    \textcard{unable to see this image in this particular hit}{2.2638}{blue!80!black}\hfill
    \textcard{I am unable to see an image above.}{2.3005}{blue!80!black}\hfill
    \textcard{I am unable to see an image above.}{2.3005}{blue!80!black}\hfill
    \textcard{I am unable to see an image above.}{2.3005}{blue!80!black}\hfill
    \textcard{I am unable to see an image above.}{2.3005}{blue!80!black}
    \caption{Bottom 18 texts}
  \end{subfigure}
  \caption{Top and bottom captions in MSCOCO captions ranked by $D_\mathrm{KL}$ through similarity score of SigLIP.}
  \label{fig:mscoco_text_samples_siglip_dkl}
\end{figure*}

\begin{figure*}[t]
  \centering
  \begin{subfigure}{\linewidth}
    \centering
    \textcard{a street sign for Library Way and Madison Avenue above a one way arrow}{34.339}{red!80!black}\hfill
    \textcard{A class picture of the Goodmayes Boys' School in April of 1929.}{33.828}{red!80!black}\hfill
    \textcard{A sign that says "Abbey Road NW8 City of Westminster".}{32.846}{red!80!black}\hfill
    \textcard{Street sign for Anza and 21st Streets in front of building.}{32.458}{red!80!black}\hfill
    \textcard{A group photo of men and boys from the Goodmayes Boys School dated April 1929.}{32.024}{red!80!black}\hfill
    \textcard{A sheet cake with a lake scene displays the words, "What is the Pickup Line Today?"}{31.716}{red!80!black}\\[0.5em]
    \textcard{A view of the street signs "W 122 St.", "Seminary Row", and "Broadway" in front of an old red brick building. }{31.507}{red!80!black}\hfill
    \textcard{Two almost identical photos with some minor cropping show two young girls standing on the coast with people in the water in the background as a plane flies overhead}{31.498}{red!80!black}\hfill
    \textcard{Two cranes behind a stop sign and a street sign indicating Fort York Blvd.}{31.297}{red!80!black}\hfill
    \textcard{Outside view of the MGM Grand in Las Vegas with people sitting and walking.}{30.966}{red!80!black}\hfill
    \textcard{A key bank sign with a clock on a building}{30.957}{red!80!black}\hfill
    \textcard{A waitress and restaurant owner showing off a pizza in front of a mural of Venice.}{30.783}{red!80!black}\\[0.5em]
    \textcard{Street sign at the corner of Ho'ohu and Pe'e roads on Kauai}{30.273}{red!80!black}\hfill
    \textcard{Two planes fly over a bridge in Sydney, Australia, with the Sydney Opera House in the background.}{30.253}{red!80!black}\hfill
    \textcard{A street sign at an intersection of Library Way and Madison Avenue}{30.237}{red!80!black}\hfill
    \textcard{A bus parked near an Apple sign during the night.}{30.107}{red!80!black}\hfill
    \textcard{An green and white overhead street sign on Interstate 278 for Queens and Bronx, showing a truck restriction.}{29.965}{red!80!black}\hfill
    \textcard{Street signs for the intersection of Anza and 21st Avenue in front of a large building}{29.895}{red!80!black}
    \caption{Top 18 texts}
  \end{subfigure}
  \vspace{1em}
  \begin{subfigure}{\linewidth}
    \centering
    \textcard{There is no image to describe for this question.}{1.7863}{blue!80!black}\hfill
    \textcard{A being is doing something as of right now that is splendid. 
    }{1.9594}{blue!80!black}\hfill
    \textcard{An individual is taken in this very picture. 
    }{1.9638}{blue!80!black}\hfill
    \textcard{An individual is taken in this very picture. 
    }{1.9638}{blue!80!black}\hfill
    \textcard{A thing is in the outline and it shows up like something 
    }{2.0022}{blue!80!black}\hfill
    \textcard{There is no image here to provide a caption for.}{2.0878}{blue!80!black}\\[0.5em]
    \textcard{There is no image here to provide a caption for.}{2.0878}{blue!80!black}\hfill
    \textcard{There is no image here to provide a caption for.}{2.0878}{blue!80!black}\hfill
    \textcard{There is no image here to provide a caption for.}{2.0878}{blue!80!black}\hfill
    \textcard{I am unable to see the image above.}{2.1347}{blue!80!black}\hfill
    \textcard{I do not know what this is supposed to be..}{2.1526}{blue!80!black}\hfill
    \textcard{There is no image to be reviewed on this hit.}{2.1852}{blue!80!black}\\[0.5em]
    \textcard{no image again there are a lot of these lately}{2.1932}{blue!80!black}\hfill
    \textcard{I am unable to see an image above.}{2.2464}{blue!80!black}\hfill
    \textcard{I am unable to see an image above.}{2.2464}{blue!80!black}\hfill
    \textcard{I am unable to see an image above.}{2.2464}{blue!80!black}\hfill
    \textcard{I am unable to see an image above.}{2.2464}{blue!80!black}\hfill
    \textcard{I am unable to see an image above.}{2.2464}{blue!80!black}
    \caption{Bottom 18 texts}
  \end{subfigure}
  \caption{Top and bottom captions in MSCOCO captions ranked by $D_\mathrm{KLR}$ through similarity score of SigLIP.}
  \label{fig:mscoco_text_samples_siglip_dklr}
\end{figure*}

\begin{figure*}[t]
  \centering
  \begin{subfigure}{\linewidth}
    \centering
    \textcard{Side by side view of two oval plates, one with fork, with chicken salad sandwiches and rosy new potatoes, by an open and an unopened bottle of lager, a pepper mill, paper towel roll, basket behind.}{17225}{red!80!black}\hfill
    \textcard{Three rectangular bowls with food; Big bowl has nine meat and sesame seed patties with brown sauce, next to it, a bowl of shredded cabbage and carrots with yogurt dollop atop, and behind that is a bowl of cut broccoli and tomatoes with seasoning.}{16587}{red!80!black}\hfill
    \textcard{An green and white overhead street sign on Interstate 278 for Queens and Bronx, showing a truck restriction.}{15674}{red!80!black}\hfill
    \textcard{Two teddy bears a pink one, and a tan one. The tan bear is wearing a pink shirt that says the harvey girls.}{15485}{red!80!black}\hfill
    \textcard{Two apples, a bowl of food with berries and a pitcher and spoon on a purple place mat.}{15061}{red!80!black}\hfill
    \textcard{a cupcake with pink icing in a pink paper cupcake holder next to a spoon.}{15027}{red!80!black}\\[0.5em]
    \textcard{Several children watch while a girl in a pink sweatshirt plays with a Wii remote while another person with a little girl on their lap snuggle in a chair in the background.}{15013}{red!80!black}\hfill
    \textcard{A woman wearing a blue t-shirt while looking at her cell phone and sitting on a bench next to a bright pink wall.}{14996}{red!80!black}\hfill
    \textcard{Two women are squatting down and petting brown and white long haired goats. }{14939}{red!80!black}\hfill
    \textcard{Four men are standing together behind a group of red chairs.}{14833}{red!80!black}\hfill
    \textcard{A styrofoam plate with shredded chicken and a dish of broccoli with cheese sauce.}{14825}{red!80!black}\hfill
    \textcard{Two very small yellow bananas on top of a wooden table next to a Malaysian coin. }{14822}{red!80!black}\\[0.5em]
    \textcard{A spoon, a large carrot, a medium carrot and a small carrot, on blue-green speckled surface}{14819}{red!80!black}\hfill
    \textcard{A street sign reading Cambridge st and Norfolk st's crossing lies broken on the ground. }{14714}{red!80!black}\hfill
    \textcard{A fork, a pork chop, green beans, an egg and parsley on a white plate.}{14679}{red!80!black}\hfill
    \textcard{Four stuffed animals, a leopard and three teddy bears, in a row sitting on a stone ledge with grass and trees behind.}{14645}{red!80!black}\hfill
    \textcard{A table with two black trays of food that include raspberries and broccoli.}{14492}{red!80!black}\hfill
    \textcard{School boys sit cross legged in front of a chalkboard sign in a vintage black and white photo.}{14481}{red!80!black}
    \caption{Top 18 texts}
  \end{subfigure}
  \vspace{1em}
  \begin{subfigure}{\linewidth}
    \centering
    \textcard{A train going down the train track. }{4099.8}{blue!80!black}\hfill
    \textcard{a train sitting on the tracks at a station}{4258.1}{blue!80!black}\hfill
    \textcard{A man who is performing a trick on a skateboard.}{4303.7}{blue!80!black}\hfill
    \textcard{A bus stopped on the side of the road.}{4330.1}{blue!80!black}\hfill
    \textcard{A train with no cars sitting on the train tracks.}{4341.5}{blue!80!black}\hfill
    \textcard{a train on a track pulling into a station }{4352.5}{blue!80!black}\\[0.5em]
    \textcard{a man going up a ramp on a skateboard}{4374.2}{blue!80!black}\hfill
    \textcard{A train traveling down tracks near a train station.}{4382}{blue!80!black}\hfill
    \textcard{a group of adults playing soccer on a field}{4384.1}{blue!80!black}\hfill
    \textcard{A train engine is on the railroad tracks.}{4396.2}{blue!80!black}\hfill
    \textcard{A group of kids playing a game of baseball.}{4411.4}{blue!80!black}\hfill
    \textcard{A train on the tracks near a train station }{4412.1}{blue!80!black}\\[0.5em]
    \textcard{A train on the train tracks in the daytime.}{4414.9}{blue!80!black}\hfill
    \textcard{This is a picture of a train coming into the station.}{4416.9}{blue!80!black}\hfill
    \textcard{A train is moving through a train station.}{4423.6}{blue!80!black}\hfill
    \textcard{a young boy is playing soccer on a field }{4425.7}{blue!80!black}\hfill
    \textcard{A man on a skateboard doing a trick.}{4435.2}{blue!80!black}\hfill
    \textcard{A baseball player preparing to bat during a game.}{4436.2}{blue!80!black}
    \caption{Bottom 18 texts}
  \end{subfigure}
  \caption{Top and bottom captions in MSCOCO captions ranked by $D_\mathrm{C}$ through similarity score of SigLIP.}
  \label{fig:mscoco_text_samples_siglip_dc}
\end{figure*}

\begin{figure*}[t]
  \centering
  \begin{subfigure}{\linewidth}
    \centering
    \textcard{Buses parked on a road outside a large bus station. }{2.5771e+05}{red!80!black}\hfill
    \textcard{A photo of a busy intersection with a bus, several cars and a clock tower.}{2.4974e+05}{red!80!black}\hfill
    \textcard{An intersection with antique cars and a bus at it.}{2.4973e+05}{red!80!black}\hfill
    \textcard{A city road with buses and people on the sidewalk}{2.4972e+05}{red!80!black}\hfill
    \textcard{a car and a bus at an intersection in front of a historical looking building}{2.477e+05}{red!80!black}\hfill
    \textcard{A street filled with lots of cars next to a boat.}{2.4403e+05}{red!80!black}\\[0.5em]
    \textcard{a group of ball players playing on a field in front of an audience}{2.4109e+05}{red!80!black}\hfill
    \textcard{People walking on the sidewalk of a city with a tour bus in the background.}{2.4098e+05}{red!80!black}\hfill
    \textcard{People crossing the street near a parked sightseeing bus.}{2.3821e+05}{red!80!black}\hfill
    \textcard{A bus and other cars driving down a multi-laned street .}{2.3718e+05}{red!80!black}\hfill
    \textcard{parked cars, motorcycles and buses on a cobblestone parking lot next to a street.}{2.3707e+05}{red!80!black}\hfill
    \textcard{Two buses in city with other cars and trees. }{2.3516e+05}{red!80!black}\\[0.5em]
    \textcard{A busy street with cars, a motorcycle and a passenger bus}{2.32e+05}{red!80!black}\hfill
    \textcard{A large bus and other vehicles on a busy street.}{2.3187e+05}{red!80!black}\hfill
    \textcard{A city road with buses and bus park}{2.3148e+05}{red!80!black}\hfill
    \textcard{An Asian city square, with people, buses, and a McDonald's.}{2.3134e+05}{red!80!black}\hfill
    \textcard{Cars and a bus stopped for a train at a crossing.}{2.3125e+05}{red!80!black}\hfill
    \textcard{Several men are playing sports on a field near some trees, wall, bus, and several buildings in the background.}{2.288e+05}{red!80!black}
    \caption{Top 18 texts}
  \end{subfigure}
  \vspace{1em}
  \begin{subfigure}{\linewidth}
    \centering
    \textcard{A picture of a person on a skateboard.}{19561}{blue!80!black}\hfill
    \textcard{There is a picture of an outside territory.
    }{23716}{blue!80!black}\hfill
    \textcard{There is no image to describe for this question.}{25238}{blue!80!black}\hfill
    \textcard{An individual is taken in this very picture. 
    }{26023}{blue!80!black}\hfill
    \textcard{An individual is taken in this very picture. 
    }{26023}{blue!80!black}\hfill
    \textcard{An individual is in the open view in the picture. 
    }{26581}{blue!80!black}\\[0.5em]
    \textcard{A statue of two people sitting on a bench.}{26713}{blue!80!black}\hfill
    \textcard{A sign on a sidewalk has a teddy bear on it.}{27071}{blue!80!black}\hfill
    \textcard{A person sits on top of a bench.}{27396}{blue!80!black}\hfill
    \textcard{a black and white photo of a train }{28376}{blue!80!black}\hfill
    \textcard{a double telescope for seeing things that are far away }{28425}{blue!80!black}\hfill
    \textcard{THERE IS A GAME A BASKETBALL GAME GOING ON}{28560}{blue!80!black}\\[0.5em]
    \textcard{Somebody is in the photograph not certain who that individual is. 
    }{28913}{blue!80!black}\hfill
    \textcard{The travelers are all traveling with their best luggage.}{28980}{blue!80!black}\hfill
    \textcard{Their is a hadron and their  in this lot,}{29067}{blue!80!black}\hfill
    \textcard{Individual is doing something at the moment that is intriguing.
    }{29491}{blue!80!black}\hfill
    \textcard{Individual is doing something at the moment that is intriguing.}{29491}{blue!80!black}\hfill
    \textcard{The hand is holding an iPhone for the picture.}{29611}{blue!80!black}
    \caption{Bottom 18 texts}
  \end{subfigure}
  \caption{Top and bottom captions in MSCOCO captions ranked by $D_\mathrm{W}$ through similarity score of SigLIP.}
  \label{fig:mscoco_text_samples_siglip_dw}
\end{figure*}

\begin{figure*}[t]
  \centering
  \begin{subfigure}{0.49\linewidth}
    \includegraphics[width=\linewidth]{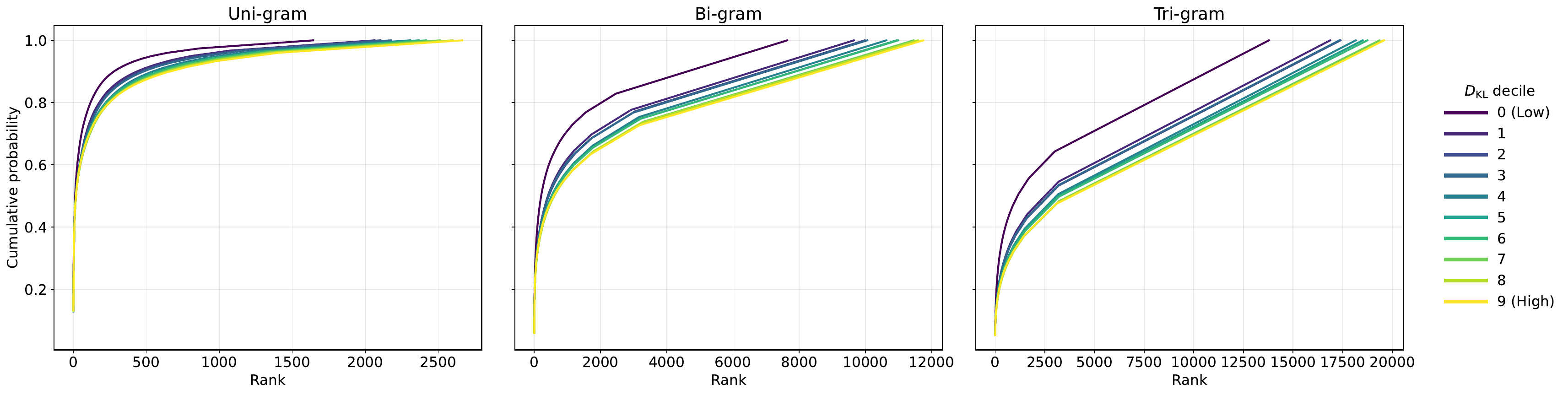}
    \caption{Cumulative N-gram probability across $D_\mathrm{KL}(i)$ deciles}
    \label{fig:mscoco_siglip_img_dkl}
  \end{subfigure}
  \vspace{1em}
  \begin{subfigure}{0.49\linewidth}
    \includegraphics[width=\linewidth]{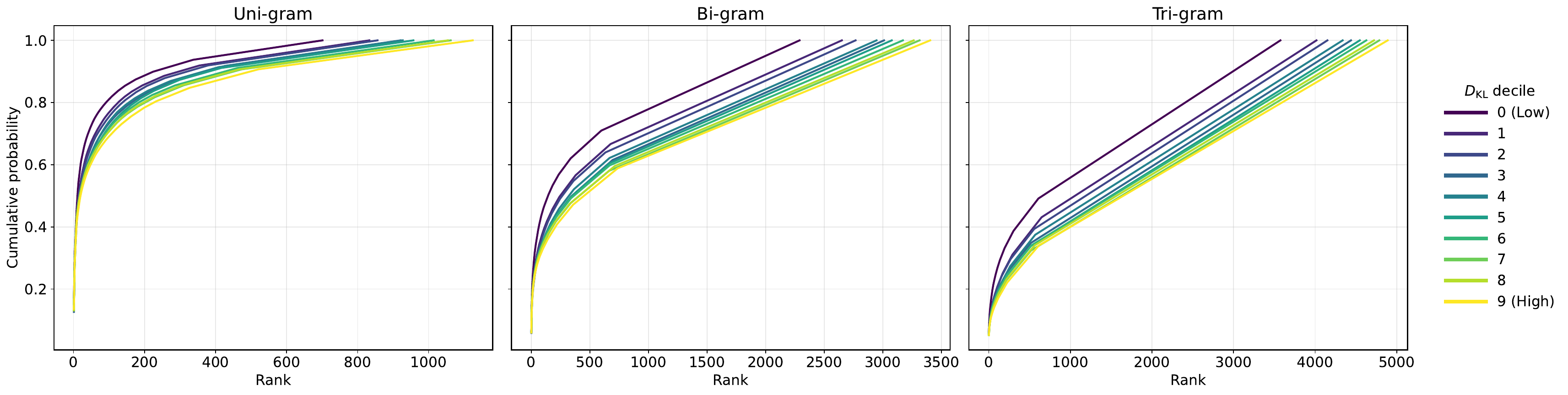}
    \caption{Cumulative N-gram probability across $D_\mathrm{KL}(t)$ deciles}
    \label{fig:mscoco_siglip_txt_dkl}
  \end{subfigure}
  \vspace{1em}
  \begin{subfigure}{0.49\linewidth}
    \includegraphics[width=\linewidth]{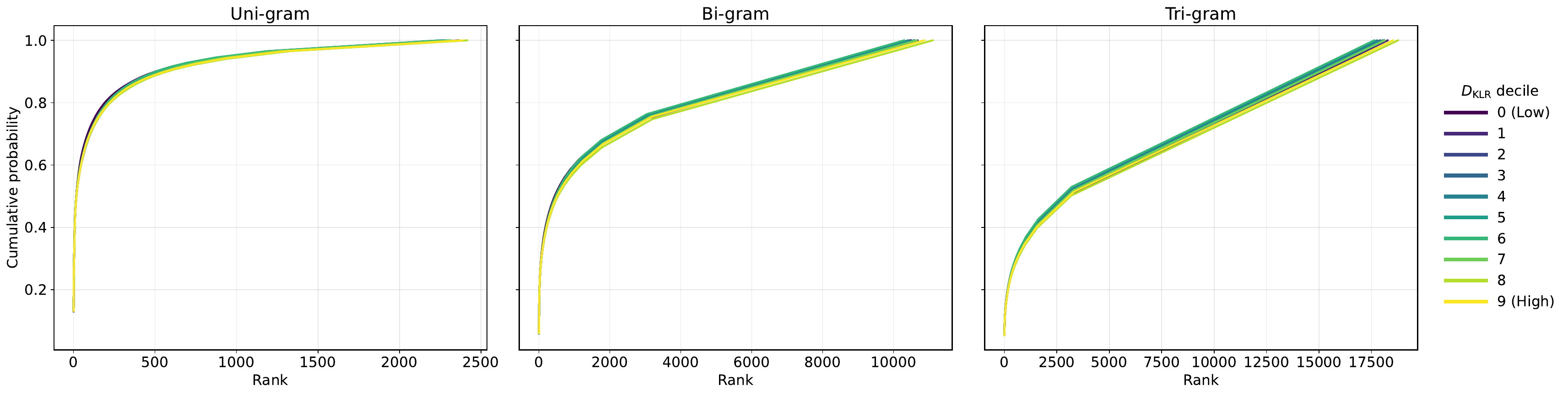}
    \caption{Cumulative N-gram probability across $D_\mathrm{KLR}(i)$ deciles}
    \label{fig:mscoco_siglip_img_dklr}
  \end{subfigure}
  \vspace{1em}
  \begin{subfigure}{0.49\linewidth}
    \includegraphics[width=\linewidth]{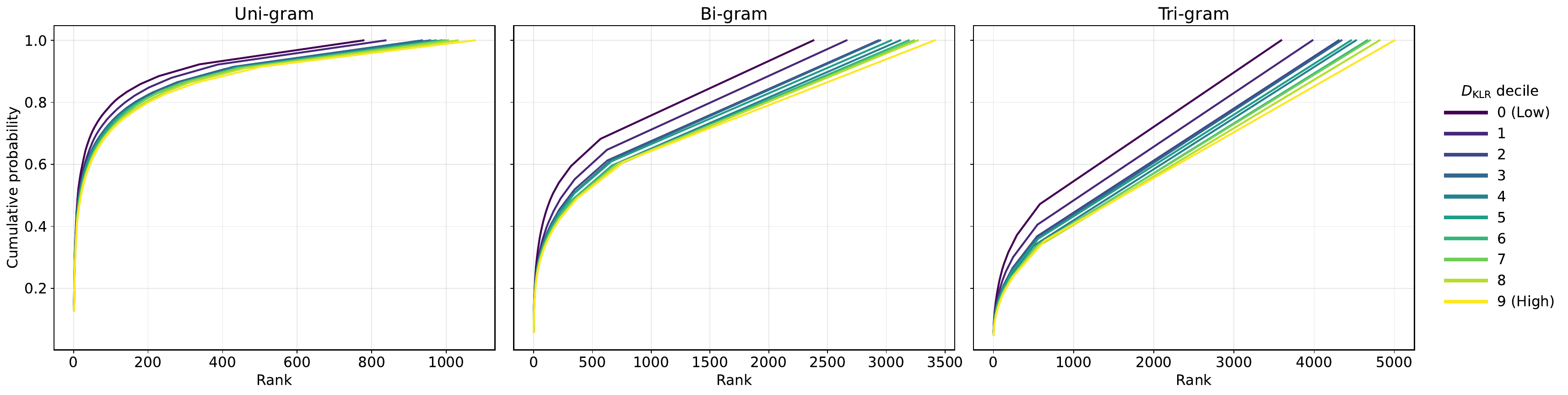}
    \caption{Cumulative N-gram probability across $D_\mathrm{KLR}(t)$ deciles}
    \label{fig:mscoco_siglip_txt_dklr}
  \end{subfigure}
  \vspace{1em}
  \begin{subfigure}{0.49\linewidth}
    \includegraphics[width=\linewidth]{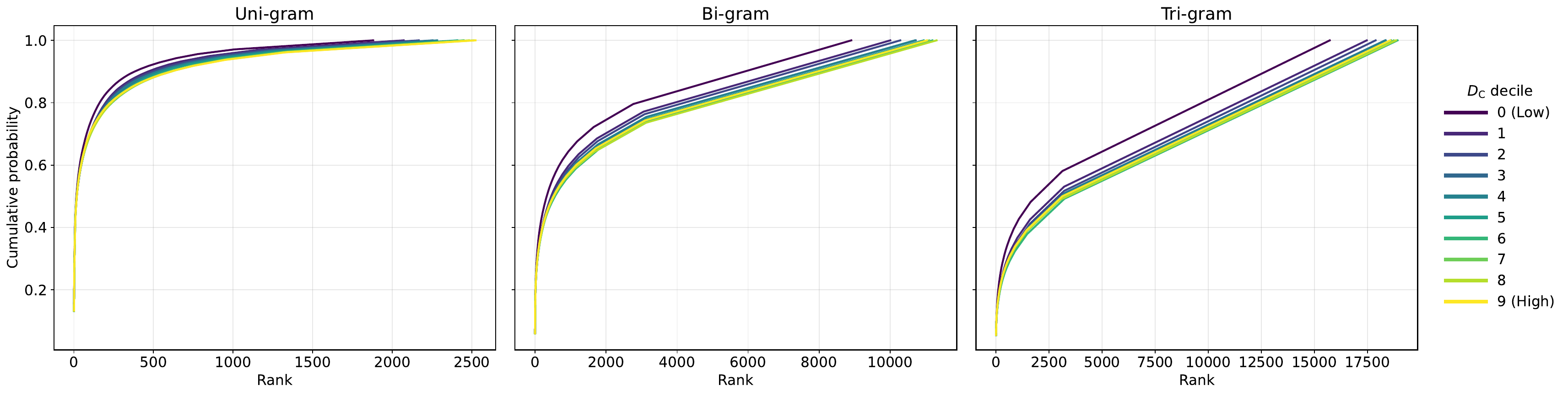}
    \caption{Cumulative N-gram probability across $D_\mathrm{C}(i)$ deciles}
    \label{fig:mscoco_siglip_img_dc}
  \end{subfigure}
  \vspace{1em}
  \begin{subfigure}{0.49\linewidth}
    \includegraphics[width=\linewidth]{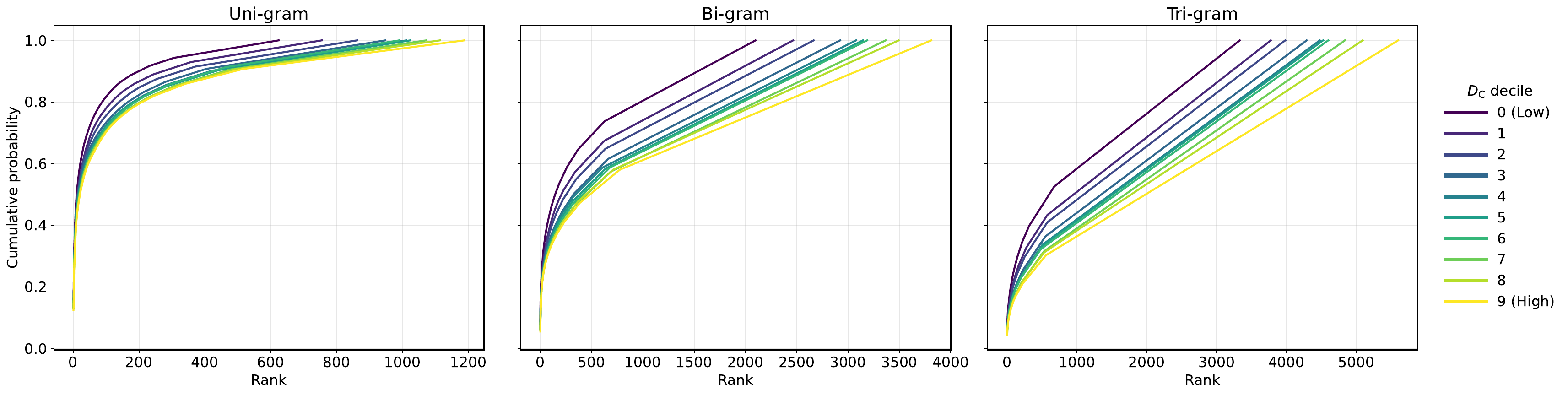}
    \caption{Cumulative N-gram probability across $D_\mathrm{C}(t)$ deciles}
    \label{fig:mscoco_siglip_txt_dc}
  \end{subfigure}
  \vspace{1em}
  \begin{subfigure}{0.49\linewidth}
    \includegraphics[width=\linewidth]{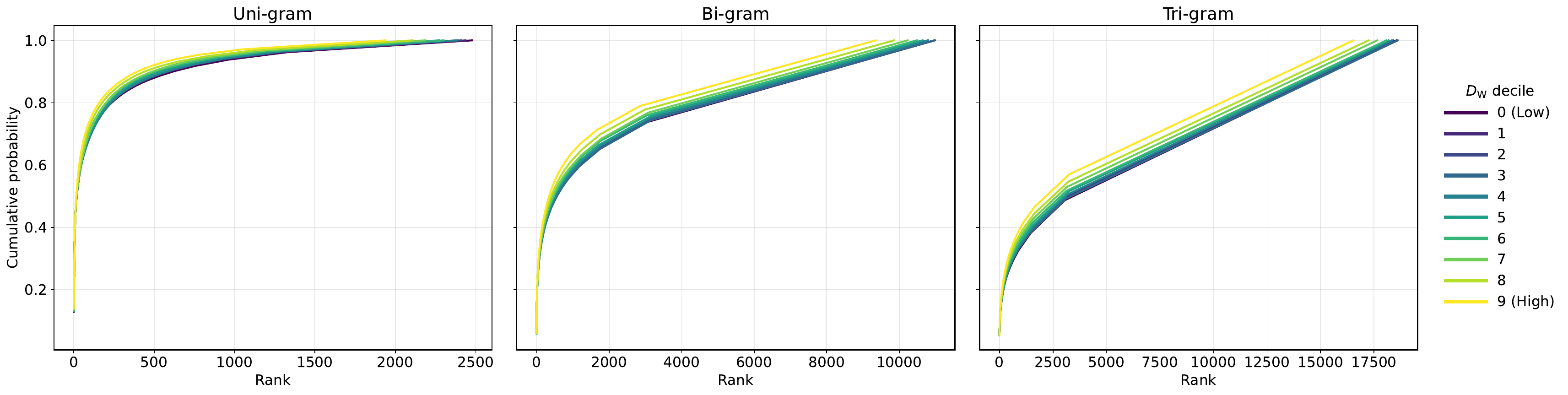}
    \caption{Cumulative N-gram probability across $D_\mathrm{W}(i)$ deciles}
    \label{fig:mscoco_siglip_img_dw}
  \end{subfigure}
  \vspace{1em}
  \begin{subfigure}{0.49\linewidth}
    \includegraphics[width=\linewidth]{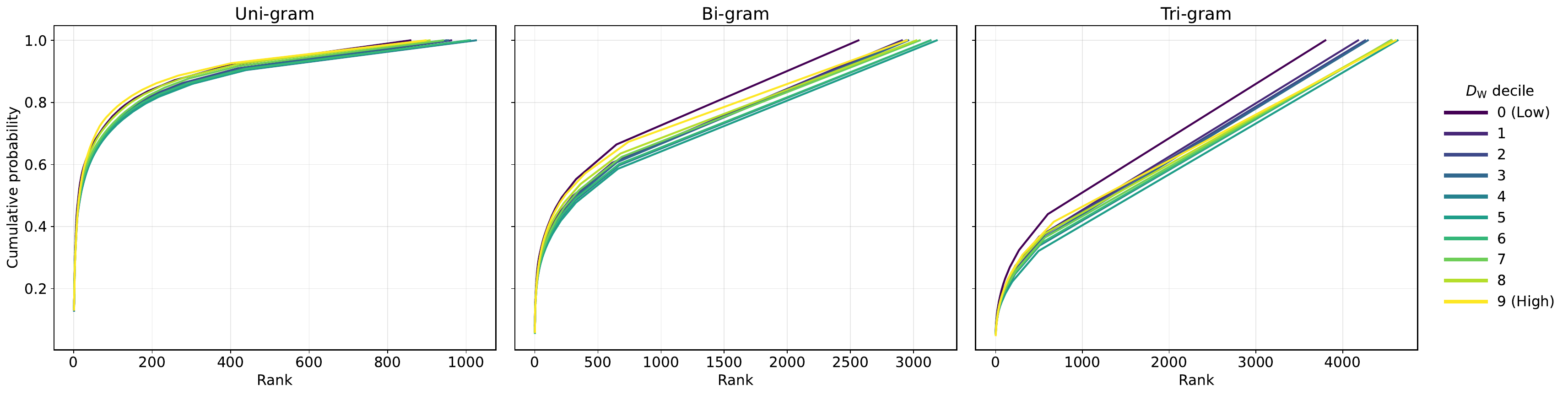}
    \caption{Cumulative N-gram probability across $D_\mathrm{W}(t)$ deciles}
    \label{fig:mscoco_siglip_txt_dw}
  \end{subfigure}
  \caption{
    N-gram probability coverage across KL deciles calculated by SigLIP. 
    Each decile in \cref{fig:mscoco_img_dklr,fig:mscoco_img_dc,fig:mscoco_img_dw} is a group of captions corresponding to images $i$ which has the same level of $D_\mathrm{KLR}(i)$, $D_\mathrm{C}(i)$ and $D_\mathrm{W}(i)$.
    Each decile in \cref{fig:mscoco_txt_dklr,fig:mscoco_txt_dc,fig:mscoco_txt_dw} is a group of captions $t$ which has the same level of each KL divergence.
    }
  \label{fig:mscoco_ngram_siglip}
\end{figure*}

\section{Ground-truth Experiments in a Toy Setting}
\begin{figure*}[t]
  \centering
  \begin{tabular}{@{}cc@{}}
    \includegraphics[width=0.48\linewidth]{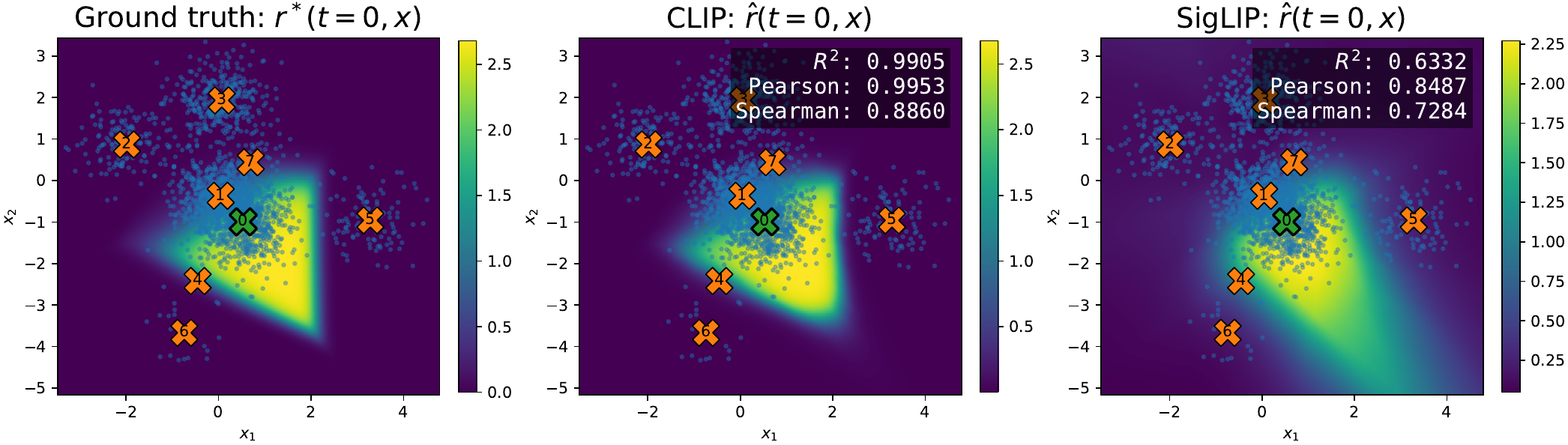} &
    \includegraphics[width=0.48\linewidth]{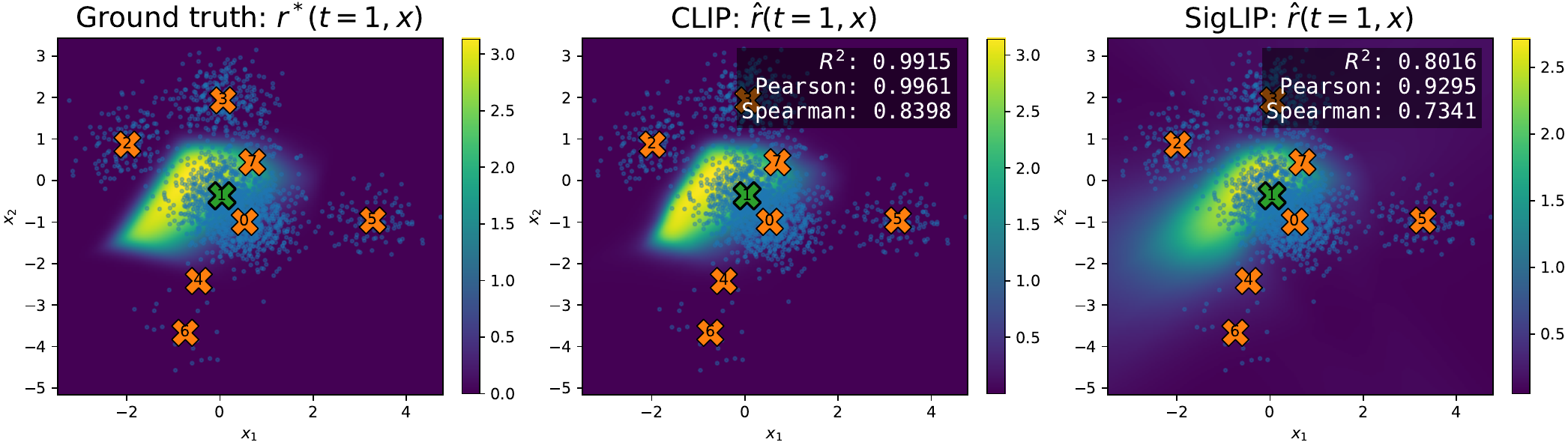} \\
    \includegraphics[width=0.48\linewidth]{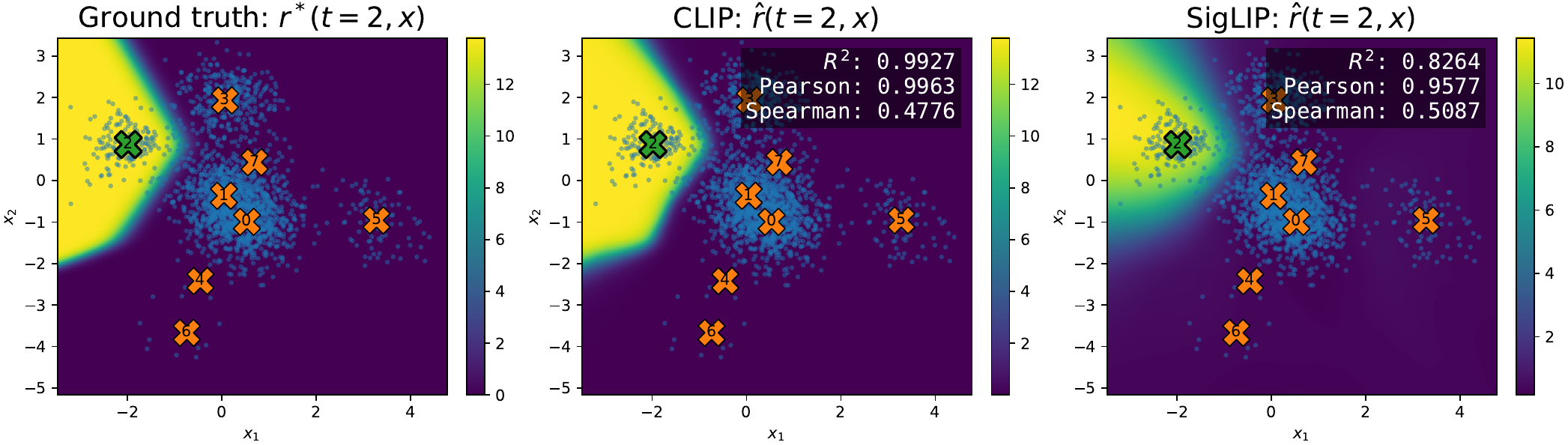} &
    \includegraphics[width=0.48\linewidth]{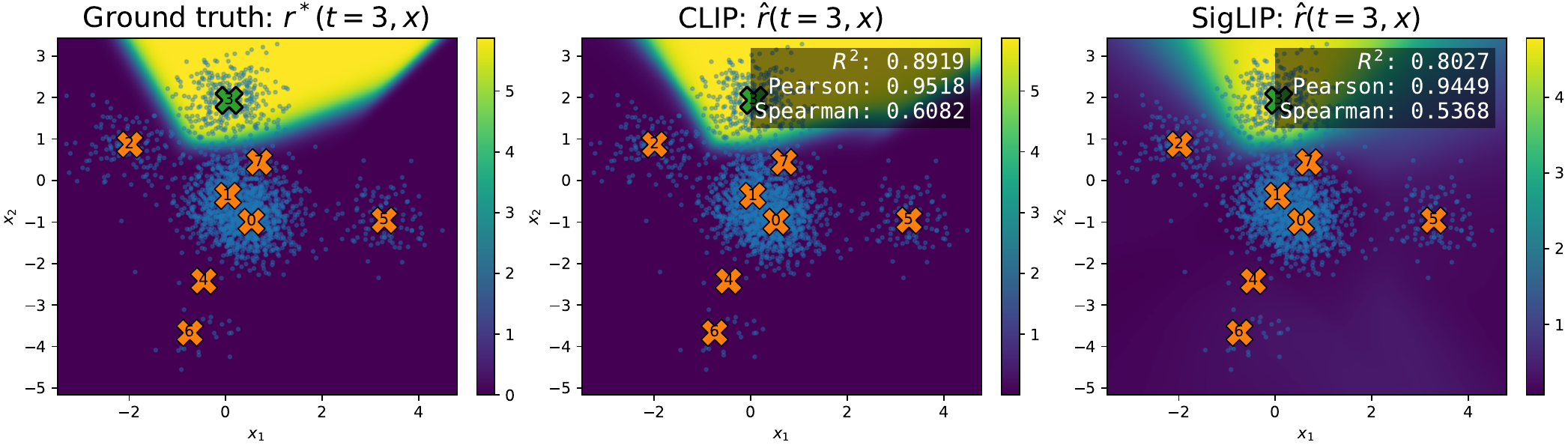} \\
    \includegraphics[width=0.48\linewidth]{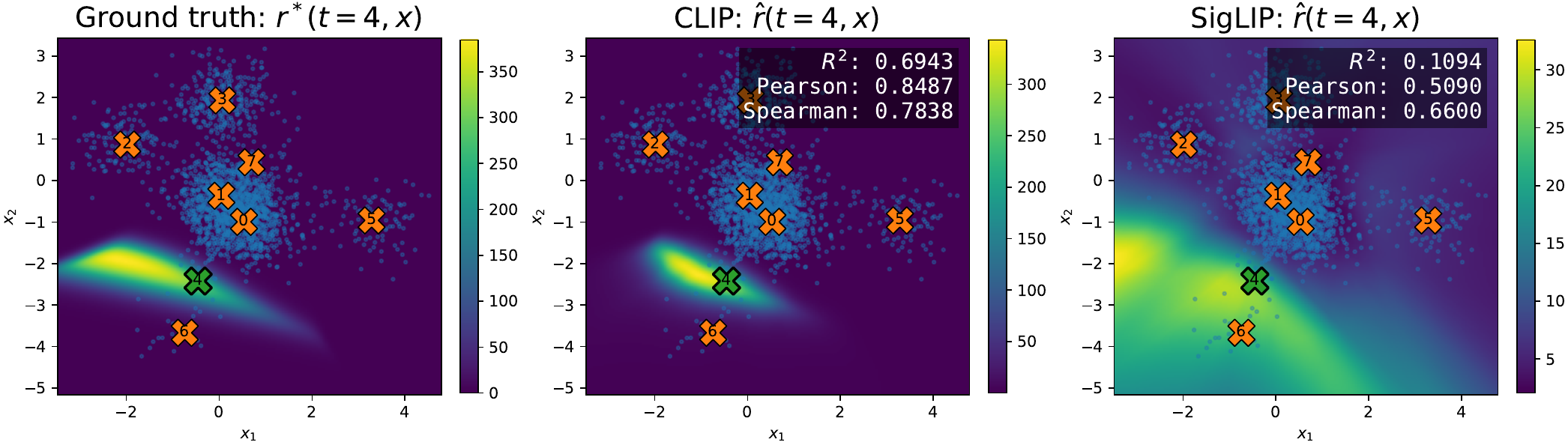} &
    \includegraphics[width=0.48\linewidth]{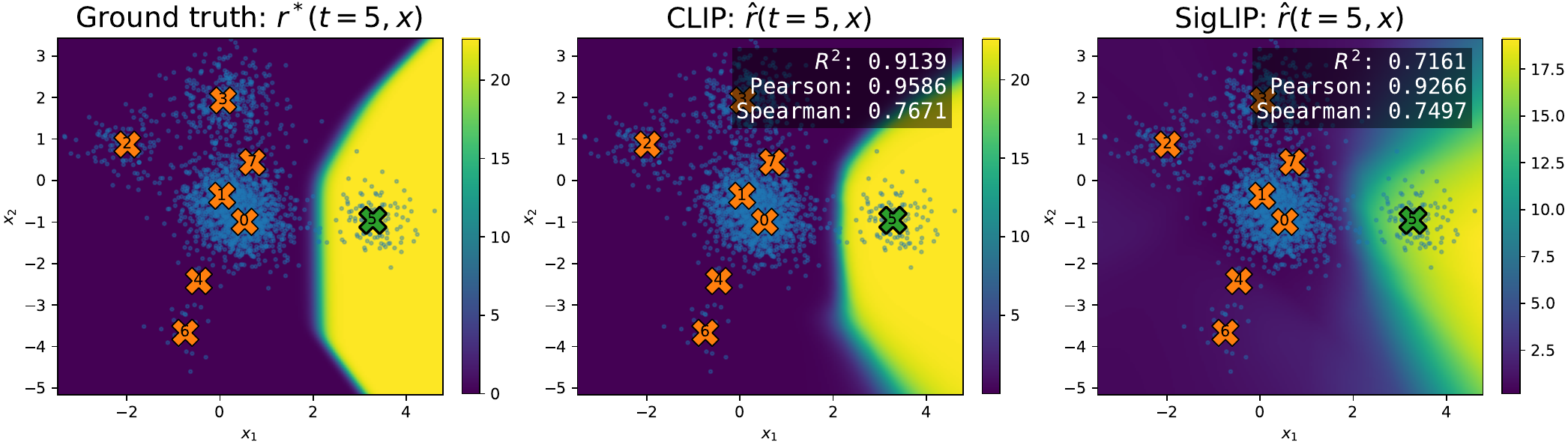} \\
    \includegraphics[width=0.48\linewidth]{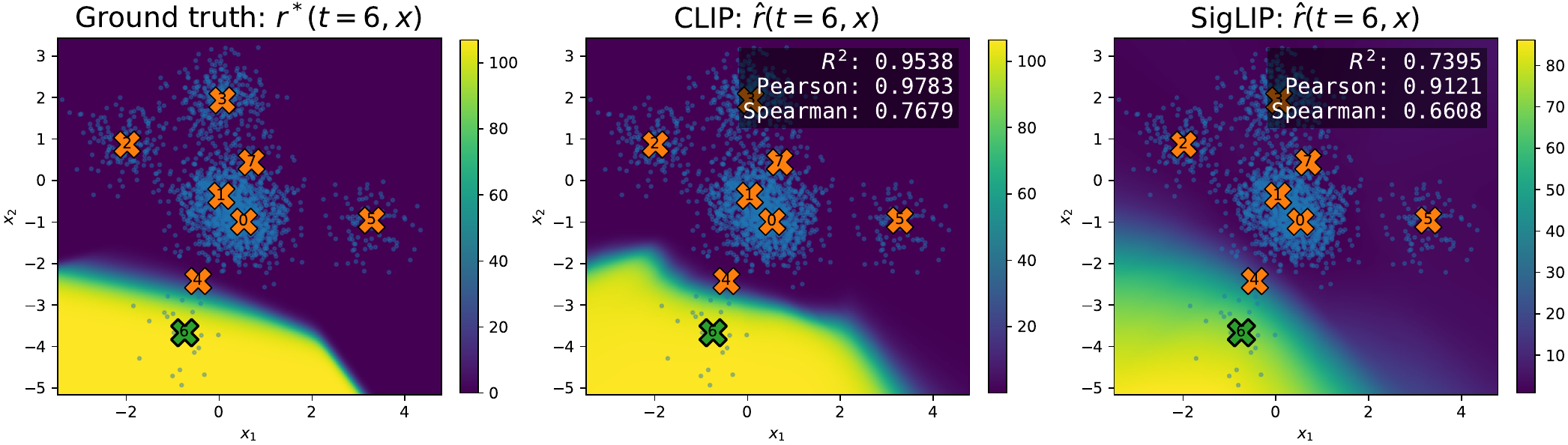} &
    \includegraphics[width=0.48\linewidth]{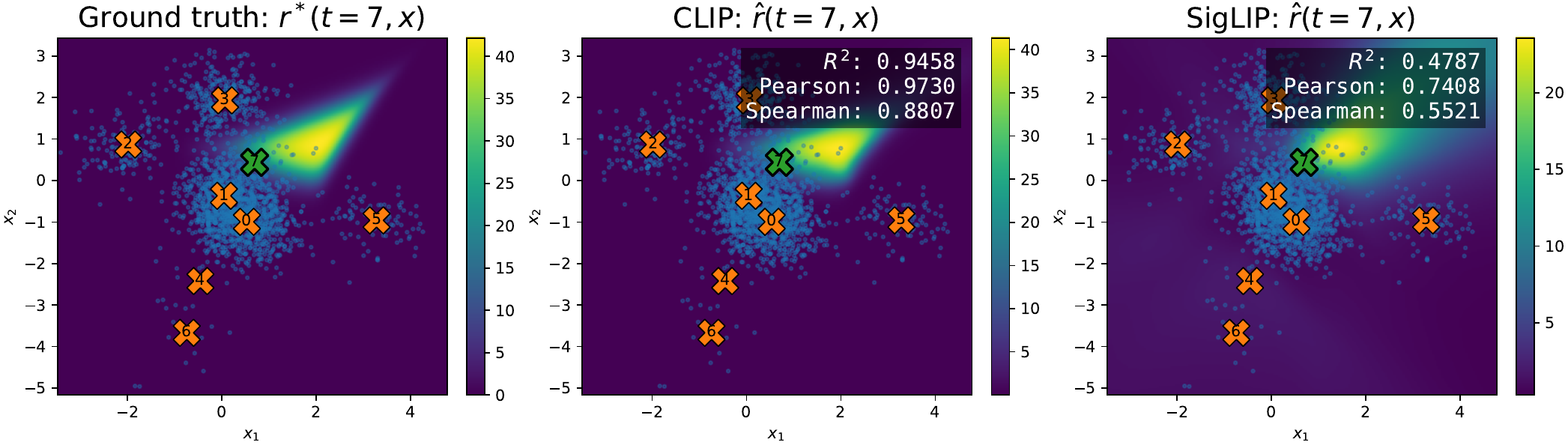} \\
  \end{tabular}
   \caption{Visualized ground truth and predicted density ratio on image space. Each cross mark is the mean vector corresponding to the text, and blue points are the sample plots. Each metric on the upper right is computed from a grid of density ratios for each mean vector.}
   \label{fig:toy}
\end{figure*}

In the main part of the paper, we have verified the density ratio modeling of CLIP-like models by implementing applications based on density ratio. On the other hand, we also conducted small-scale experiments to directly validate the density ratio modeling of the CLIP-like models. For $K\in \mathbb{N}$, we prepared eight artificial class labels $t\in \{t_j\}_{j=1}^K$ sampled from a categorical distribution. This corresponds to captions in CLIP-like models. Then, an image $i\in \mathbb{R}^d (d\in\mathbb{N})$ sampled from a Gaussian mixture model; each peak is a Gaussian distribution $\mathcal{N}(\mu_j, 4I)$ and $\mu_j\in \mathbb{R}^d$ corresponds to $t_j$. We trained lightweight CLIP-like encoders, implemented as three-layer MLPs, using the InfoNCE and NCE objectives. As shown in Figure \ref{fig:toy}, the predicted density ratios in 2D image space with eight labels closely match and have high correlation coefficients with the ground-truth ratio. Moreover, \cref{tab:synthetic} demonstrates that CLIP achieves almost perfect estimation across all settings, while SigLIP shows varying accuracy depending on dimensionality. This validates our theoretical interpretation that the contrastive objectives learn density ratios, complementing the real-data applications in the main paper.

\begin{table}[t]
\centering
\resizebox{\linewidth}{!}{%
\begin{tabular}{cc|ccc|ccc}
\toprule
$K$ & $d$ & \multicolumn{3}{c|}{CLIP} & \multicolumn{3}{c}{SigLIP} \\
& & R$^2$$\uparrow$ & MSE$\downarrow$ & Pearson$\uparrow$ & R$^2$$\uparrow$ & MSE$\downarrow$ & Pearson$\uparrow$ \\
\midrule
8 & 2 & 0.9938 & 4.46e-01 & 0.9971 & 0.7722 & 1.63e+01 & 0.9457 \\
 & 8 & 1.0000 & 1.05e-02 & 1.0000 & 0.9250 & 1.51e+02 & 0.9978 \\
 & 64 & 1.0000 & 9.60e-04 & 1.0000 & 0.9723 & 4.84e+01 & 0.9983 \\
 & 128 & 1.0000 & 1.03e-03 & 1.0000 & 0.9924 & 1.12e+01 & 0.9998 \\
 & 512 & 1.0000 & 5.35e-03 & 1.0000 & 0.9823 & 4.04e+01 & 0.9997 \\
\midrule
16 & 2 & 0.9947 & 4.25e-01 & 0.9974 & 0.9250 & 6.02e+00 & 0.9847 \\
 & 8 & 0.9998 & 9.53e-01 & 0.9999 & 0.1484 & 4.05e+03 & 0.3980 \\
 & 64 & 0.9998 & 3.30e+00 & 1.0000 & 0.0927 & 1.33e+04 & 0.3595 \\
 & 128 & 1.0000 & 2.23e-01 & 1.0000 & 0.7778 & 1.62e+03 & 0.9589 \\
 & 512 & 1.0000 & 9.38e-02 & 1.0000 & 0.9763 & 1.11e+02 & 0.9991 \\
\bottomrule
\end{tabular}}
\caption{Density ratio estimation accuracy (R$^2$, MSE, Pearson) on synthetic data. $K$ is the number of the artificial labels, and $d$ is the dimension of image space. Each metric is calculated using density ratios in test data.}
\label{tab:synthetic}
\end{table}

\section{Estimation Error Analysis}
\label{sec:kl_err}
All approximations of the KL divergences need a sample set. 
Let us consider estimates of the image KL divergence.
Right sides of \cref{eq:dkl_approx,eq:dklr_approx} employ sample text set $\mathcal{D}_T$, and \cref{eq:dw_approx_img,eq:dc_approx_img} use both text set and image set for calculating $\hat{G}_T$, $\hat{G}_I$, $\hat{v}_T$ and $\hat{v}_I$.
We evaluated approximation bias and its variance arising from finite-sample approximation using the bootstrap method.

\subsection{Bias, Variance and RMSE}

To analyze the error of finite sample approximation, we prepared $B$ sample sets of texts and images $\{\mathcal{D}_T^b\}_{b=1}^B, \{\mathcal{D}_I^b\}_{b=1}^B$ where each $\mathcal{D}_T^b$ and $\mathcal{D}_I^b$ are sampled from the original $\mathcal{D}_T$ and $\mathcal{D}_I$ with replacement.

Let $\hat{D}_\mathrm{KL}(i;b), \hat{D}_\mathrm{KLR}(i;b)$ denote the estimated value of $D_\mathrm{KL}(i), D_\mathrm{KLR}(i)$ using $\mathcal{D}_T^b$.
We also denote the estimated value of $D_\mathrm{C}(i), D_\mathrm{W}(i)$ using $\mathcal{D}_T^b$ and $\mathcal{D}_I^b$ as $\hat{D}_\mathrm{C}(i;b), \hat{D}_\mathrm{W}(i;b)$
We collectively denote these estimations using original samples $\mathcal{D}_T$ (and $\mathcal{D}_I$) as $\hat{D}_*(i)$, and these estimations using $\mathcal{D}_T^b$ (and $\mathcal{D}_I^b$) as $\hat{D}_*(i;b)$.

For a fixed image $i$, the estimation bias and variance can be calculated as
\begin{align}
    \mathrm{Bias}(i) &= \frac{1}{B}\sum_{b=1}^B \hat{D}_*(i;b) - \hat{D}_*(i) \\
    \mathrm{Variance}(i) &= \frac{1}{B} \sum_{b=1}^B \bigg(\hat{D}_*(i;b) - \bar{D}_*(i)\bigg)^2
\end{align}
where $\bar{D}_*(i) := \frac{1}{B}\sum_{b=1}^B \hat{D}_*(i;b)$.
Note that Variance(i) is not the variance of the KL divergence across images, but across sample sets.

Root Mean Square Error (RMSE) is then computed as follows:

\begin{align}
    \mathrm{RMSE}^2(i) &= \frac{1}{B}\sum_{b=1}^B \bigg(\hat{D}_*(i;b) - \hat{D}_*(i)\bigg)^2 \notag\\
    &= \frac{1}{B}\sum_{b=1}^B \bigg(\hat{D}_*(i;b)- \bar{D}_*(i)+ \bar{D}_*(i) - \hat{D}_*(i)\bigg)^2 \notag\\
    &= \mathrm{Variance}(i) + \mathrm{Bias}(i)^2 \notag\\
    \therefore \mathrm{RMSE}(i) &= \sqrt{\mathrm{Variance}(i) + \mathrm{Bias}(i)^2}.
\end{align}

We compared each approximation by relative errors; Bias and RMSE are divided by ``scale'', which is the standard deviation of $\hat{D}_*(i)$ over $i$, and Variance is divided by squared scale (\ie variance of $\hat{D}_*(i)$ across $i$). We sampled 100 sets of texts and images from 50,000 image-text pairs in MSCOCO for bootstrapping.

\begin{figure*}[t]
  \centering
  \includegraphics[width=0.32\textwidth]{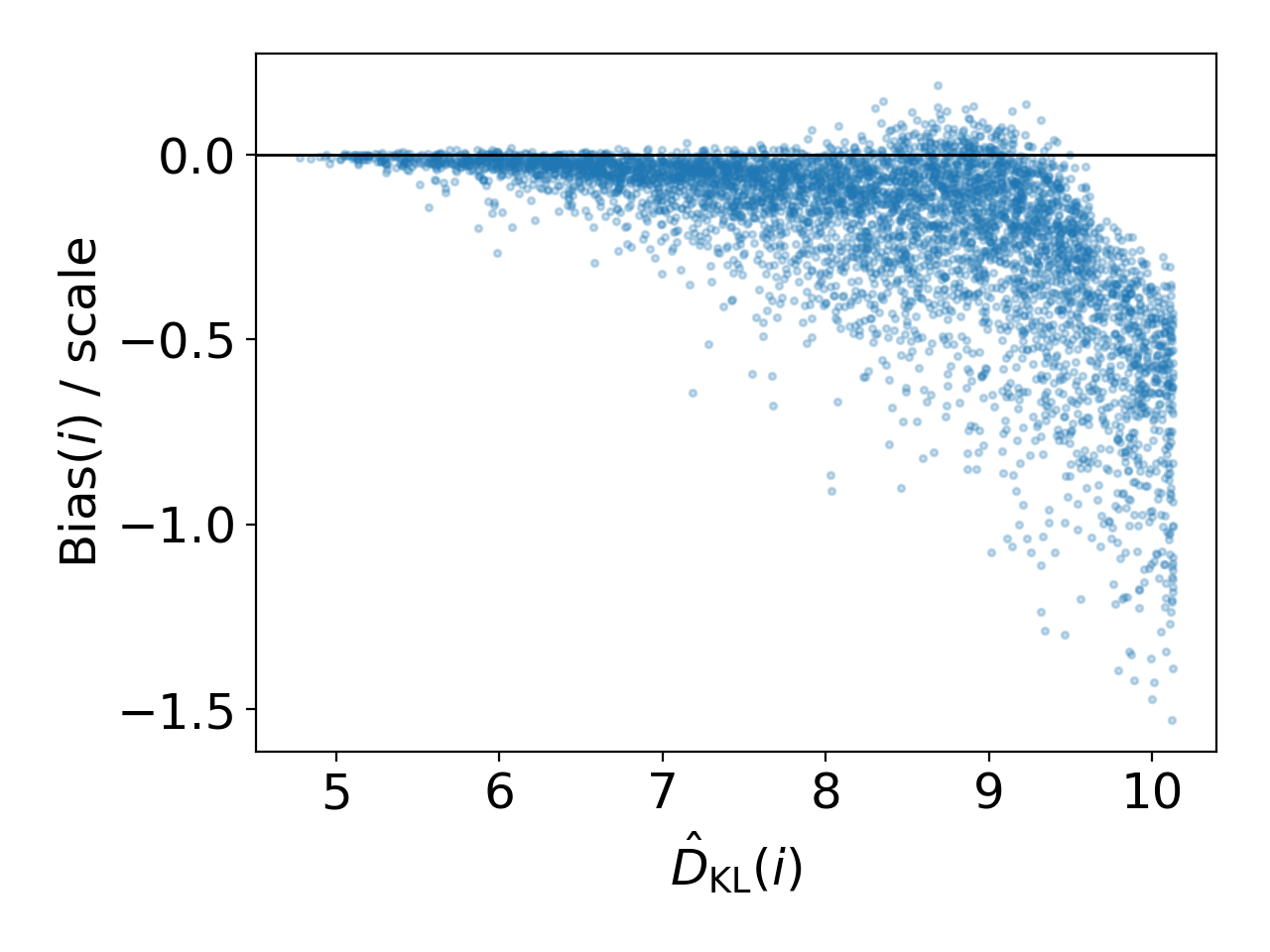}
  \includegraphics[width=0.32\textwidth]{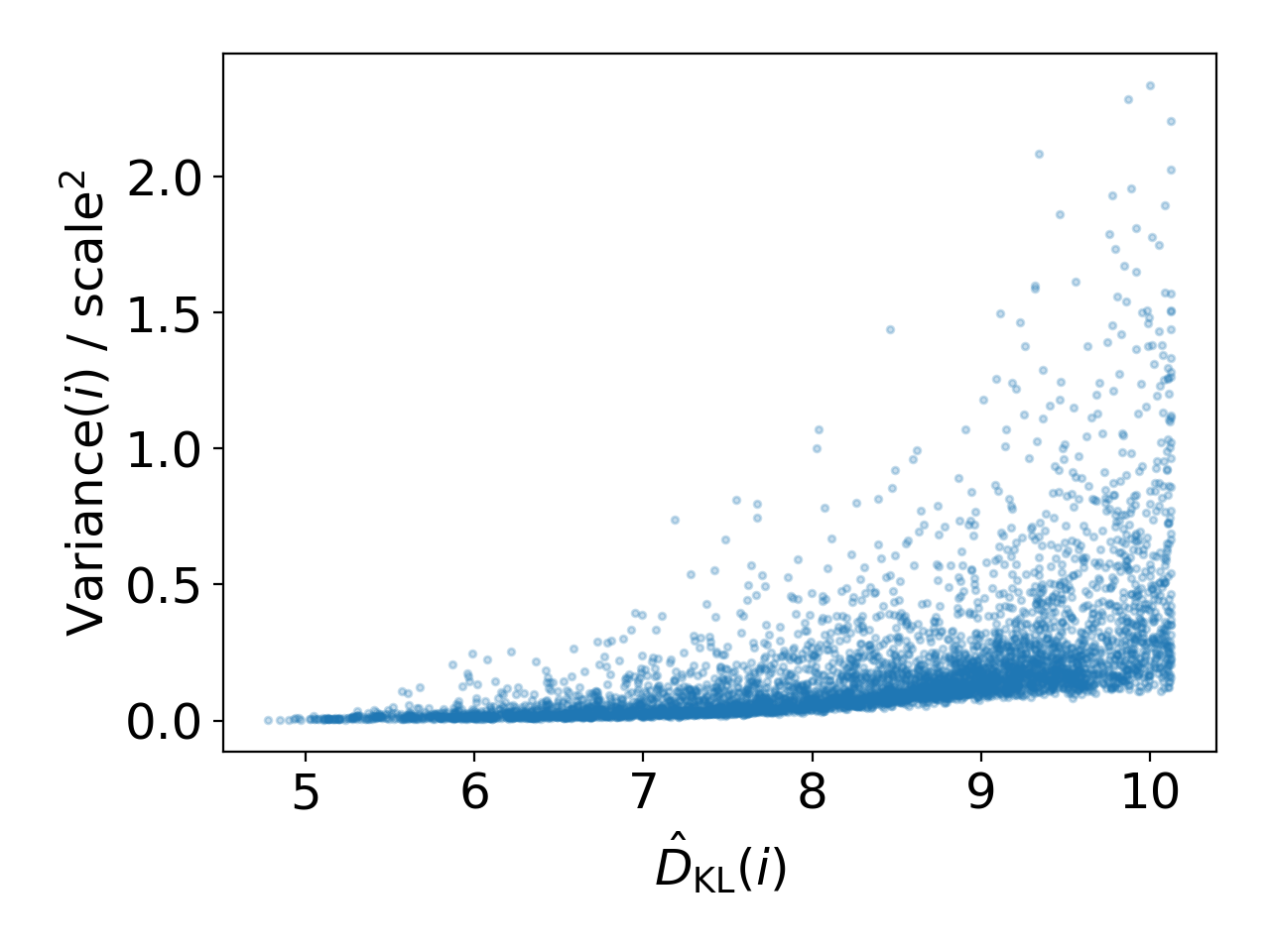}
  \includegraphics[width=0.32\textwidth]{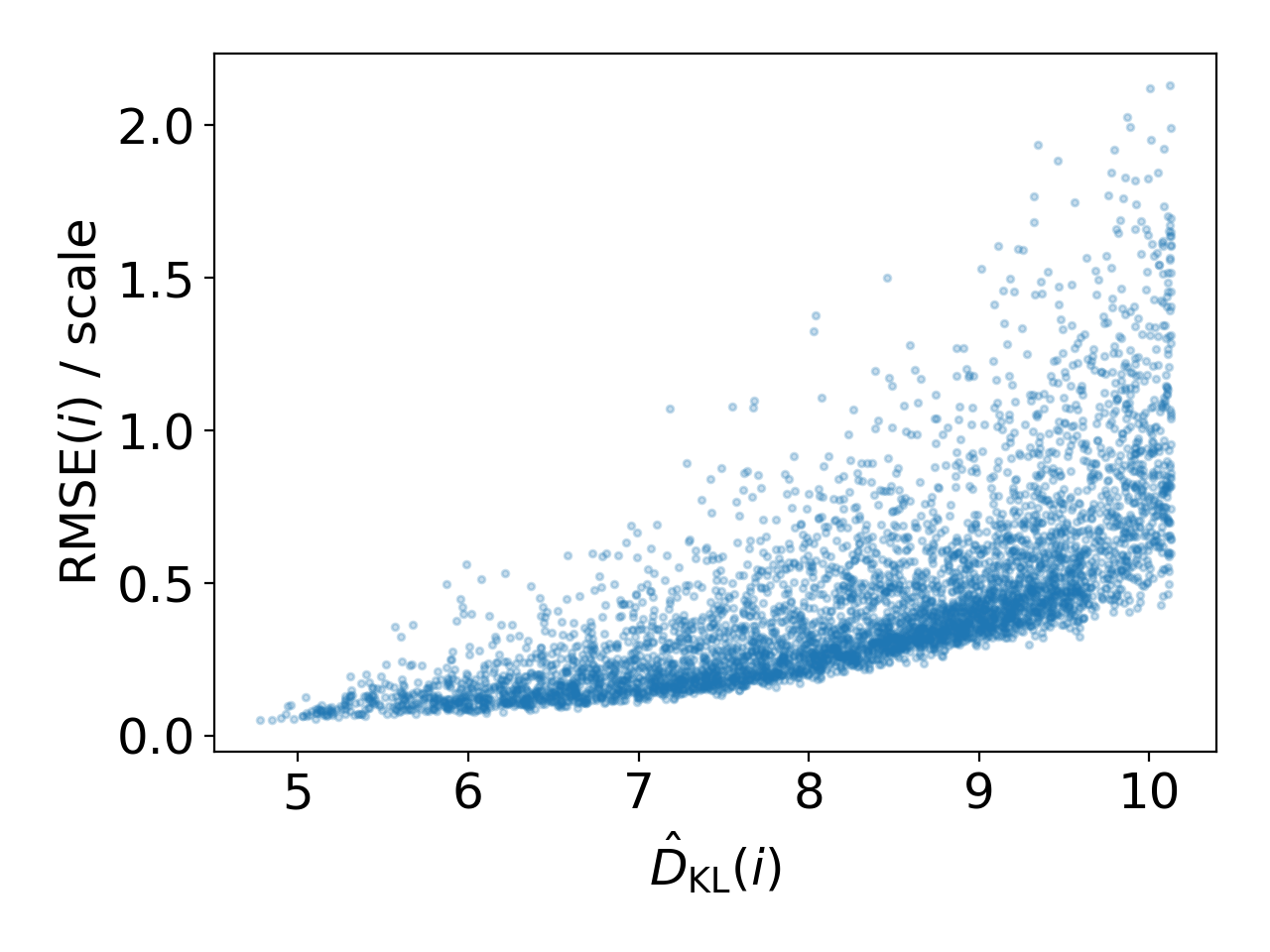}
  
  \medskip
  
  \includegraphics[width=0.32\textwidth]{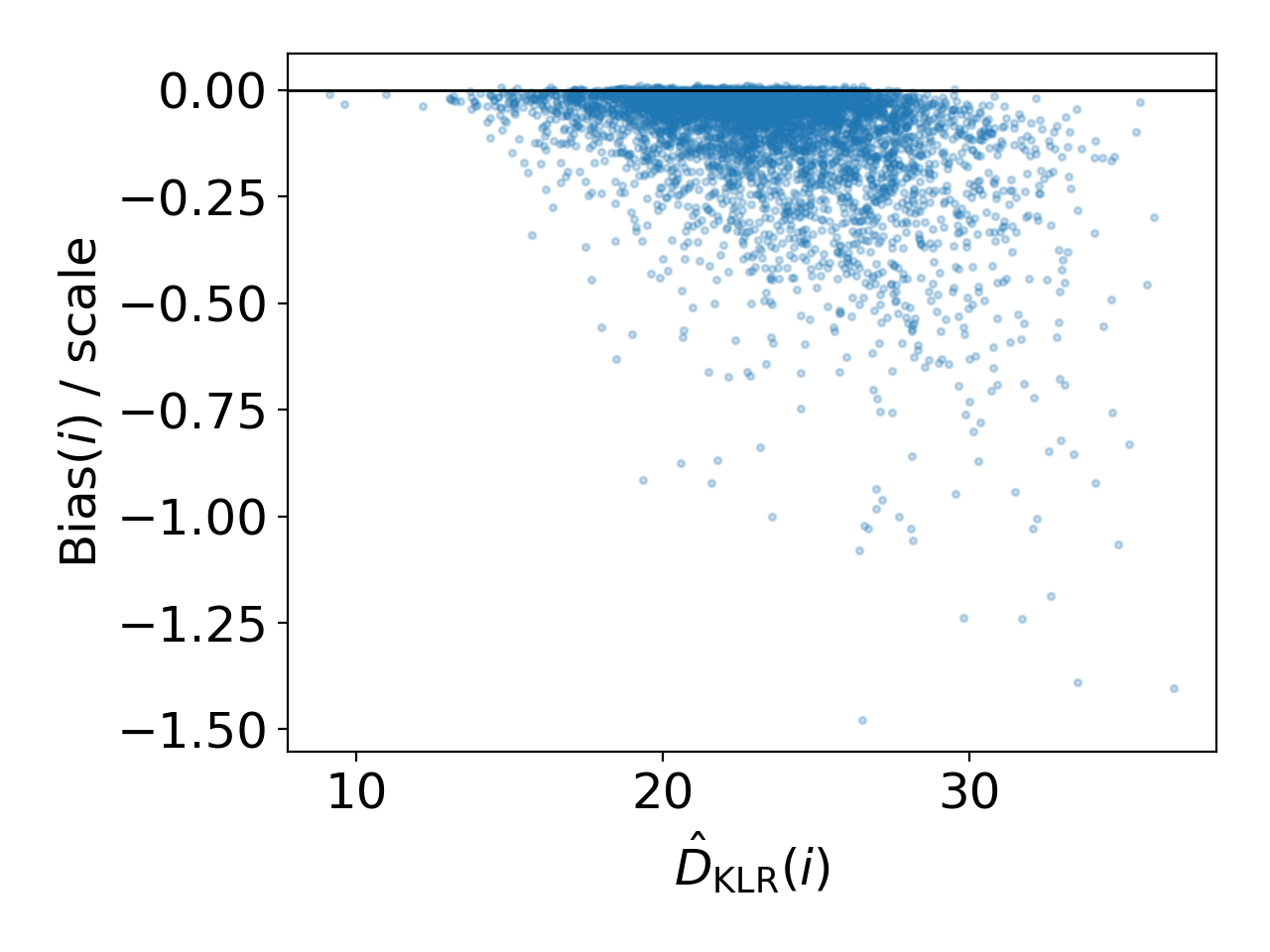}
  \includegraphics[width=0.32\textwidth]{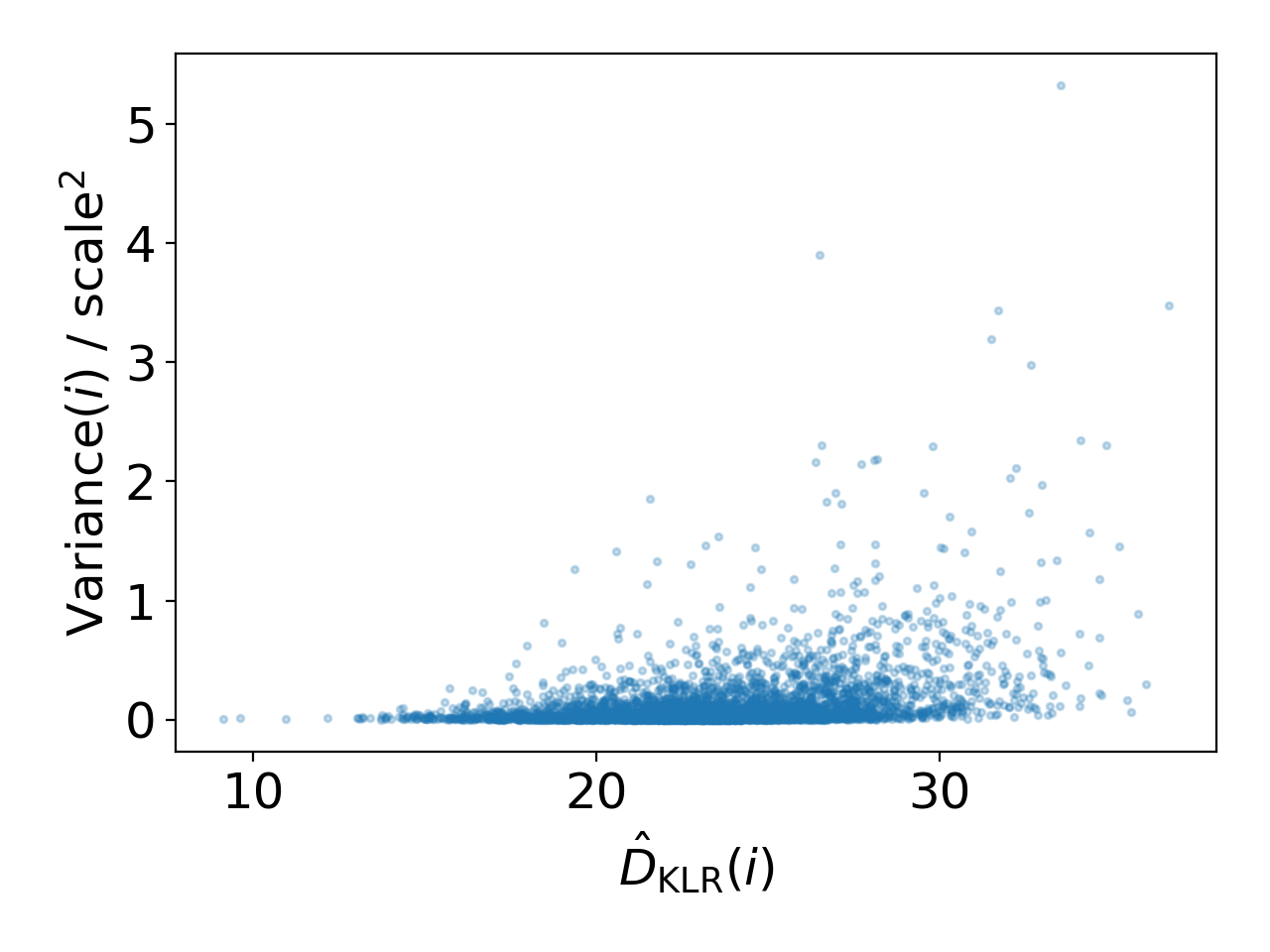}
  \includegraphics[width=0.32\textwidth]{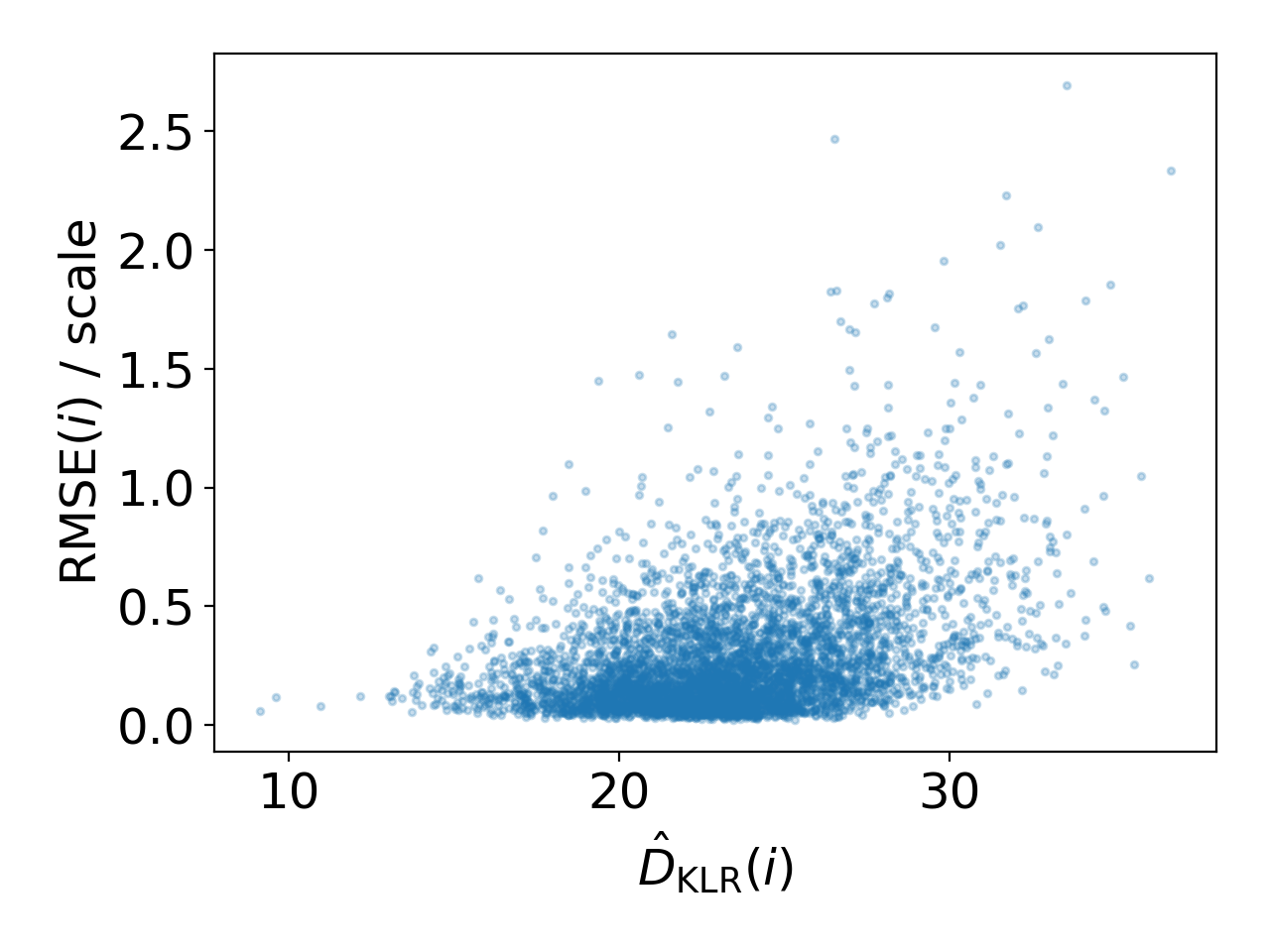}
  
  \medskip
  
  \includegraphics[width=0.32\textwidth]{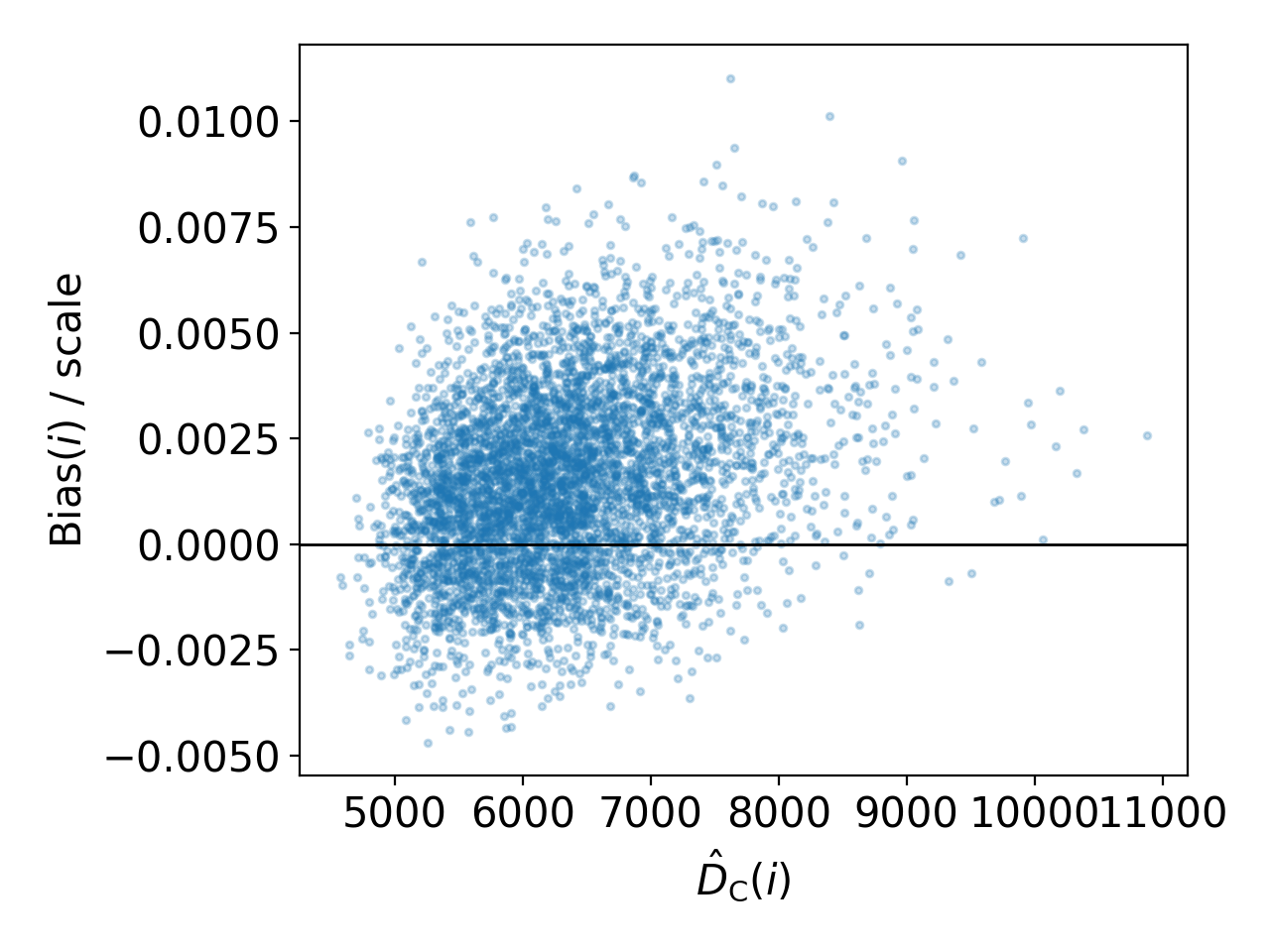}
  \includegraphics[width=0.32\textwidth]{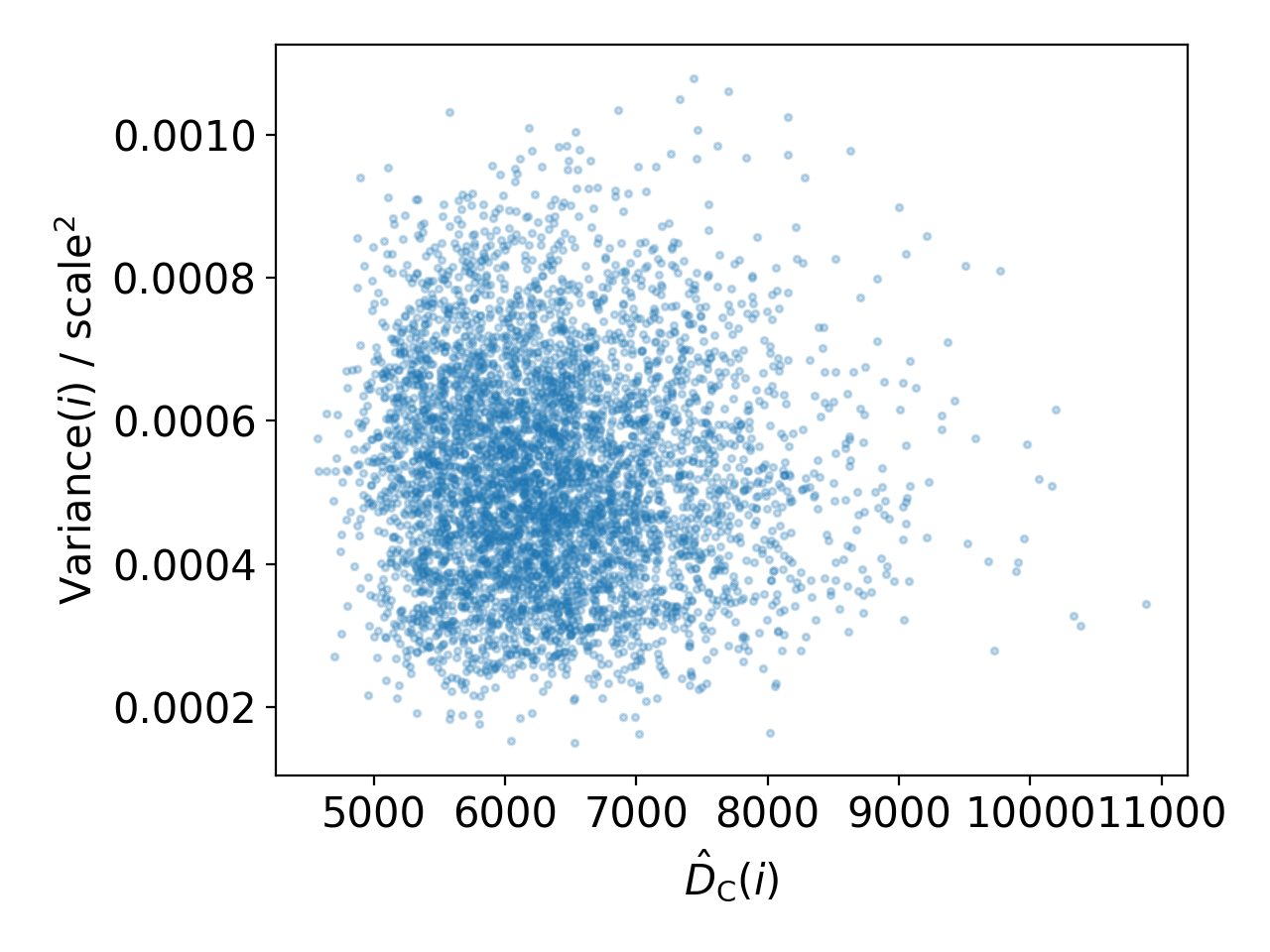}
  \includegraphics[width=0.32\textwidth]{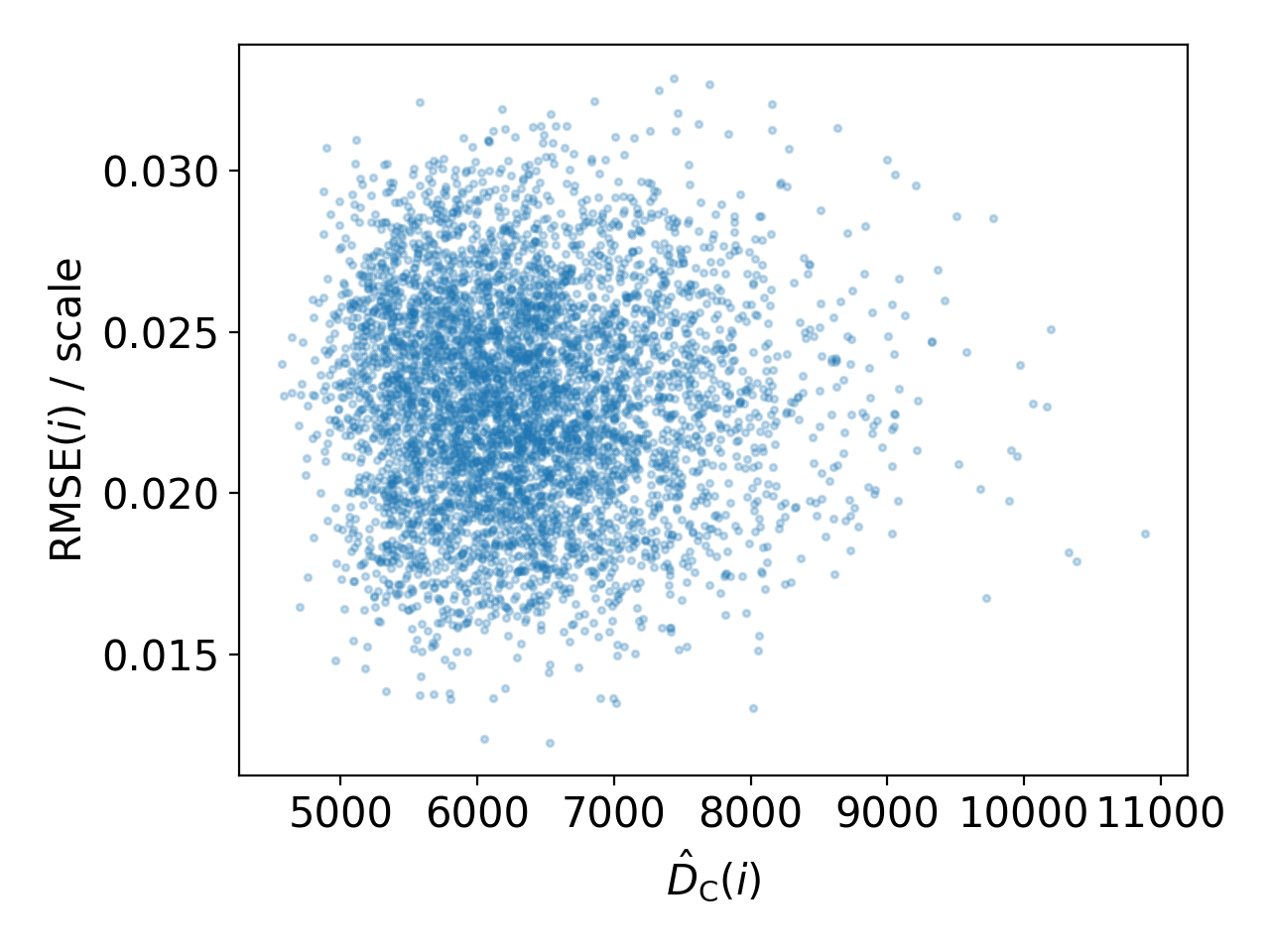}
  
  \medskip
  
  \includegraphics[width=0.32\textwidth]{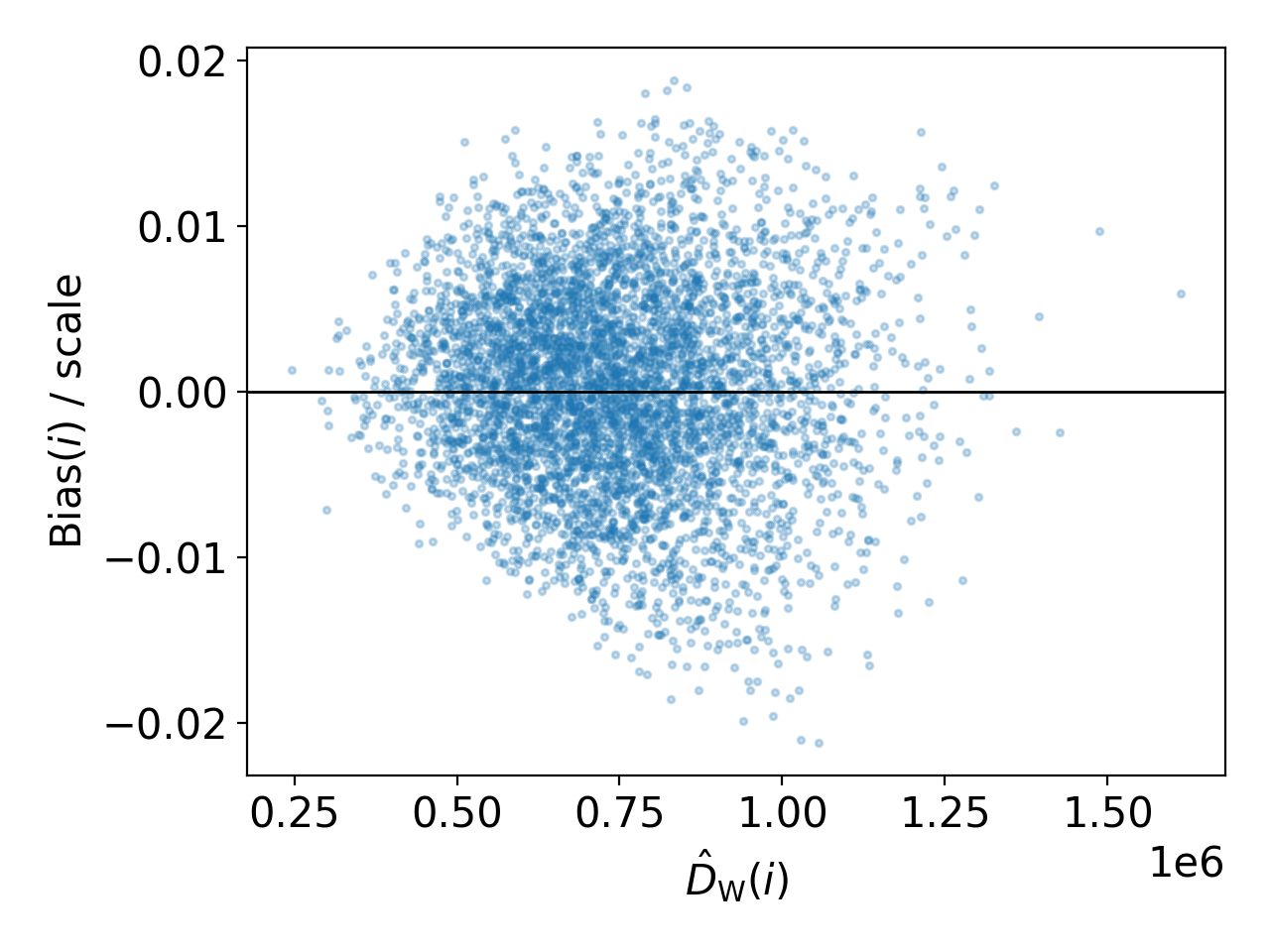}
  \includegraphics[width=0.32\textwidth]{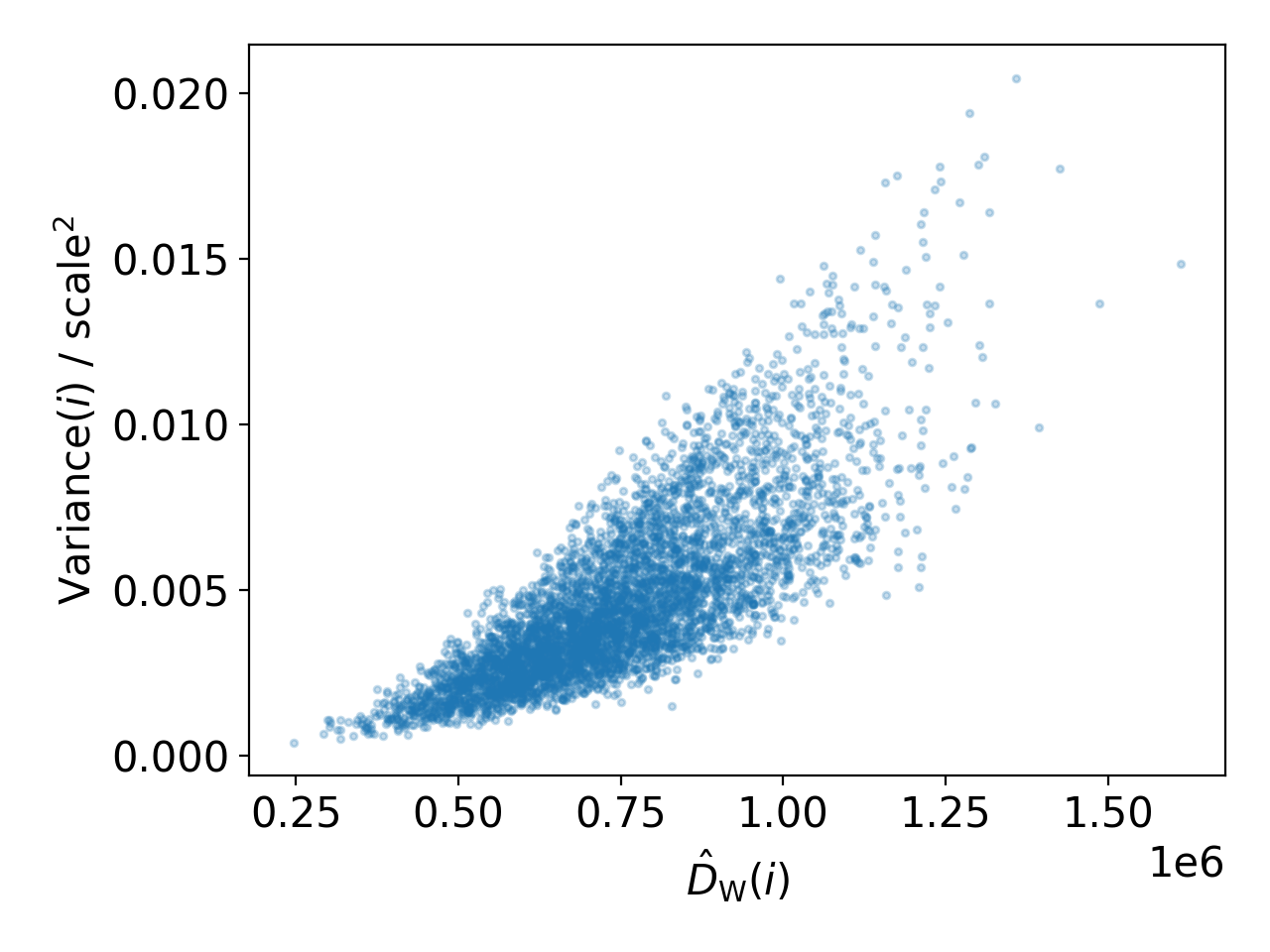}
  \includegraphics[width=0.32\textwidth]{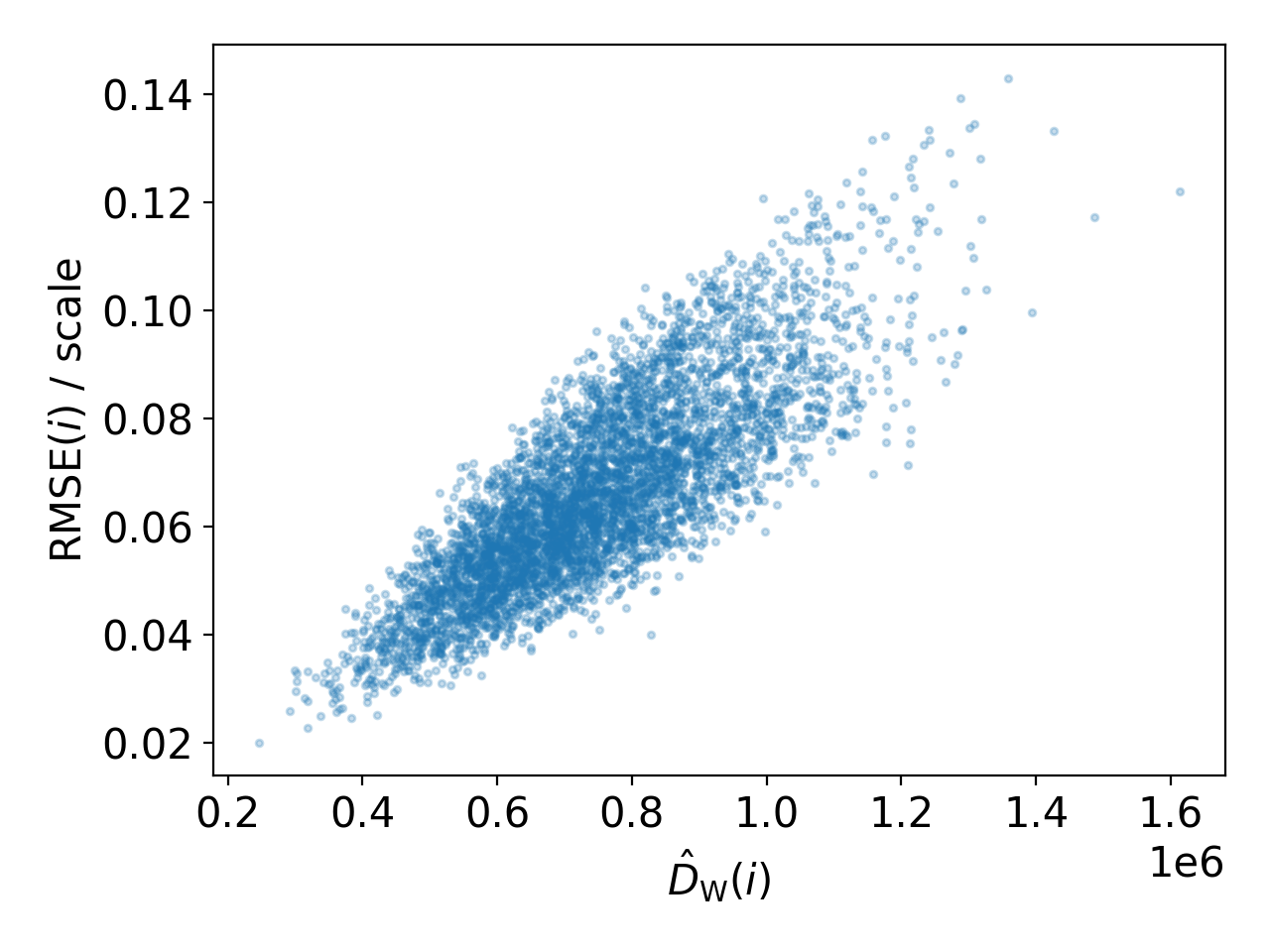}
  \caption{Scatter plots of $\hat{D}_*(i)$ and normalized $\mathrm{Bias}(i)$ (left column), $\mathrm{Variance}(i)$ (center column), $\mathrm{RMSE}(i)$ (right column). Each row represents the errors of each estimation method. KL divergences are computed using CLIP.}
  \label{fig:kl_err_clip}
\end{figure*}

\begin{figure*}[t]
  \centering
  \includegraphics[width=0.32\textwidth]{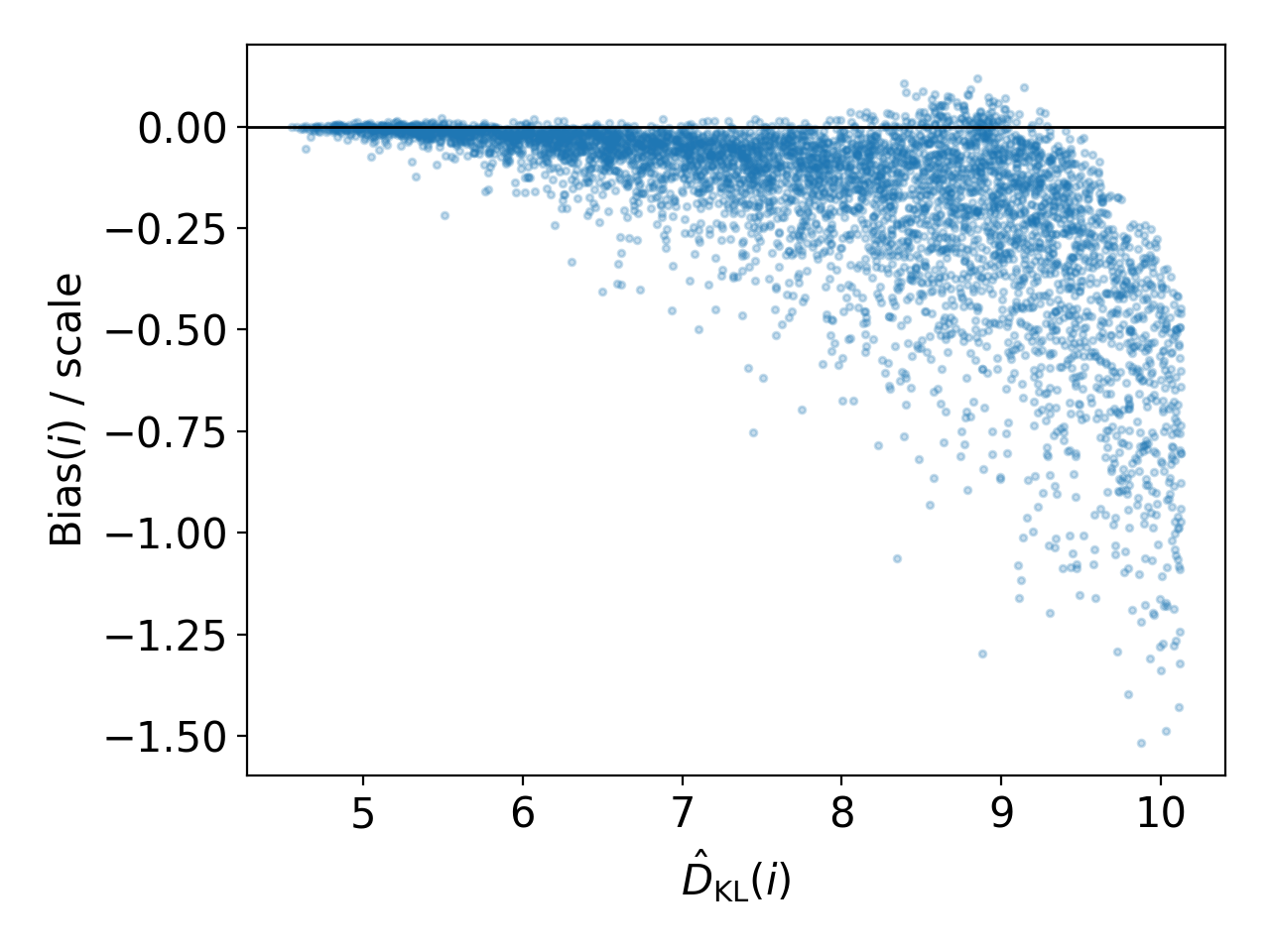}
  \includegraphics[width=0.32\textwidth]{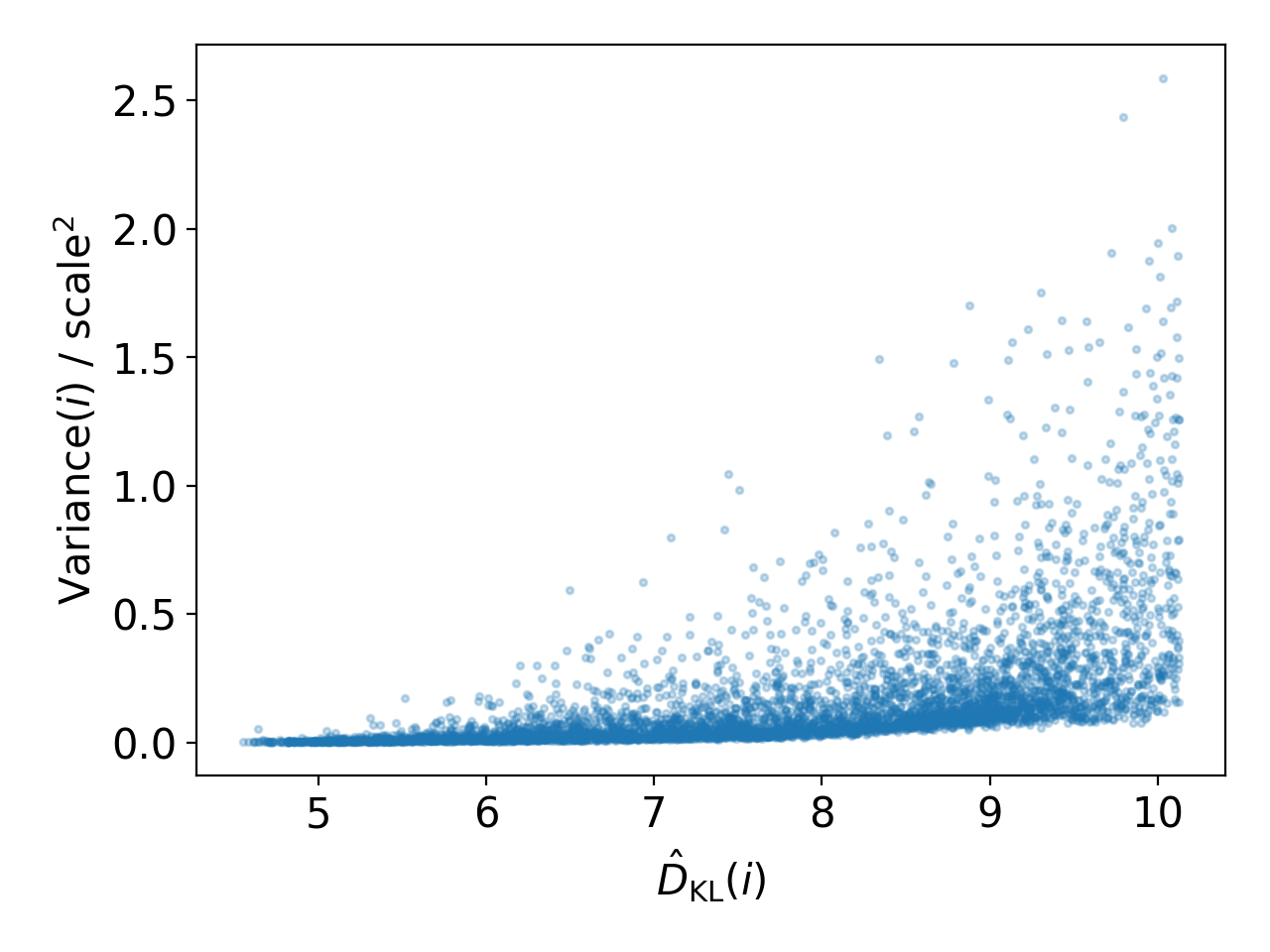}
  \includegraphics[width=0.32\textwidth]{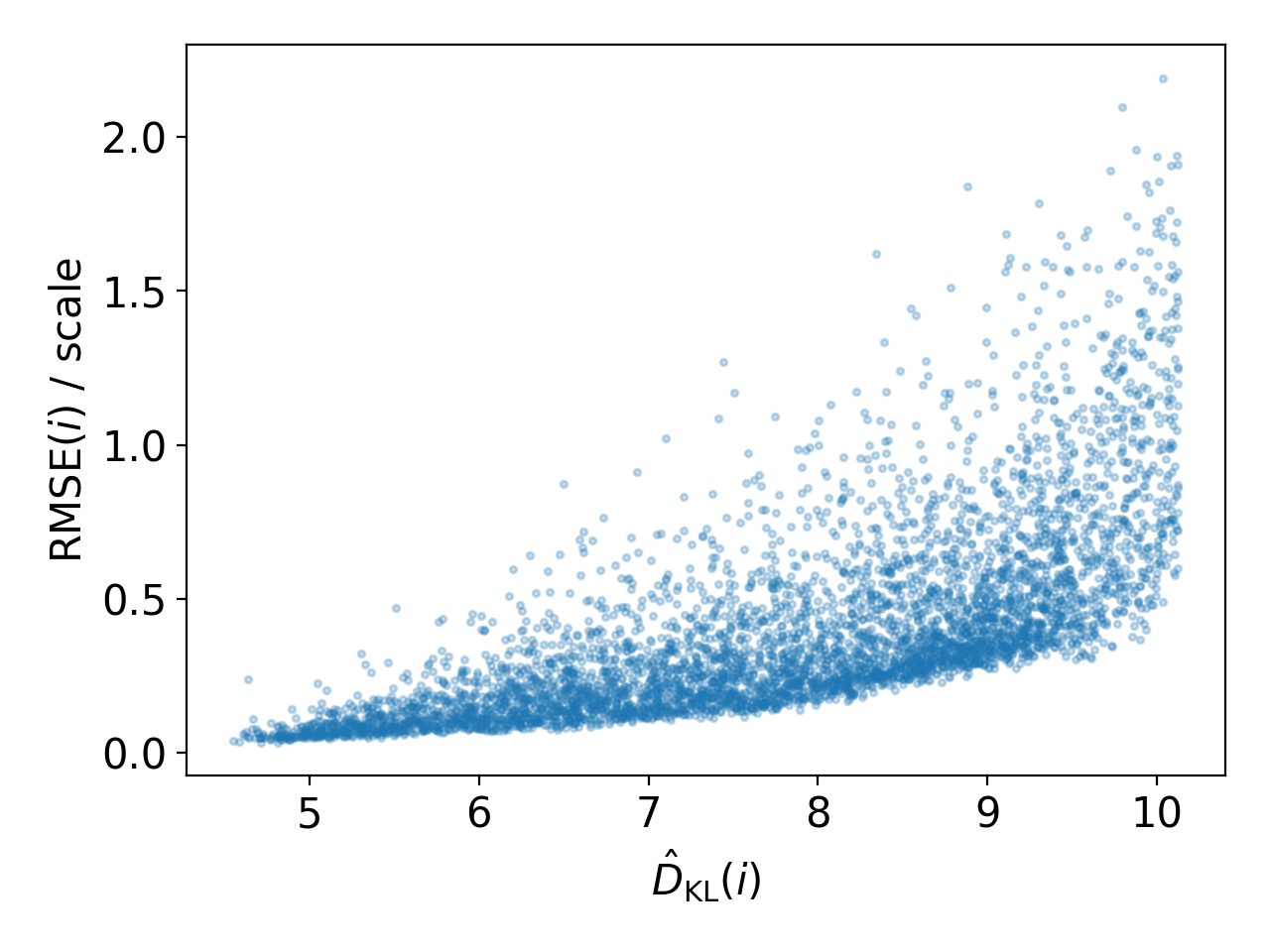}
  
  \medskip
  
  \includegraphics[width=0.32\textwidth]{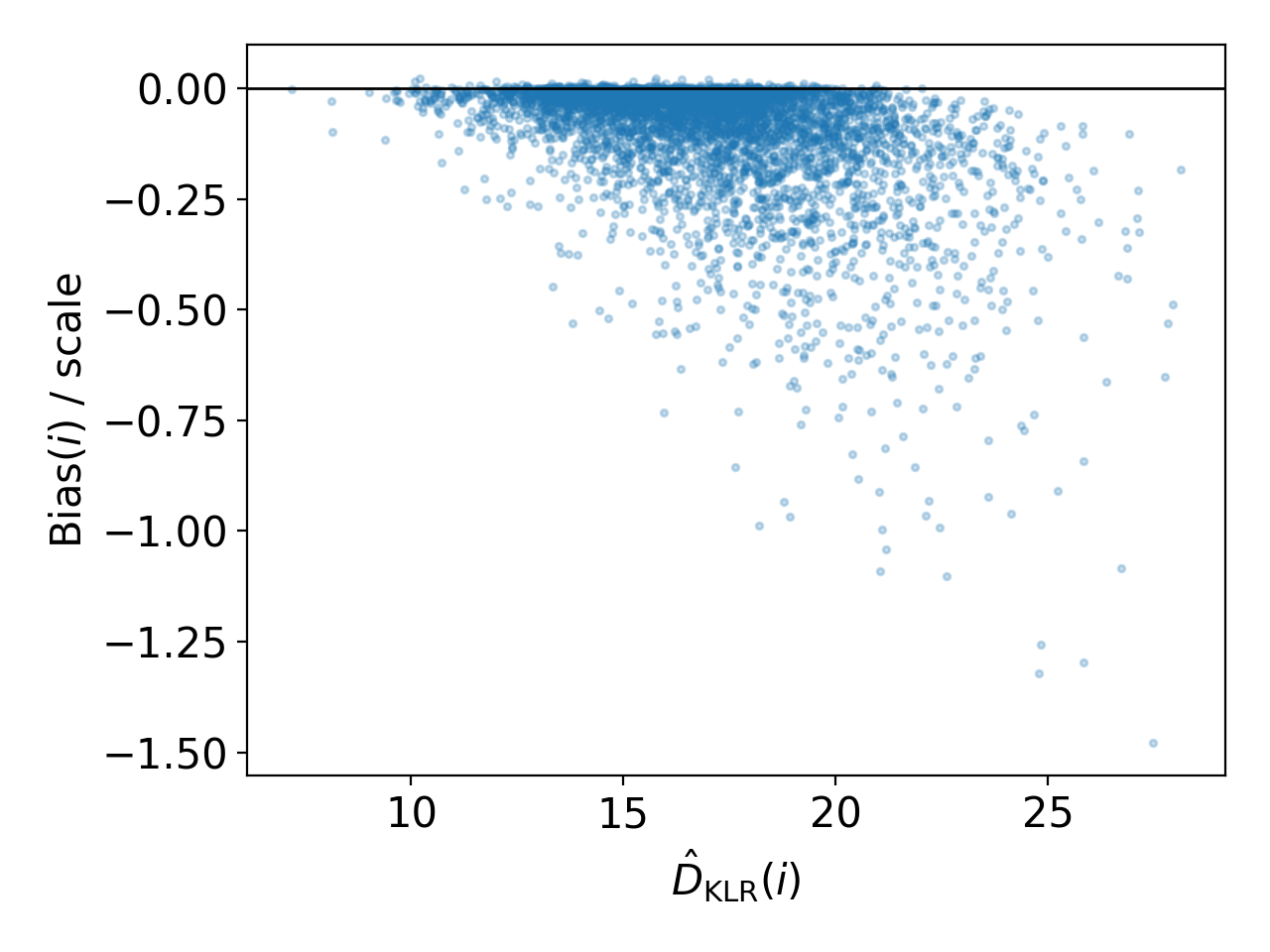}
  \includegraphics[width=0.32\textwidth]{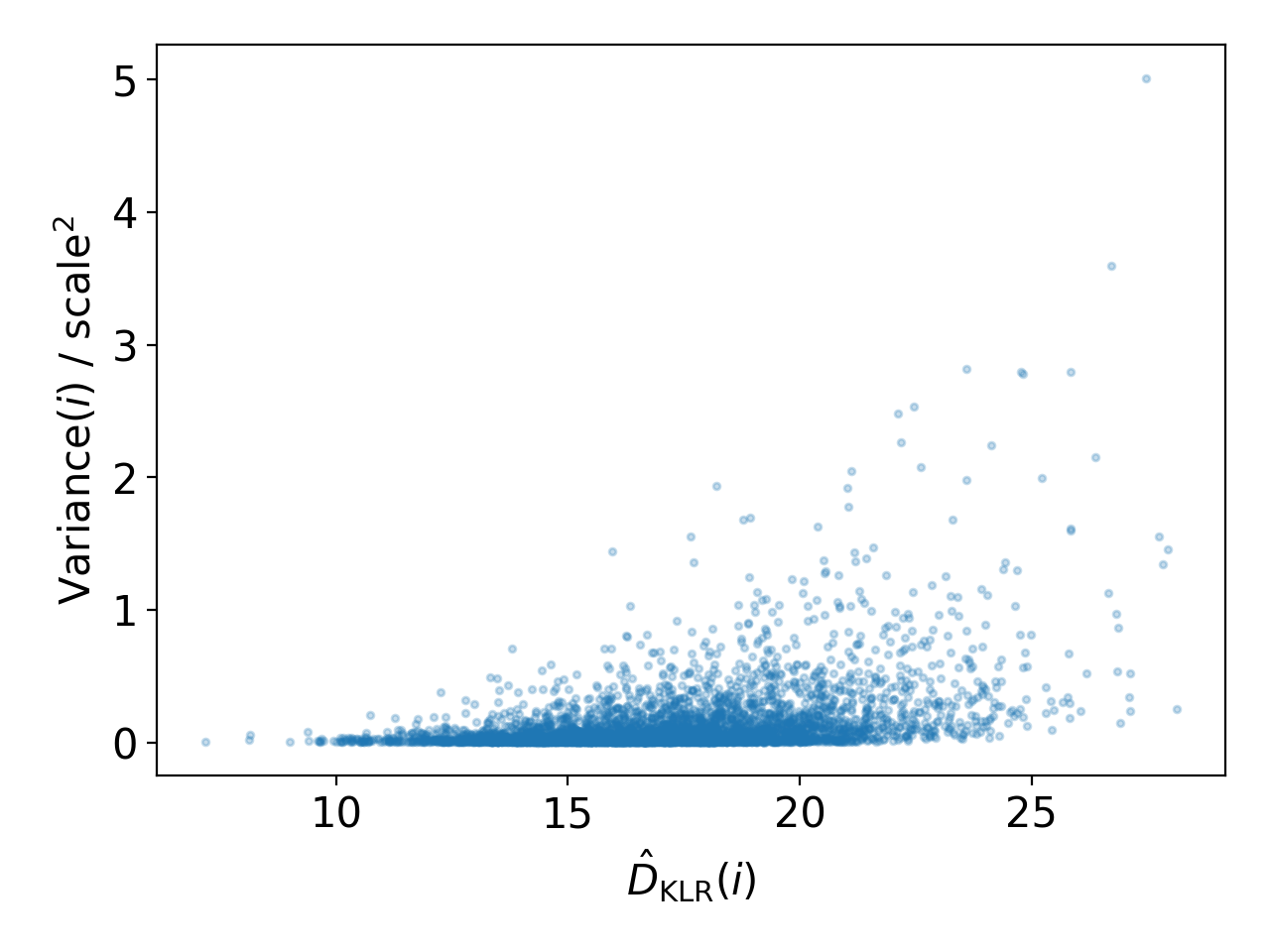}
  \includegraphics[width=0.32\textwidth]{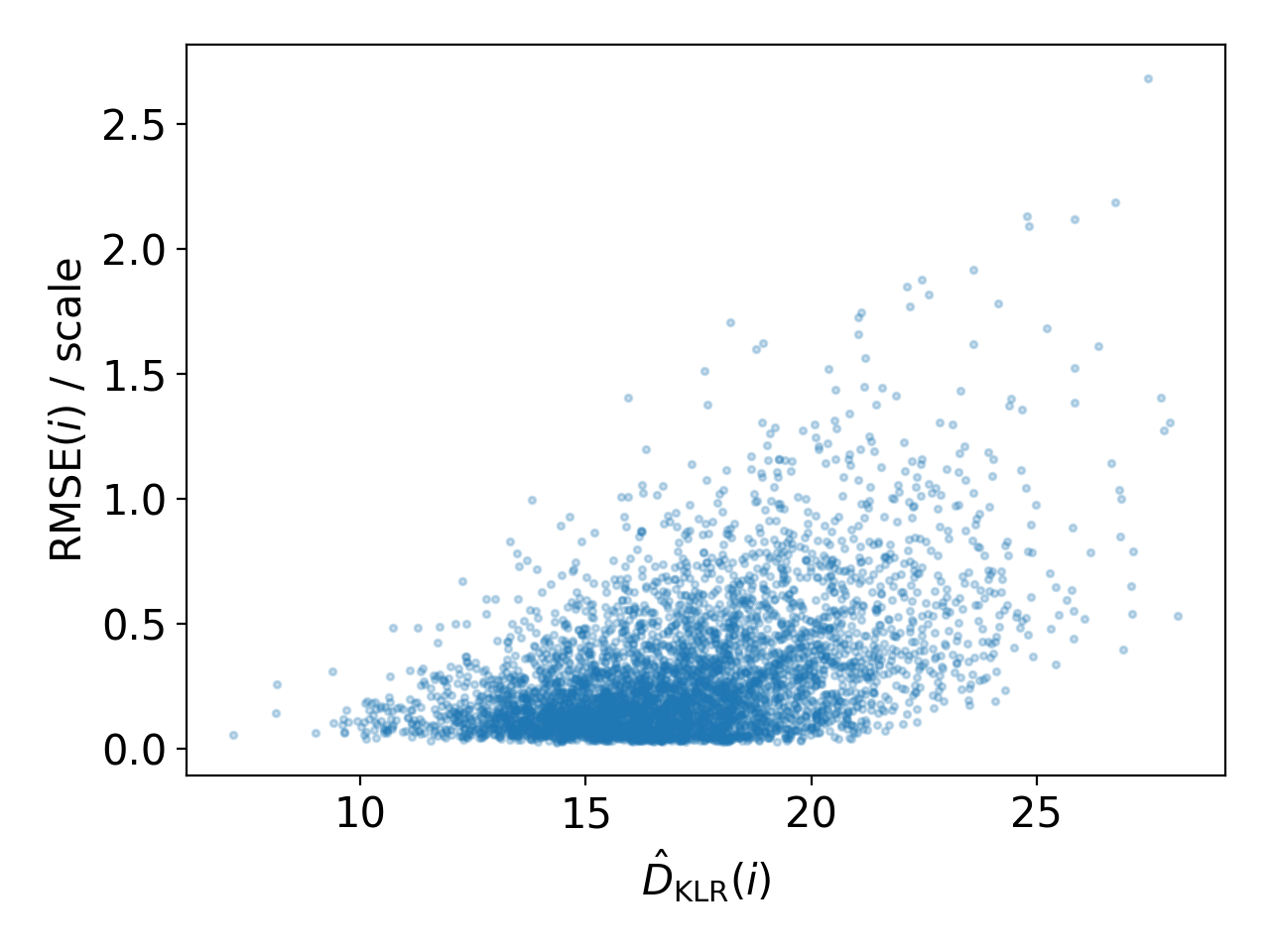}
  
  \medskip
  
  \includegraphics[width=0.32\textwidth]{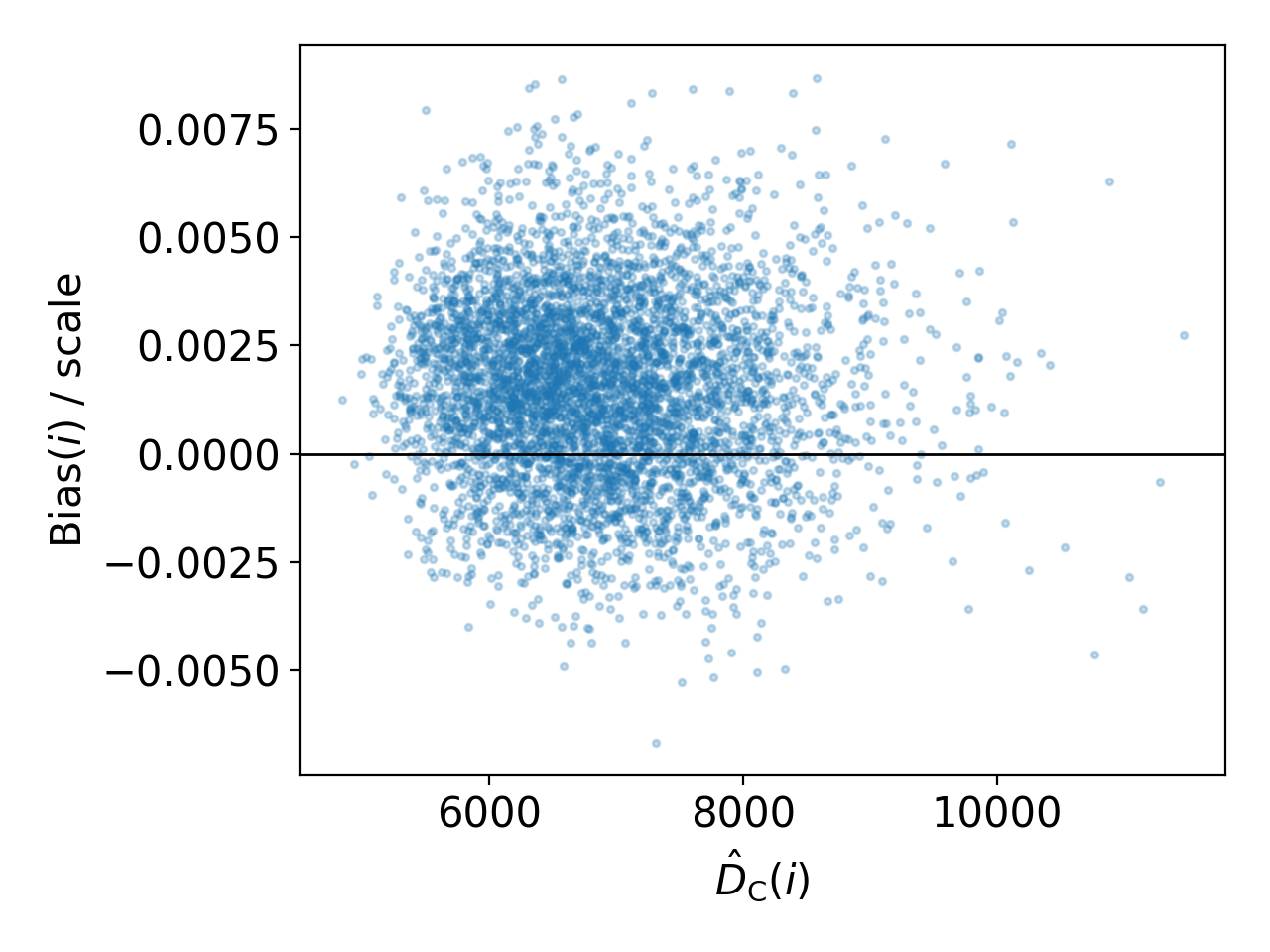}
  \includegraphics[width=0.32\textwidth]{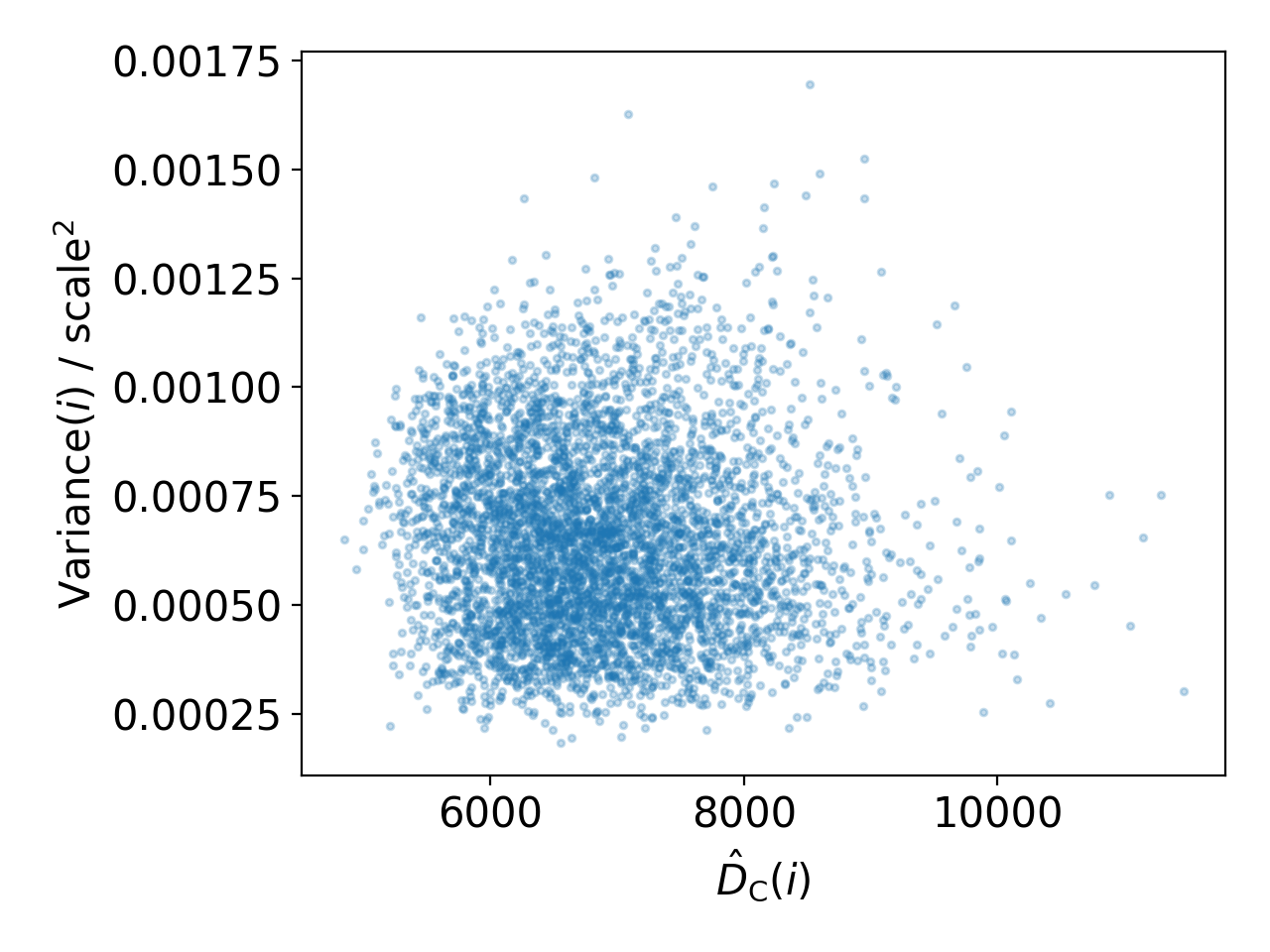}
  \includegraphics[width=0.32\textwidth]{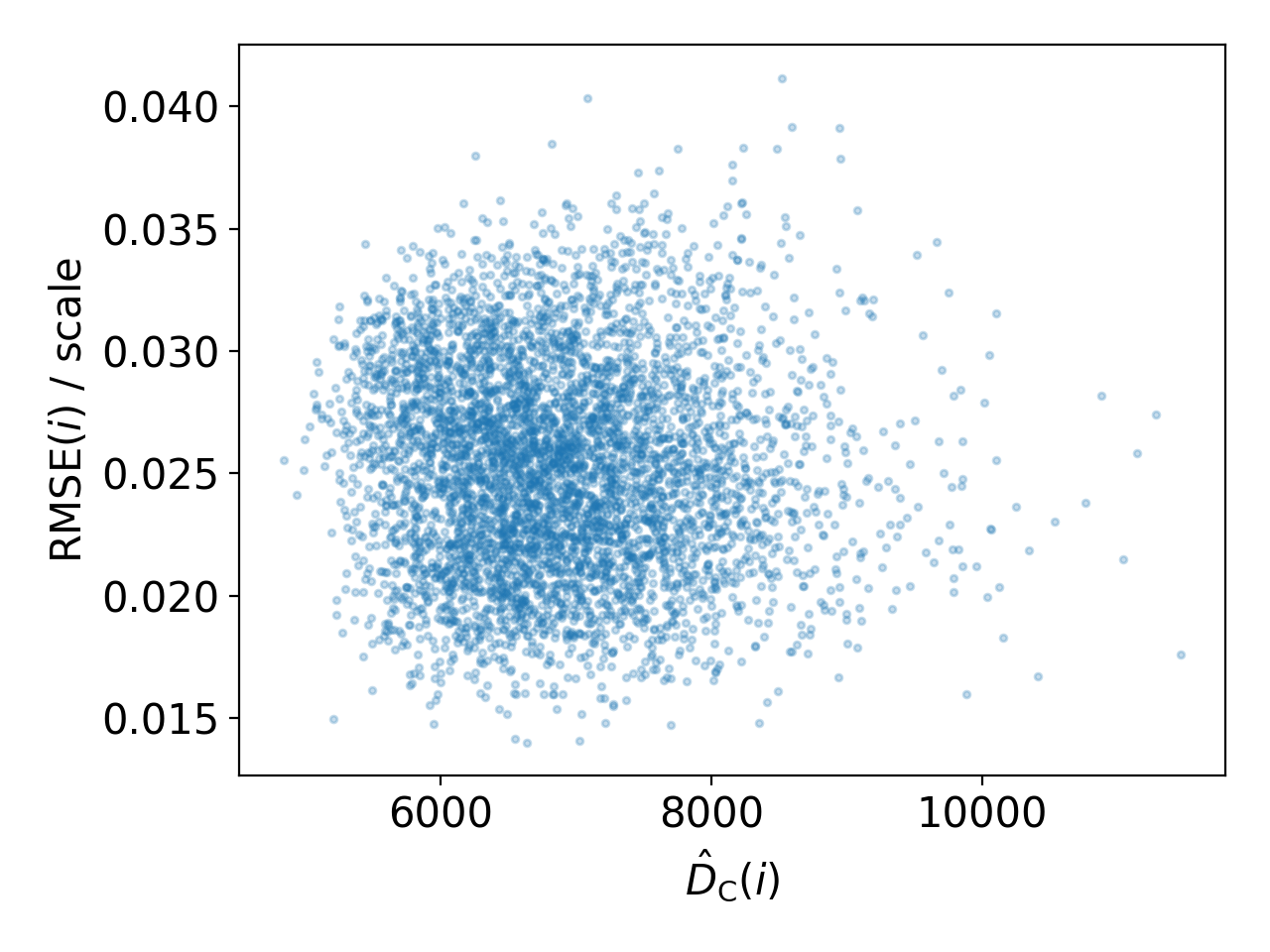}
  
  \medskip
  
  \includegraphics[width=0.32\textwidth]{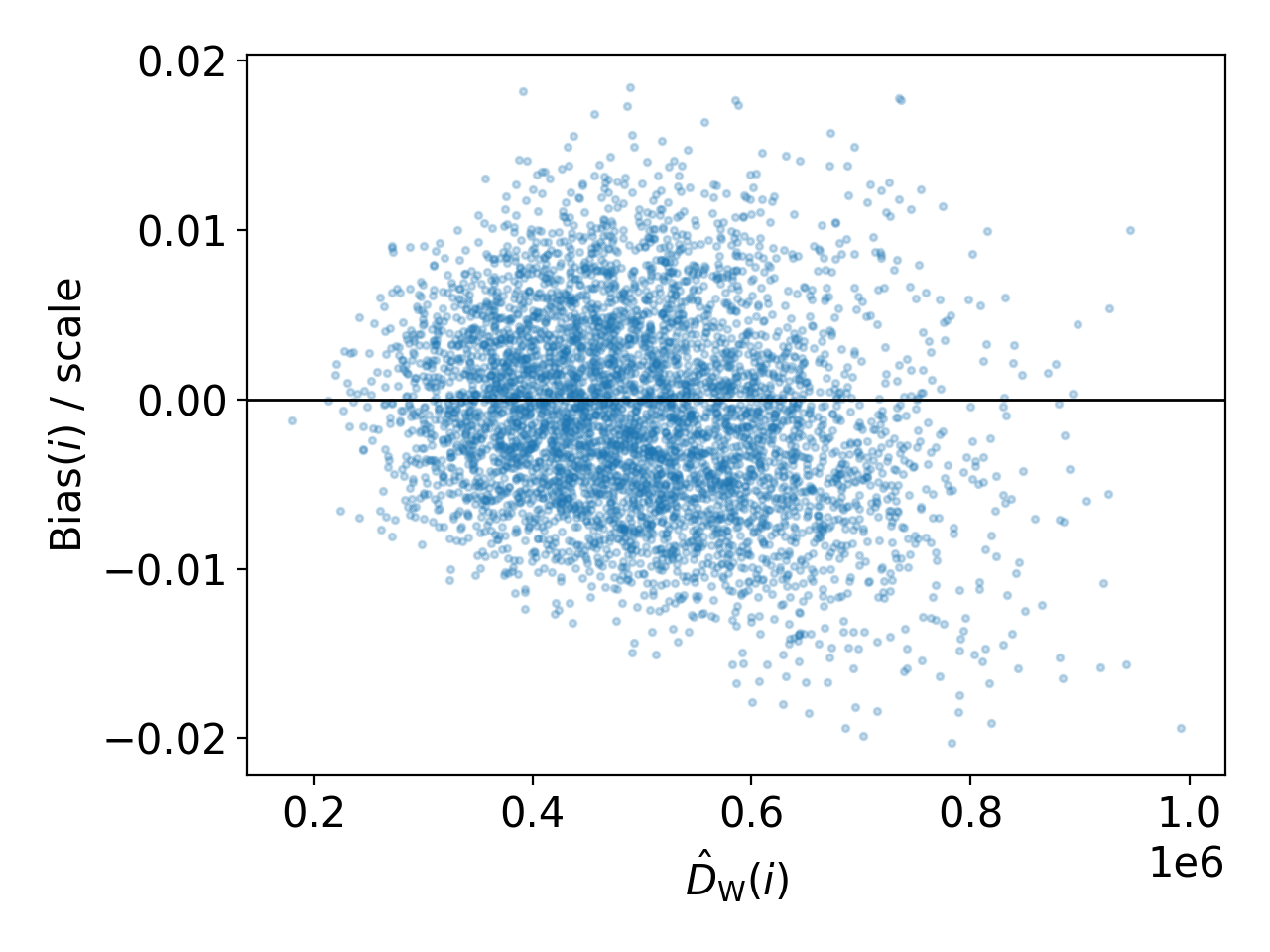}
  \includegraphics[width=0.32\textwidth]{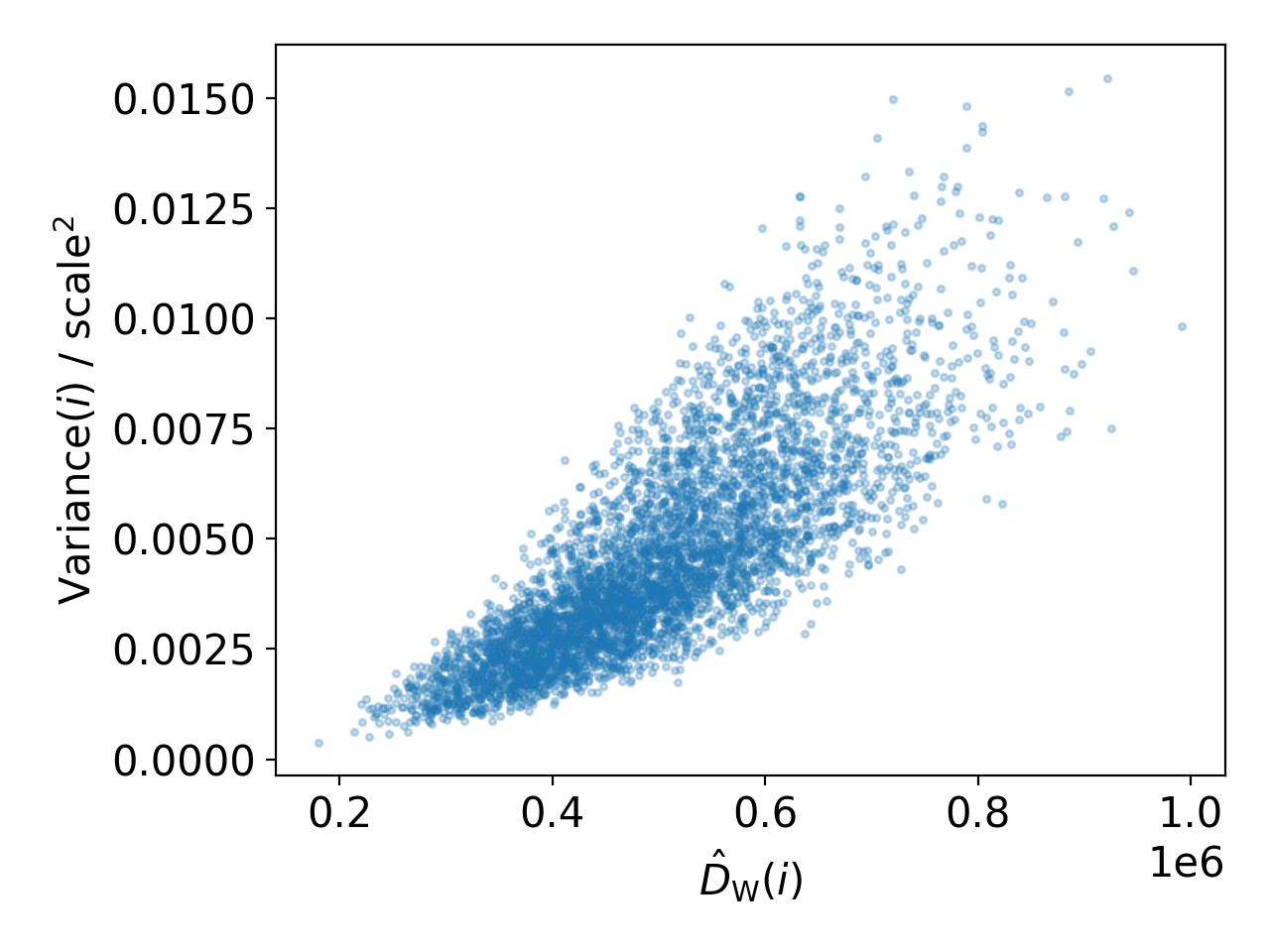}
  \includegraphics[width=0.32\textwidth]{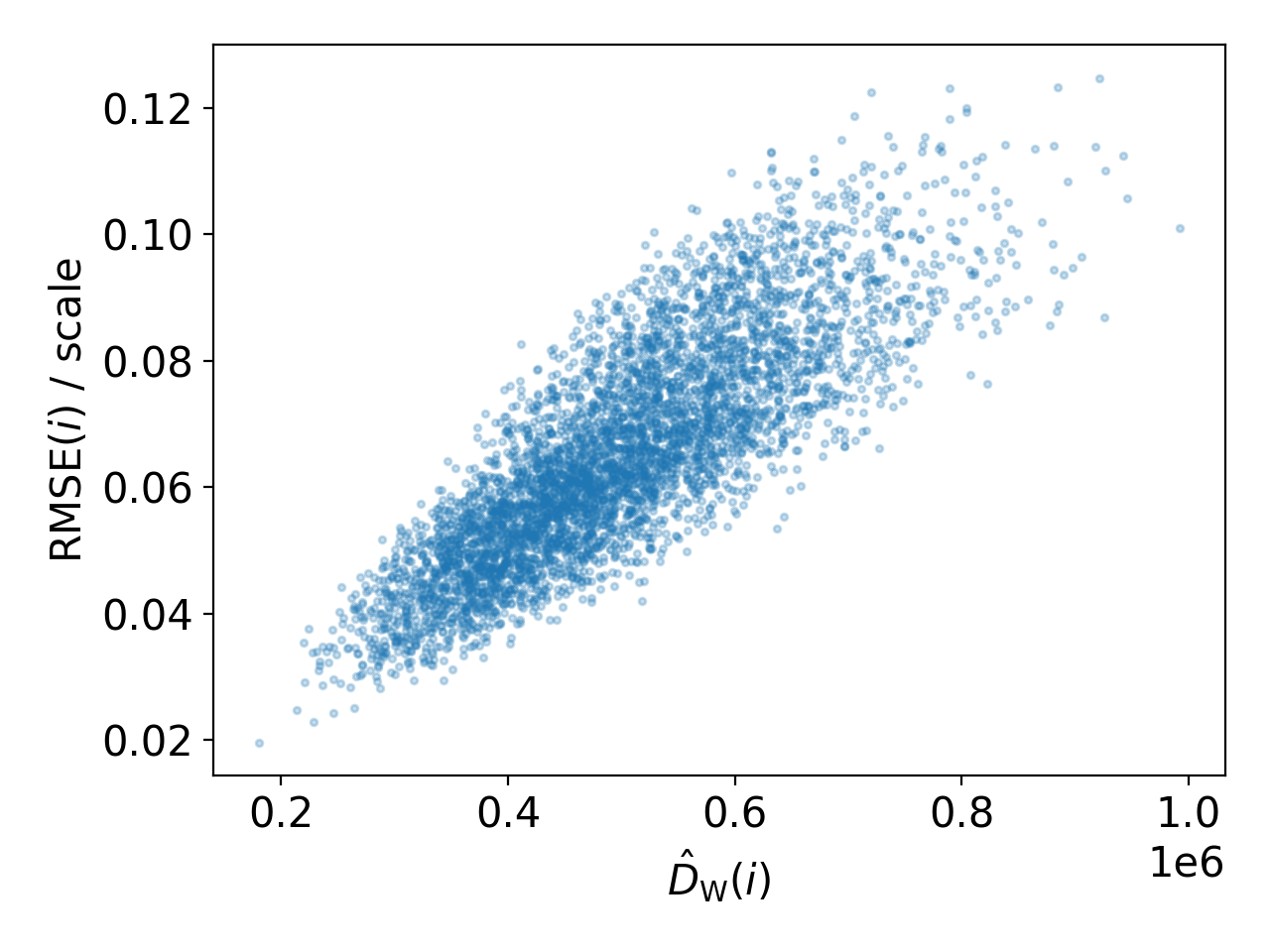}
  \caption{Scatter plots of $\hat{D}_*(i)$ and normalized $\mathrm{Bias}(i)$ (left column), $\mathrm{Variance}(i)$ (center column), $\mathrm{RMSE}(i)$ (right column). Each row represents the errors of each estimation method. KL divergences are computed using SigLIP.}
  \label{fig:kl_err_siglip}
\end{figure*}

\Cref{fig:kl_err_clip,fig:kl_err_siglip} shows estimation bias, variance, and RMSE of each approximation method. $\hat{D}_\mathrm{KL}(i)$ and $\hat{D}_\mathrm{KLR}(i)$ have negative estimation bias. The greater the KL divergence, the greater the bias and variance. On the other hand, $\hat{D}_\mathrm{C}(i)$ and $\hat{D}_\mathrm{W}(i)$ have little bias of finite sample approximation. Though the estimation variance of $\hat{D}_\mathrm{W}$ linearly grows when $\hat{D}_\mathrm{W}$ increases, the relative ratio of estimation error is still much less than that of $\hat{D}_\mathrm{KL}(i)$ and $\hat{D}_\mathrm{KLR}(i)$. These results suggest that nonlinear operations in these methods cause finite-sample approximation bias. However, all estimation methods have low estimation errors when the KL divergence is not high.

\subsection{Sample size and approximation error}
\begin{figure*}[t]
    \centering
    \begin{subfigure}{\linewidth}
        \centering
        \begin{minipage}{0.24\textwidth}
            \centering
            \includegraphics[width=\textwidth]{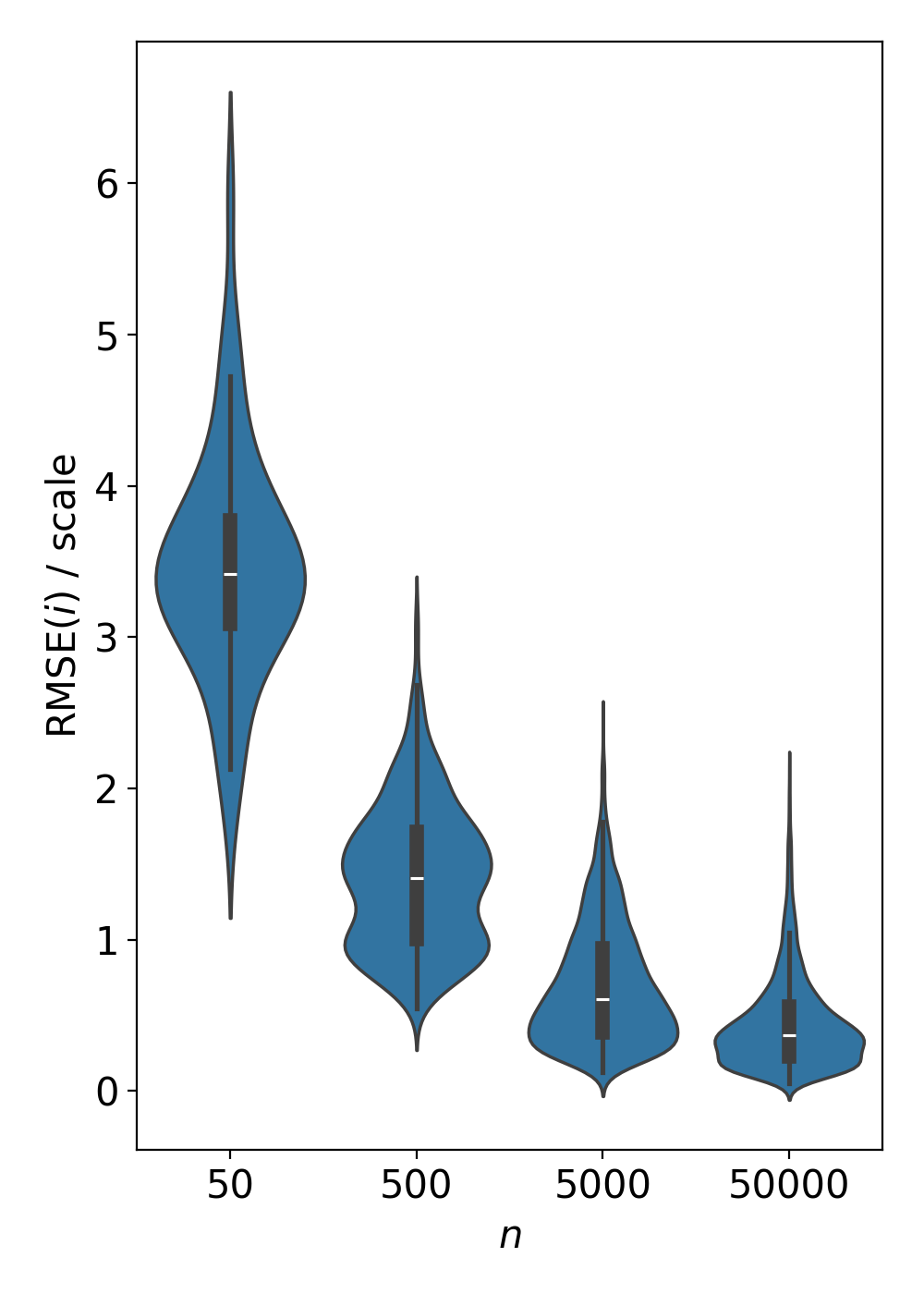}
            \subcaption{$D_\mathrm{KL}(i)$}
        \end{minipage}
        \begin{minipage}{0.24\textwidth}
            \centering
            \includegraphics[width=\textwidth]{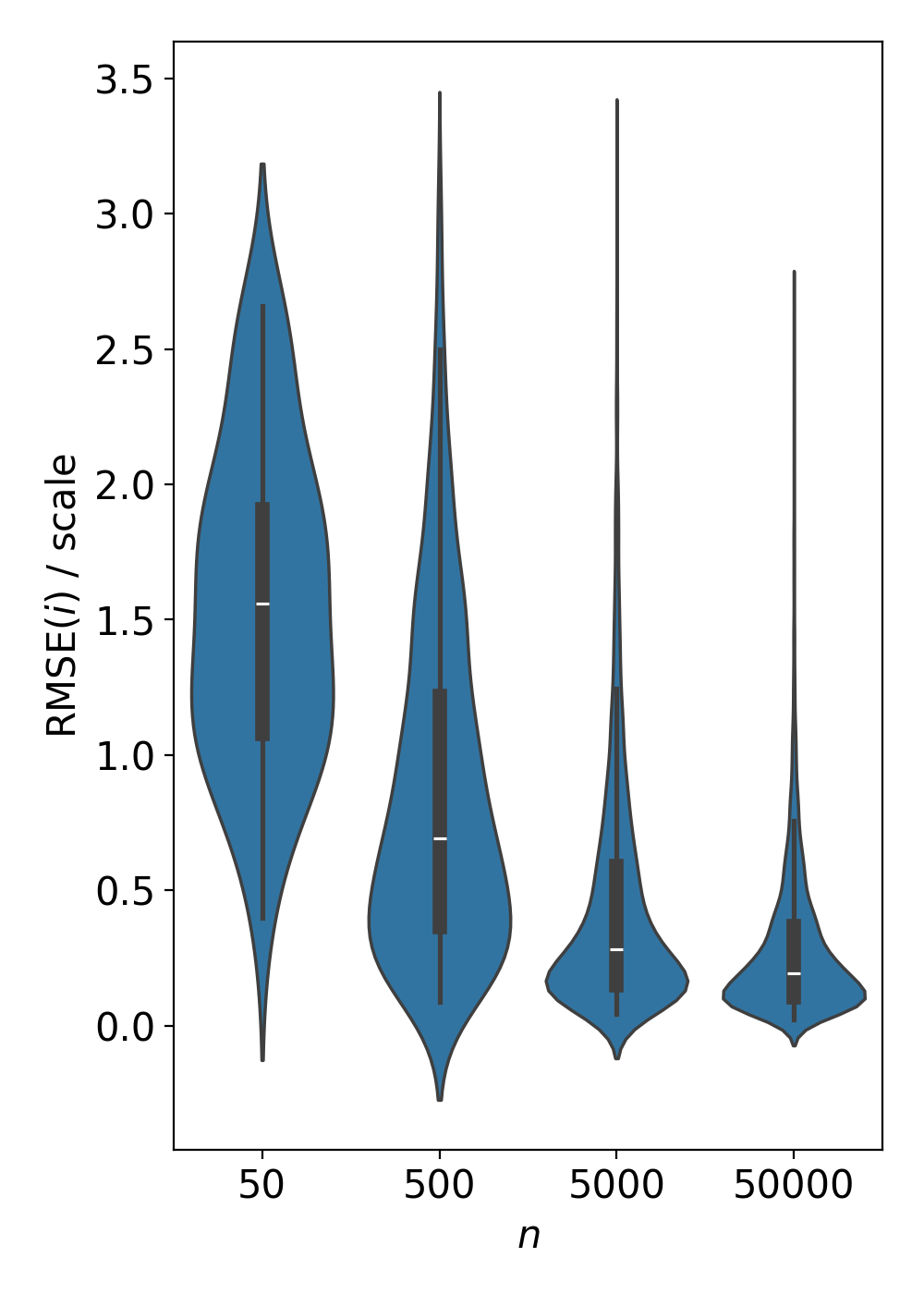}
            \subcaption{$D_\mathrm{KLR}(i)$}
        \end{minipage}
        \begin{minipage}{0.24\textwidth}
            \centering
            \includegraphics[width=\textwidth]{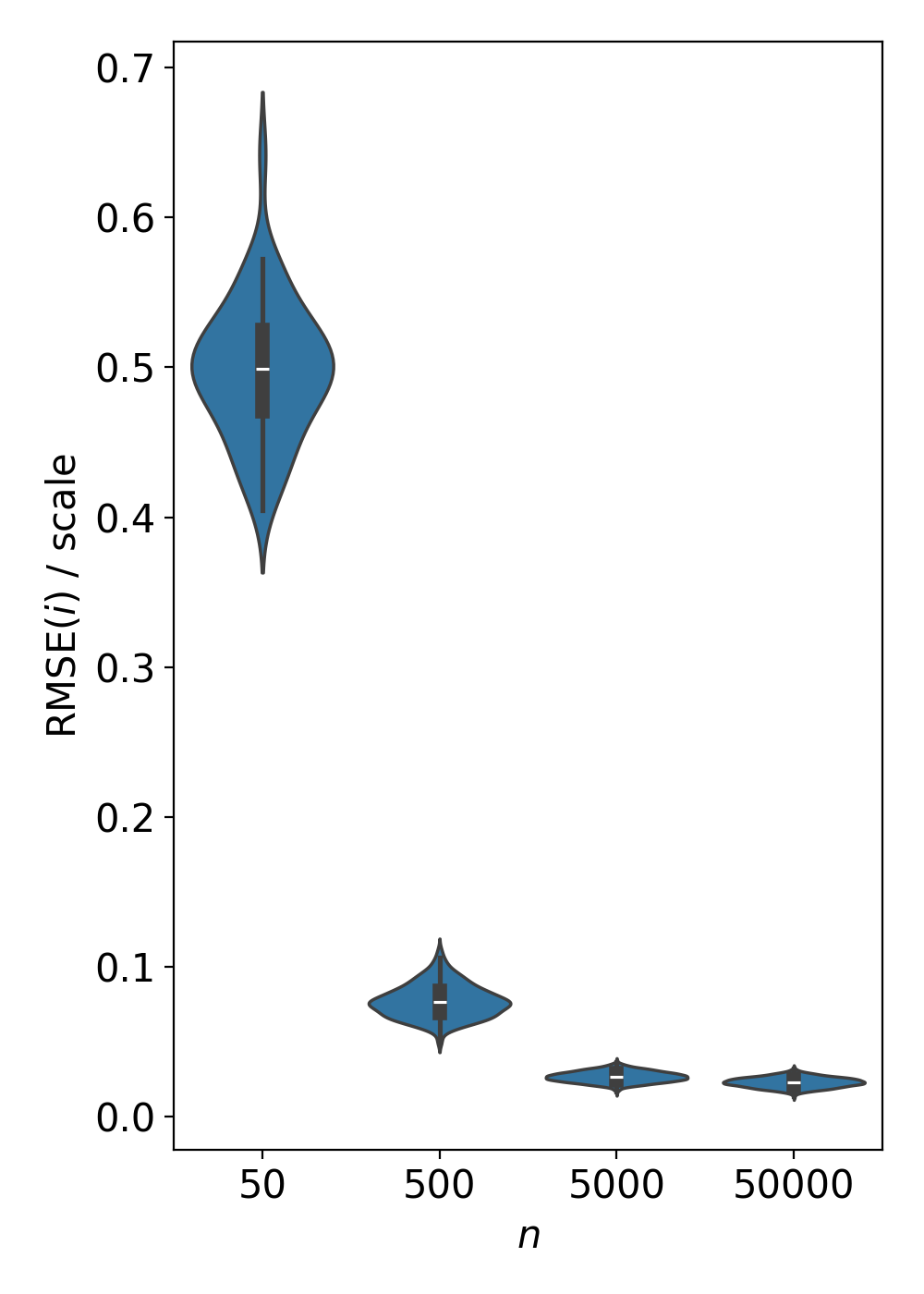}
            \subcaption{$D_\mathrm{C}(i)$}
        \end{minipage}
        \begin{minipage}{0.24\textwidth}
            \centering
            \includegraphics[width=\textwidth]{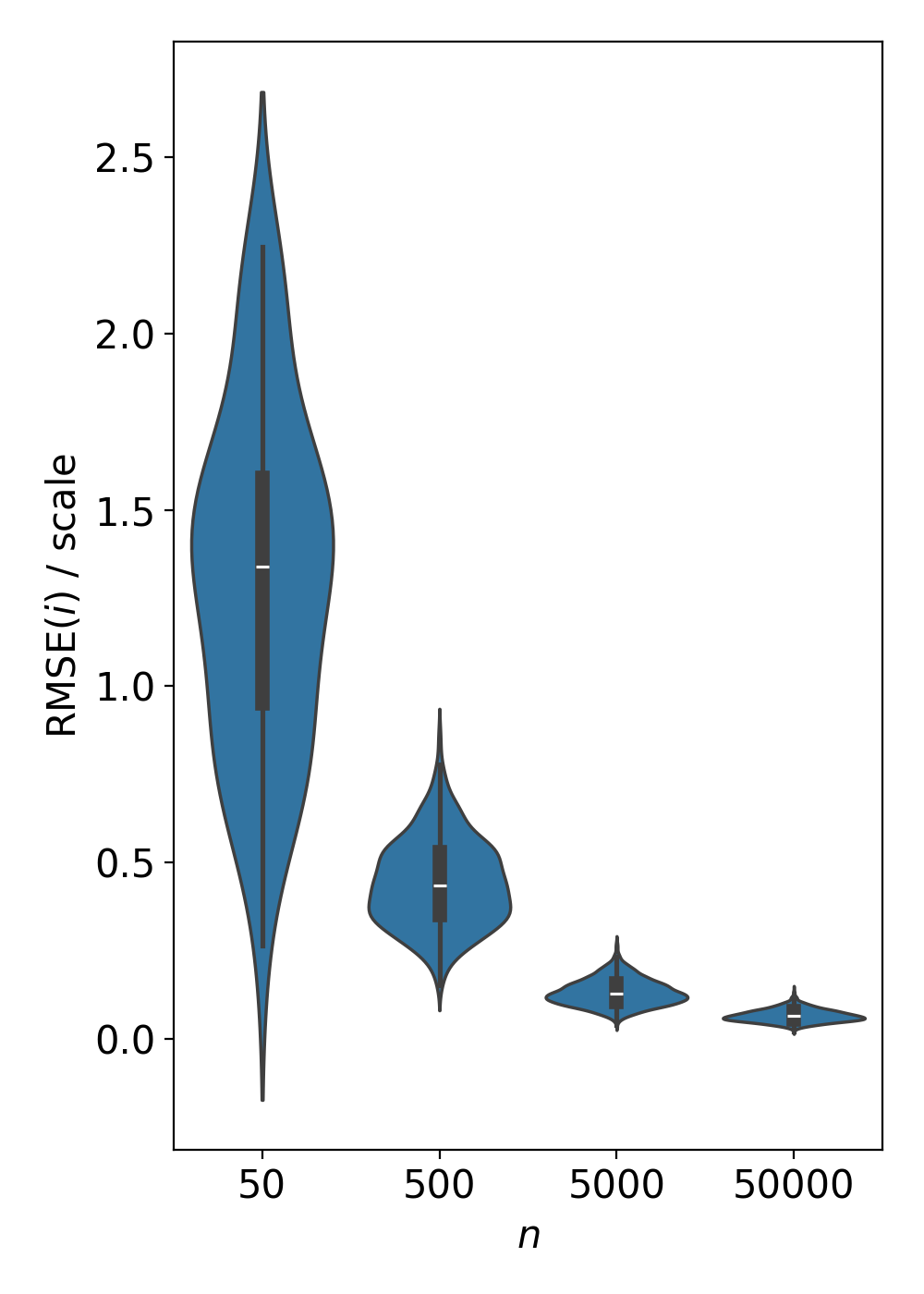}
            \subcaption{$D_\mathrm{W}(i)$}
        \end{minipage}
        \caption*{CLIP results}
    \end{subfigure}
    \hfill
    \vspace{1em}
    \begin{subfigure}{\linewidth}
        \centering
        \begin{minipage}{0.24\textwidth}
            \centering
            \includegraphics[width=\textwidth]{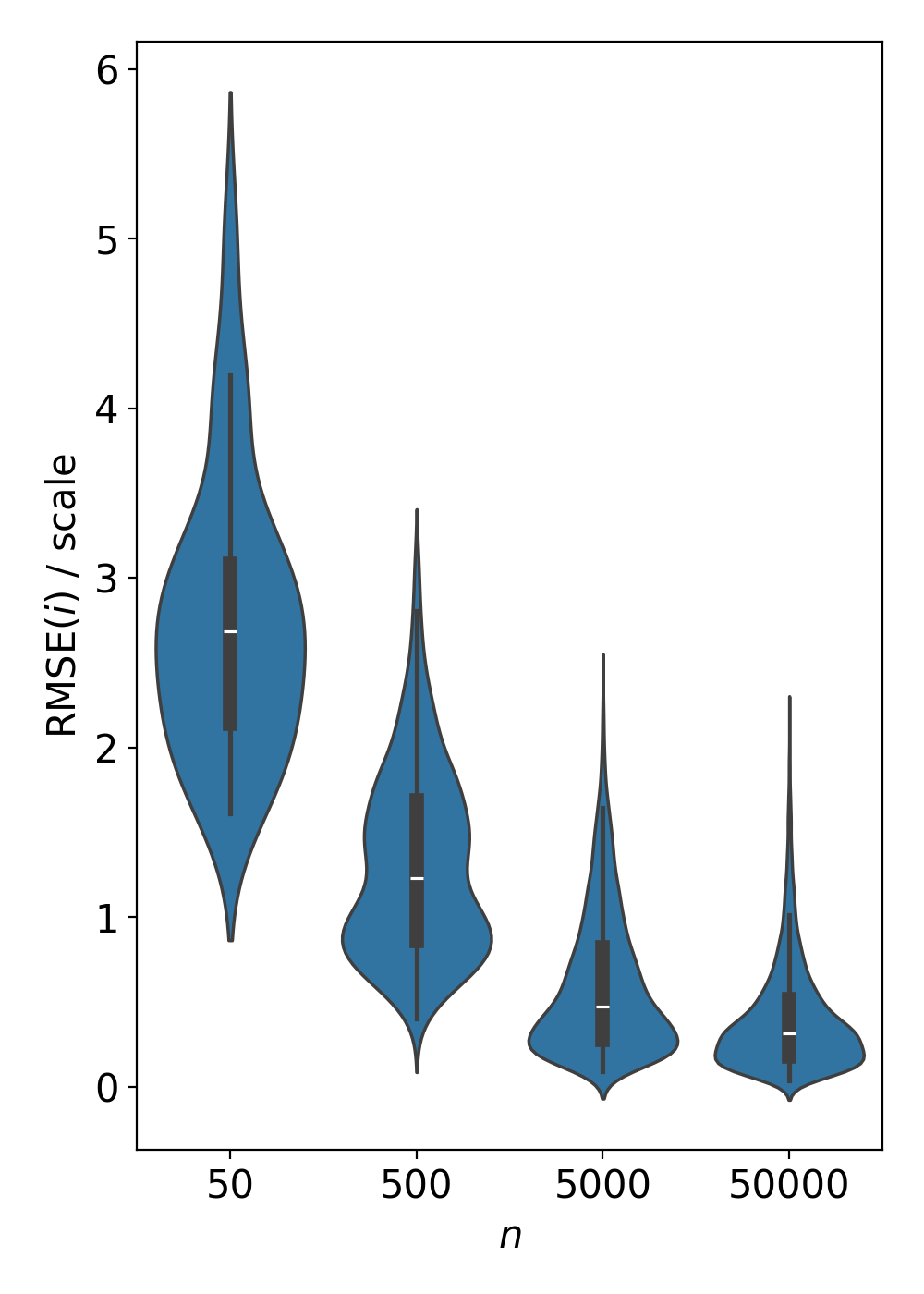}
            \subcaption{$D_\mathrm{KL}(i)$}
        \end{minipage}
        \begin{minipage}{0.24\textwidth}
            \centering
            \includegraphics[width=\textwidth]{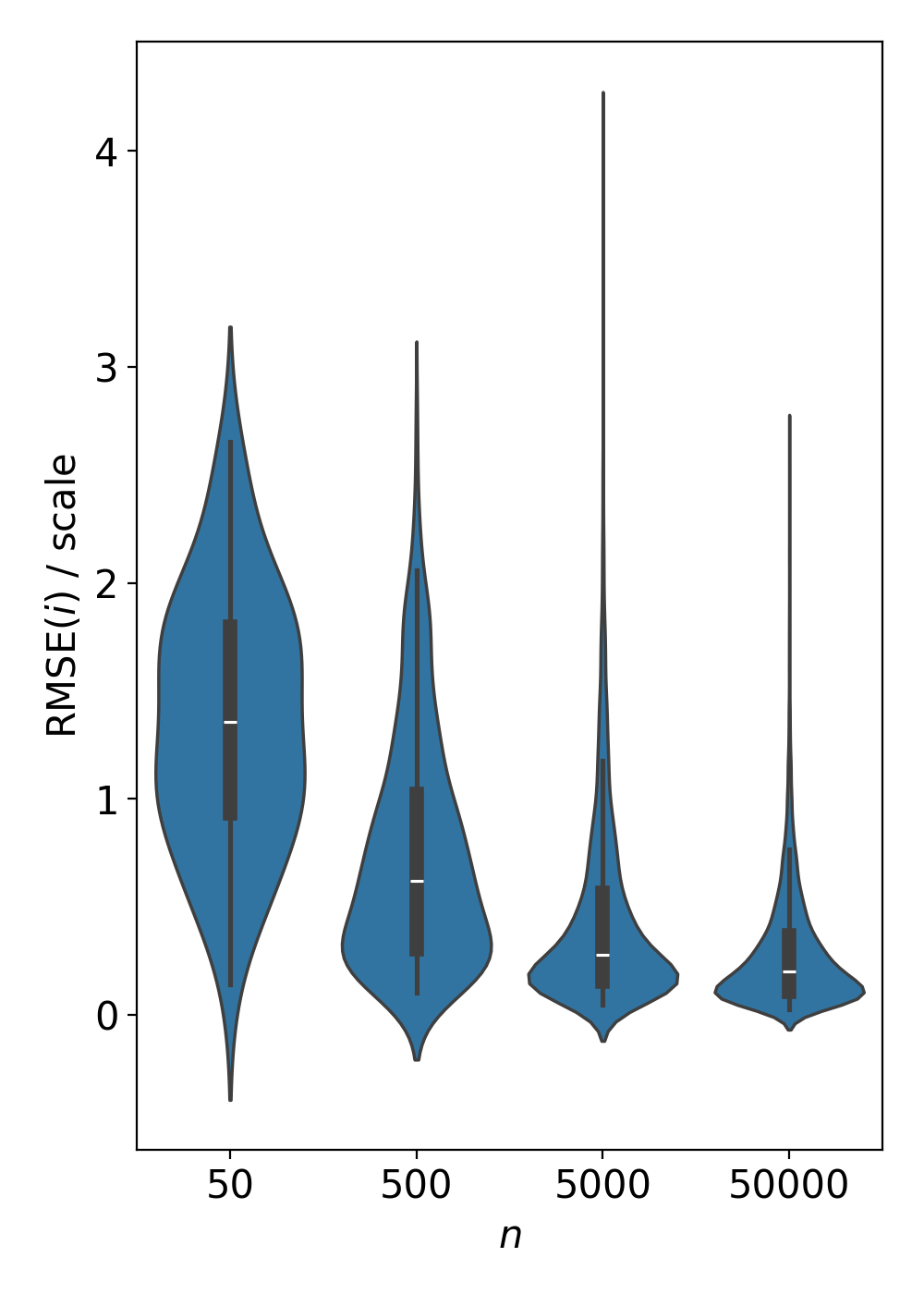}
            \subcaption{$D_\mathrm{KLR}(i)$}
        \end{minipage}
        \begin{minipage}{0.24\textwidth}
            \centering
            \includegraphics[width=\textwidth]{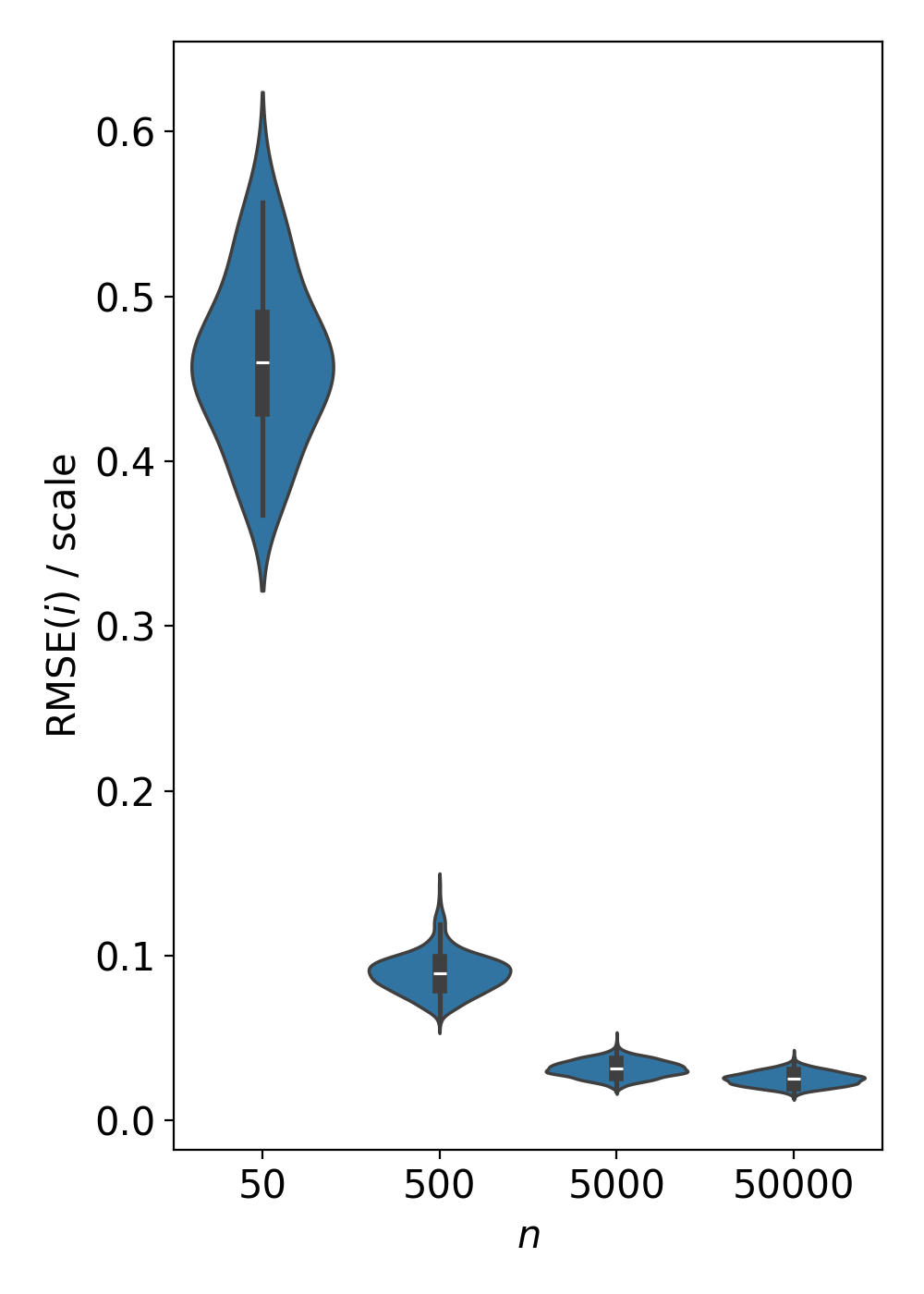}
            \subcaption{$D_\mathrm{C}(i)$}
        \end{minipage}
        \begin{minipage}{0.24\textwidth}
            \centering
            \includegraphics[width=\textwidth]{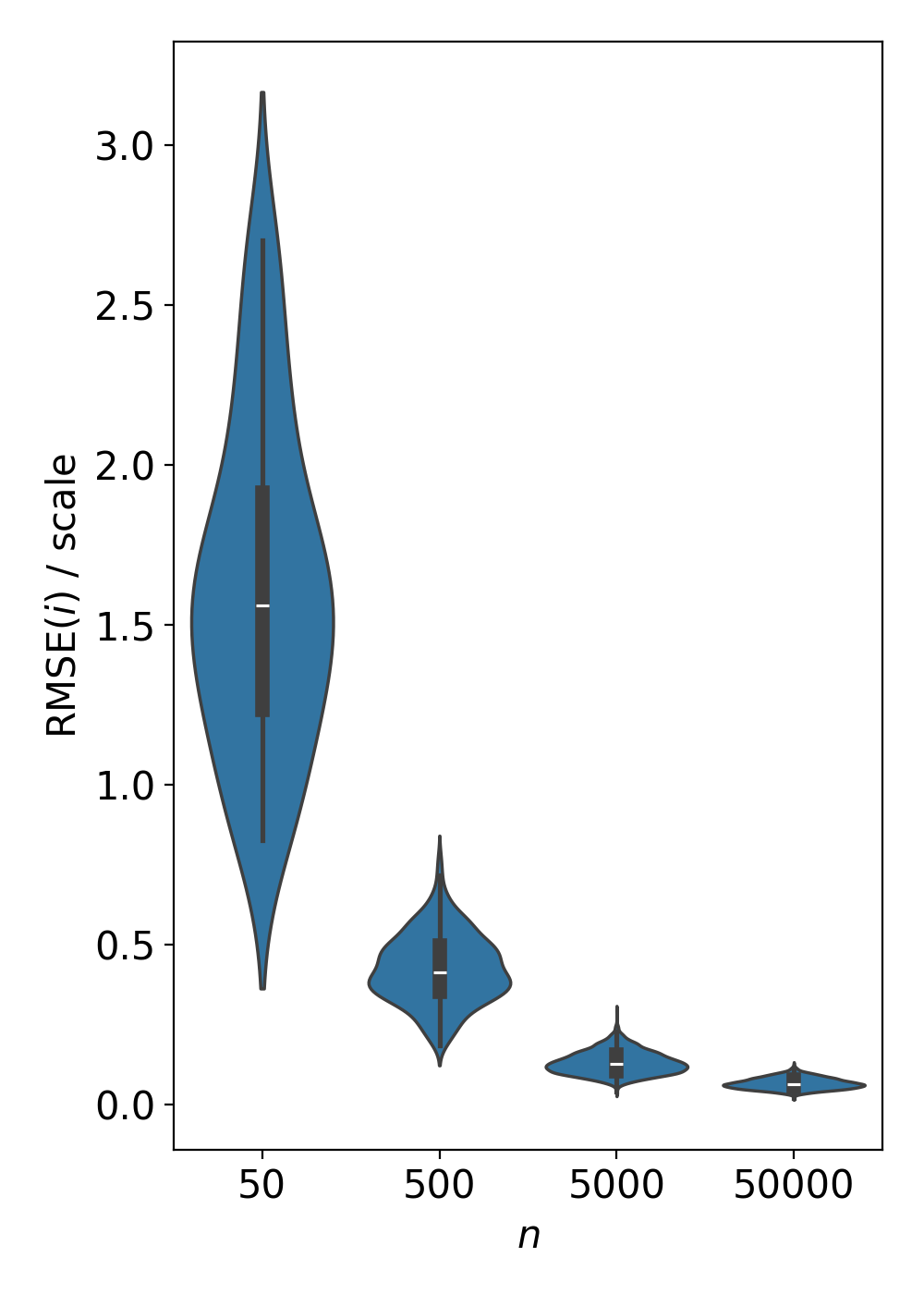}
            \subcaption{$D_\mathrm{W}(i)$}
        \end{minipage}
        \caption*{SigLIP results}
    \end{subfigure}
    \caption{Violin plot of $\mathrm{RMSE}(i)$ on each estimation method across different sample size $n$.}
    \label{fig:err_n}
\end{figure*}

We also examined how the sample size $n$ affects the error arising from finite-sample approximation. For each $n$, we sampled $n$ image-text pairs from 50,000 pairs in MSCOCO 100 times, and calculated estimation RMSE using the bootstrap method. \Cref{fig:err_n} shows RMSE distribution on each sample size. This indicates that around 5000 samples are sufficient for all KL divergence estimation methods.

\end{document}